%% file: main.tex
\documentclass[runningheads]{llncs}
\usepackage{graphicx}
\usepackage{tikz}
\usepackage{comment}
\usepackage{amsmath,amssymb} 
\usepackage{color}
\usepackage[normalem]{ulem}
\usepackage{booktabs}
\usepackage{multirow}
\usepackage{subcaption}

\usepackage{pifont}
\usepackage[export]{adjustbox}
\definecolor{MyGreen}{cmyk}{100, 0, 100, 0}
\usepackage[pagebackref,breaklinks,colorlinks,citecolor=MyGreen]{hyperref}

\newcommand{\figcspace}{\vspace{0mm}}
\newcommand{\figspace}{\vspace{-5mm}}
\newcommand{\tabcspace}{\vspace{0mm}}
\newcommand{\tabspace}{\vspace{-5mm}}
\newcommand{\fix}[1]{\textcolor{black}{#1}}

\newcommand{\etal}{\textit{et al.}}
\DeclareCaptionLabelFormat{andtable}{#1~#2  \&  \tablename~\thetable}
\begin{document}

\pagestyle{headings}
\mainmatter

\title{Pay Attention to Hidden States for Video Deblurring: Ping-Pong Recurrent Neural Networks and Selective Non-Local Attention} 

\titlerunning{PAHS for Video Deblurring}
\authorrunning{J. Park et al.}

\author{Joonkyu Park$^{1}$ \and Seungjun Nah$^{1,3}$ \and Kyoung Mu Lee$^{1,2}$}
\institute{
$^{1}$Dept. of ECE \& ASRI, $^{2}$IPAI, Seoul National University, Korea \quad $^{3}$NVIDIA\\
}

\maketitle

\input{sections/0_abstract}
\input{sections/1_introduction}
\input{sections/2_relatedworks}
\input{sections/3_proposedmethods}

\input{sections/4_experimentalresults}
\input{sections/5_conclusion}


\clearpage
\bibliographystyle{splncs04}
\bibliography{main}
\end{document}


\pagestyle{headings}
\mainmatter

\title{\emph{Supplementary Material for}\\Pay Attention to Hidden States for Video Deblurring: Ping-Pong Recurrent Neural Networks and Selective Non-Local Attention}

\renewcommand{\thetable}{S\arabic{table}}
\renewcommand{\thesection}{S\arabic{section}}
\renewcommand{\thefigure}{S\arabic{figure}}
\renewcommand{\theequation}{S\arabic{equation}}
\renewcommand{\thealgorithm}{S\arabic{algorithm}}

\newcommand{\citenumber}[1]{[\textcolor{MyGreen}{#1}]}
\newcommand{\refnumber}[1]{\textcolor{red}{#1}}

\titlerunning{PAHS for Video Deblurring}
\authorrunning{J. Park et al.}

\author{Joonkyu Park$^{1}$ \and Seungjun Nah$^{1,3}$ \and Kyoung Mu Lee$^{1,2}$}
\institute{
$^{1}$Dept. of ECE \& ASRI, $^{2}$IPAI, Seoul National University, Korea \quad $^{3}$NVIDIA\\
}

\maketitle
\section{Introduction}
In the main manuscript, we proposed a new RNN-based video deblurring framework, PAHS (Pay Attention to Hidden States) that greatly enhance the deblurring performance by maximizing the power of hidden states.
In this supplementary material, we provide the implementation details, further ablation studies, and more comparative test results of our proposed PAHS.  
In Section~\ref{sec_supp:model_specifics}, the model specifications that we used for the experiments in the main manuscript are described.
In Section~\ref{sec_supp:ablations}, we further justify the design choice of PPRNN and SNLA by comparing with the other design configurations, respectively.
In Section~\ref{sec_supp:cnn_problem}, we justify the problem of CNN-based video deblurring methods~\citenumber{14,21}, which fail to restore the structure of the scene in the consecutive blurry frames.
In Section~\ref{sec_supp:large_motion}, we validate the efficacy of our PAHS in restoring sharp details on large blur dataset.
In Section~\ref{sec_supp:visual_comparisons}, we provide additional qualitative comparisons of deblurred results with the other state-of-the-art video deblurring methods.
\input{supplementary/fig/model_all}

\section{Architecture Specifications}
\label{sec_supp:model_specifics}
In Figure~\ref{sup_fig:model}, we show the overall architecture of our PAHS framework again.
As explained in the main manuscript, PPRNN handles the hidden states to gather useful information from the previous and the current features, and SNLA solves the misalignment issue between the hidden states~$\tilde{h}_{t-1}$ and the current blurry feature~$f_{B_t}$.
In this section, we cover the detailed specifications of each module~(feature extractor, reconstructor, hidden state extractor, PPRNN, SNLA).
Let us denote the number of channels of the feature~$f_{B_t}$, extracted from the blurry image~${B_t}$, as $c$.
For our final model, we set $c = 192$.

\subsection{Feature Extractor Architecture}
\input{supplementary/table/feandre}
In Table~\ref{sup_tab:feature_extractor}, we show the architecture details of the feature extractor module.
We build the feature extractor with a convolutional layer, ResBlocks, and intermediate strided convolution layers.
From an input blurry frame $B_{t} \in \mathbb{R}^{h \times w \times 3}$, a blurry frame feature $f_{B_{t}} \in \mathbb{R}^{h/4 \times w/4 \times c}$ is extracted, where $h$ and $w$ denote height and width of the input image, and $c$ denotes the number of channels of $f_{B_{t}}$, respectively.
In our implementation, the feature extractor consists of two strided convolution layers for downsampling, and five residual blocks are employed after each convolution layers.

\subsection{Reconstructor Architecture}
In Table~\ref{sup_tab:reconstructor}, we show the architecture details of the feature extractor module.
We set the reconstructor by taking the reverse of the order of feature extractor and replacing the strided convolutions with transposed convolutions.
However, we used less number of ResBlocks than the feature extractor.
In order to extract the hidden state, we split the reverse module into first 3 ResBlocks and the rest, which is the reconstructor.

\subsection{Hidden State Extractor Architecture}
Table~\ref{sup_tab:hidden_state_extractor} shows the architecture of the hidden state extractor.
We design the hidden state extractor with two convolution layers and a residual block in between.
Given a latent frame feature $f_{L_t} \in \mathbb{R}^{c \times h/4 \times w/4}$, it outputs the hidden state $h_{t} \in \mathbb{R}^{c/3 \times h/4 \times w/4}$.
\input{supplementary/table/heandpprnn}

\subsection{PPRNN Architecture}
\input{supplementary/table/pprnn_algo}
In Figure~\ref{sup_fig:pprnn}, we illustrate the architecture of a PPRNN cell.
PPRNN updates the input hidden state $h_{t-1}^{(i-1)}$ alternately with $f_{B_{t}}$ and $f_{L_{t-1}}$ in each ping-pong step.
Specifically, the weights used to update the input state via different inputs are shared.
The architecture details are shown in Table~\ref{sup_tab:pprnn} with the inference process shown in Algorithm~\ref{algo:pprnn}.

\subsection{SNLA Architecture}
\input{supplementary/table/snla_algo}
In Figure~\ref{sup_fig:snla}, we illustrate the architecture of SNLA module structure.
From the blurry frame feature $f_{B_{t}}$ and the updated hidden state from PPRNN $h_{t-1}^{(n)}$, query and the key features are computed by a strided convolution, respectively.
The query and key are matrix-multiplied to produce a non-local attention map, $S_{\text{NL}}$.
In order to suppress undesirably high attention scores from small correlation, we compute a selection score, $S_{\text{Sel}}$ that ranges from 0 to 1 with a filtering module.
$S_{\text{NL}}$ and $S_{\text{Sel}}$ are multiplied to produce a scaled attention map, and then matrix-multiplied with the value feature from $h_{t-1}^{(n)}$.
Finally, $\tilde{h}_{t-1}$ is generated by the following transposed convolutions.
The architecture details are shown in Table~\ref{sup_tab:snla} with the inference process shown in Algorithm~\ref{algo:snla}.
\input{supplementary/table/snla}

\section{Further Ablation studies}
\label{sec_supp:ablations}
\subsection{PPRNN Ablation: Order of $f_{B_{t}}$ and $f_{L_{t-1}}$}
In the main manuscript, we proposed PPRNN that updates the hidden state 
$h_{t-1}^{(i-1)}$ by $f_{B_{t}}$ first, and then with $f_{L_{t-1}}$.
Table~\ref{sup_tab:pprnn_ablation} shows that the proposed update order gives consistently better results than the reverse order for different recurrence numbers.
As $h_{t-1}^{(0)}$ was generated from $f_{L_{t-1}}$ in the previous time step, updating with $f_{L_{t-1}}$ would cause redundancy and may lead to inferior performance.
Also, if the last update is with $f_{B_{t}}$, the following concatenation with $f_{B_{t}}$ with or without SNLA could also make the feature redundant, harnessing the extraction of complementary information.
\input{supplementary/table/pprnn_ablation}

\subsection{SNLA Ablation: Cross-Attention}
In contrast to the commonly used self-attention~\citenumber{27}, in our SNLA, we use the cross-attention between $h_{t-1}^{(n)}$ and $f_{B_{t}}$ to update the hidden state.
Table~\ref{sub_tab:snla_ablation} shows that our cross-attention with heterogeneous inputs to SNLA exhibits better performance than the traditional self-attention scheme.
As we aim to rearrange $h_{t-1}^{(n)}$ to help find the deblurred image from the blurry frame feature $f_{B_{t}}$, using $f_{B_t}$ as a query leads to better performance.
\input{supplementary/table/snla_ablation}

\subsection{Smaller version of PAHS: PAHS-S}
In addition to our PAHS, we show a smaller version of PAHS, PAHS-S, to validate the efficacy of our framework in Figure~\ref{sup_tab:pahs_s}.
our PAHS-S with $c$, the channel of the feature~$f_{B_t}$, set to $92$, achieves comparable PSNR as ARVo~\citenumber{14}.
Specifically, PAHS-S is 4.3 times smaller and 27.8 times faster than ARVo~(model size: 4.8MB/fps: 4.17).
\input{supplementary/table/pahs-s}

\section{Problem of CNN-based video deblurring methods}
\label{sec_supp:cnn_problem}
In Line 565-571 in the main manuscript, we maintain the problem of CNN-based methods~\citenumber{14,21} that they fail to deblur the scene objects if the consecutive frames are severely blurred.
We evaluate the deblurring performance by replacing one of the consecutive blurry frames with the ground truth sharp frame.
While our method PAHS does not enhance the performance by replacing the neighboring frame~(34.24/0.938 to 34.21/0.937), replacing a neighboring frame with a ground truth frame in CNN-based video deblurring methods enhances the performance quantitatively and qualitatively in Figure~\ref{sup_fig:cnn_based}. 
This justifies that CNN-based video deblurring methods can suffer in consecutive blurry frames.
\input{supplementary/fig/cnn_based}

\section{Robustness to abrupt motions}
\label{sec_supp:large_motion}
In Line 578-583 in the main manuscript, we claim our PAHS's robustness to large and abrupt motions.
We quantitatively evaluate the deblurring performance in large-displacement data by choosing even-numbered test frames from GOPRO~\citenumber{19} dataset in Table~\ref{sup_tab:large_motion}.
Our PAHS and PAHS-S exhibit large performance gaps.
\input{supplementary/table/large_motion}

\section{Visual Comparison of Deblurred Videos}
\label{sec_supp:visual_comparisons}
We show more visual comparisons in Figures~\ref{fig:real_comparisons}, \ref{fig:dvd_comparisons}, \ref{fig:gopro_comparisons}, \ref{fig:reds_comparisons} with the deblurred results on real blur datasets, DVD~\citenumber{24}, GOPRO~\citenumber{19}, and REDS~\citenumber{18}, respectively.
These results clearly  demonstrate that our proposed PAHS produces much sharper and clearer outputs for various scenes with different blurs including real ones than the state-of-the-art methods.

\subsection{Results Videos in .mp4 Format}
\label{sec_supp:video_results}
We present the video clips showing the deblurred results of our PAHS with the blurry input in the supplementary videos.
Please see the attached videos for the comprehensive comparison.

\noindent
\textbf{\fontfamily{pcr}\selectfont DVD.mp4}:\\
Comparison between DVD blur video~\citenumber{24} and deblurred video with PAHS.

\noindent
\textbf{\fontfamily{pcr}\selectfont GOPRO.mp4}:\\
Comparison between blur GOPRO video~\citenumber{19} and deblurred video with PAHS.

\noindent
\textbf{\fontfamily{pcr}\selectfont REDS.mp4}:\\
Comparison between blur REDS video~\citenumber{18} and deblurred video with PAHS.

\noindent
\textbf{\fontfamily{pcr}\selectfont REAL.mp4}:\\
Comparison between real blur video and deblurred video with PAHS.

\clearpage
\input{supplementary/fig/real_comparison}
\input{supplementary/fig/dvd_comparison}
\input{supplementary/fig/gopro_comparison}
\input{supplementary/fig/reds_comparison}


\clearpage

%% file: sections/0_abstract.tex
\begin{abstract}
    Video deblurring models exploit information in the neighboring frames to remove blur caused by the motion of the camera and the objects.
    Recurrent Neural Networks~(RNNs) are often adopted to model the temporal dependency between frames via hidden states.
    When motion blur is strong, however, hidden states are hard to deliver proper information due to the displacement between different frames.
    While there have been attempts to update the hidden states, it is difficult to handle misaligned features beyond the receptive field of simple modules.
    Thus, we propose 2 modules to supplement the RNN architecture for video deblurring.
    First, we design Ping-Pong RNN~(PPRNN) that acts on updating the hidden states by referring to the features from the current and the previous time steps alternately.
    PPRNN gathers relevant information from the both features in an iterative and balanced manner by utilizing its recurrent architecture.
    Second, we use a Selective Non-Local Attention~(SNLA) module to additionally refine the hidden state by aligning it with the positional information from the input frame feature.
    The attention score is scaled by the relevance to the input feature to focus on the necessary information.
    By paying attention to hidden states with both modules, which have strong synergy, our PAHS framework improves the representation powers of RNN structures and achieves state-of-the-art deblurring performance on standard benchmarks and real-world videos.

\end{abstract}

%% file: sections/1_introduction.tex
\section{Introduction}
\input{sections/figs/overview}


Video recordings often suffer from motion blur due to the scene dynamics with various moving objects.
Recovering sharp frames from the observed blurry frames is a challenging problem due to the complexity of blur and the changes of scene content.
With the spatio-temporally varying information in multiple frames, the key to deblurring a video has been the way to aggregate complementary information in the neighboring frames.
However, gathering relevant information from the misaligned frames is a nontrivial issue.

While the misalignment between the video frames can be handled by compensating the motion explicitly~\cite{Kim_2018_ECCV,li2021arvo,Pan_2020_CVPR,Su_2017_CVPR}, spatio-temporally varying nonlinear dynamics is difficult to model.
In order to avoid the modeling complexity, RNN-based methods were developed to extract and deliver necessary information, implicitly by hidden states~\cite{Kim_2017_ICCV,Nah_2019_CVPR,Wieschollek_2017_ICCV,zhong2020efficient}.
As computationally expensive optical flow can be omitted, RNN-based methods often exhibit fast processing speed.
However, hidden states are extracted without knowing the scene dynamics at the next time steps and may contain irrelevant or misaligned information.
Thus, with the propagation of the hidden states to future frames, the way to manipulate them has played a key role in RNN architectures.


Therefore, most RNN-based methods tried to handle the scene discrepancy between the time steps by updating the hidden states to adapt to the next frames~\cite{Kim_2017_ICCV,Nah_2019_CVPR,Zhou_2019_ICCV}.
Dynamic temporal blending~\cite{Kim_2017_ICCV}, intra-frame iterations~\cite{Nah_2019_CVPR}, and filter-adaptive convolutions~\cite{Zhou_2019_ICCV} were used to inject information and adapt it to the target blurry frame.
However, as they utilize only the current blurry frame feature to update hidden states, the adaptation could be biased toward the current time and may lead to forgetting what was provided from the past~\cite{Nah_2019_CVPR}.
Also, the previous methods have limitations in handling large displacements, bounded by the receptive field of the convolutional layers due to the fixed structure~\cite{Kim_2017_ICCV,Zhou_2019_ICCV}.




To overcome the limitation of conventional RNNs in handling hidden states jointly with the current blurry frame features, we propose two modules supplementing the architecture.
First, we present a Ping-Pong RNN~(PPRNN) module that operates on the hidden states in a balanced manner via internal recurrence.
Unlike IFI-RNN~\cite{Nah_2019_CVPR}, PPRNN uses not only the current blurry frame feature but also the {\em past deblurred frame feature} jointly to update the hidden states.
\fix{We use the module with the term, {\em ping-pong}, to indicate that both the past and the present features update the hidden states alternately in succession like ping-pong players.}
Using both the current and past frame features together, PPRNN can extract useful information by alternating optimization and prevent loss of necessary information from either side.
Furthermore, the recurrence nature allows the indefinitely deep computational path to find more optimal hidden states without having to increase the model size.

Second, we propose a Selective Non-Local Attention~(SNLA) module to refine the hidden state processed by PPRNN.
\fix{Unlike RNNs managing long-term dependencies by flowing the hidden states from the temporal steps, SNLA plays a role in handling long-range spatial dependencies, utilizing spatially distant positional information.}
SNLA computes non-local spatial correspondence between the current frame feature and the hidden state to gather information from distant spatial locations.
Different from the typical self-attention scheme~\cite{vaswani2017attention} that constrains the sum of attention values to be 1, our heterogeneous inter-feature correlation is not always guaranteed to be strong, especially in case of strong scene changes.
Thus, we employ an attention filtering technique to effectively suppress unnecessary information and extract only the useful part aligned with the frame to be deblurred.


Finally, we use the overall RNN architecture in a bidirectional manner to further improve the deblurring performance.
By using the past and the future information jointly, the restoration accuracy could be significantly boosted with a slight modification.
We validate the effectiveness of our PPRNN and SNLA modules with rigorous ablation study and comparisons with state-of-the-art video deblurring methods.
Furthermore, our method can be plugged into any RNN-based video deblurring methods, as shown in the generalization experiments.

We summarize our contributions as follows:
\begin{itemize}
    \item We introduced a novel Ping-Pong RNN (PPRNN) module to update hidden states effectively with balance by alternately providing information from the past and the current features.
    \item We proposed a Selective Non-Local Attention (SNLA) module to find a proper attentive correlation between the hidden state and the video frame feature in a non-local manner to handle misalignment from strong dynamics.
    \item We show that our proposed PAHS video deblurring framework that consists of PPRNN and SNLA, achieves superior video deblurring performance compared with state-of-the-art methods on various benchmarks both qualitatively and quantitatively.
\end{itemize}

%% file: sections/figs/overview.tex
\begin{figure}[t]
    \centering
    \captionsetup[subfloat]{font=tiny}
    \renewcommand{\wp}{0.165\linewidth}
    \subfloat[Input]{\includegraphics[width=\wp]{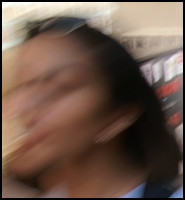}}
    \hfill
    \subfloat[IFI-RNN~\cite{Nah_2019_CVPR}]{\includegraphics[width=\wp]{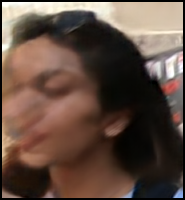}}
    \hfill
    \subfloat[STFAN~\cite{Zhou_2019_ICCV}]{\includegraphics[width=\wp]{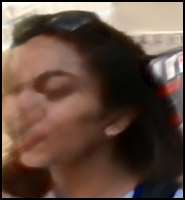}}
    \hfill
    \subfloat[ESTRNN~\cite{zhong2020efficient}]{\includegraphics[width=\wp]{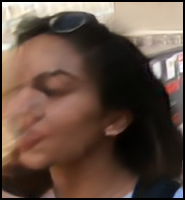}}
    \hfill
    \subfloat[\textbf{PAHS}]{\includegraphics[width=\wp]{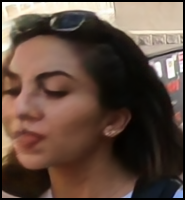}}
    \hfill
    \subfloat[GT]{\includegraphics[width=\wp]{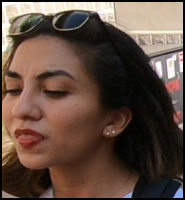}}
    \figcspace
    \caption{
        \textbf{Qualitative comparisons of different video-deblurring models on REDS~\cite{Nah_2019_CVPR_Workshops_REDS} dataset.}
        Our model manages hidden states to be updated properly with proposed PPRNN and SNLA. Our methods improve the restoration quality by large margin.
    }
    \label{fig:overview}
    \figspace
\end{figure}

%% file: sections/2_relatedworks.tex
\section{Related Works}
\label{sec:related_works}
\input{sections/figs/model_all}


\noindent
\textbf{Video Deblurring with Explicit Motion Models.}
Early works on video deblurring~\cite{cho2012registration,matsushita2006full} tried to find and aggregate sharper texture from the neighboring frames.
Energy optimization-based approaches~\cite{Kim_2015_CVPR,kim2017dynamic,li2010generating,Pan_2020_CVPR,Wulff_2014_ECCV,Zhang_2014_CVPR} estimated blur kernels from the associated patches and restored the deblurred frames through the joint optimization process.
However, spatio-temporally varying blur kernels make the optimization process to be complex and computationally expensive.

Su~\etal~\cite{Su_2017_CVPR} introduced a learning-based method by feeding stacked blurry frames into CNN layers to deblur the center frame.
They aligned neighboring frames using homography and optical flows.
However, it is challenging to acquire accurate pixel-level correspondences of frames under blur artifacts.
With focusing on the execution speed, Deng~\etal~\cite{deng2021multi} employed RCSA module with both channel-wise and spatial-wise attention to efficiently deblur the video in the real-time processing speed.
Pan~\etal~\cite{Pan_2020_CVPR} estimated sharpness as a prior, which can represent the global feature in the temporal sequences and support the learning of deblurring the target frame.
On the other hand, ARvo~\cite{li2021arvo} proposed an all-range correlation volume pyramid that acquired all-range correlation by matching pixel pairs in all spatial ranges between the target frame and neighboring frames.
In order to help finding inter-frame correspondences, \cite{li2021arvo,Pan_2020_CVPR} estimated optical flow from successive latent frames using PWC-Net~\cite{Sun_2018_CVPR}.
However, estimating the optical flow requires large computational costs, which makes the overall model heavy and slow.

\noindent
\textbf{Video Deblurring with Implicit Motion Models.}
Besides the explicit alignment to compensate for the motion blurs, recent studies digged into modeling motion information implicitly.
RDN~\cite{Wieschollek_2017_ICCV}, as a recurrent network, employed temporal skip connections in multi-scale features in the residual dense blocks.
Stepping forward, OVD~\cite{Kim_2017_ICCV} extracted hidden states with a corresponding module and further adapted the states to the next time step with dynamic temporal blending module.
IFI-RNN~\cite{Nah_2019_CVPR} more actively updated the hidden states in an iterative manner.
By reusing the recurrent architecture and a precomputed blurry input feature, intra-frame iterations injected information of the target time step.
STFAN~\cite{Zhou_2019_ICCV}, on the other hand, proposed to use filter-adaptive convolutions to align the hidden states and the features from the target frames.
ESTRNN~\cite{zhong2020efficient} used multiple features extracted at different time steps by the recurrent network structure.
The set of features were aggregated via spatio-temporal attention to produce a deblurred frame.

Among the previous works, RNN-based methods~\cite{Kim_2017_ICCV,Nah_2019_CVPR,zhong2020efficient,Zhou_2019_ICCV} could serve to encode long-term dependency by conveying temporal information with hidden states.
However, a hidden state is unaware of the information of the target frame as it is extracted from the past frames and may not be very optimal in deblurring if used as is.
To overcome such an issue, several works~\cite{Nah_2019_CVPR,Zhou_2019_ICCV} tried to adapt the hidden states to the current frame by referring to the current feature.
However, IFI-RNN~\cite{Nah_2019_CVPR} only referred current feature which can fade out the crucial information from the past.
Also, STFAN~\cite{Zhou_2019_ICCV} only filtered the hidden states locally to the target frame, which cannot handle the misalignment of hidden states in case of abrupt large motions.



\noindent
\textbf{Non-Local Attention.}
With limited resources, allocating more resources for the appropriate components can be a natural strategy.
Since Vaswani~\etal~\cite{vaswani2017attention} first proposed the transformers for machine translation, that learn to focus on the informative components, have been widely used in vision community~\cite{chen2021pre,dosovitskiy2020image}.
They viewed inputs as a set of key-value pairs and mapped the query to the key-value pairs to produce the output.
Wang~\etal~\cite{wang2018non} proposed non-local attention to obtain the spatial attention by exploring the whole region of the features.
Non-local attention~\cite{wang2018non} allowed distant pixels to contribute to enhancing features with matrix multiplication between the embedded features.
However, despite non-local attention~\cite{wang2018non} displayed sound performances, it needed a massive amount of memory and computational time, which was not compatible with large inputs.
Criss-Cross attention module~\cite{huang2019ccnet} tried to solve the above problem by capturing contextual information using horizontal and vertical directions.
Liu~\etal~\cite{mairal2009non} also fixed up the problems of non-local attention with a sparse attention module.
They chose the top $k$ most related contextual attention by sorting the information with spherical LSH~\cite{gionis1999similarity}.
With selective sparse attention, the model enjoyed robustness by preventing the model from attending of unrelated and noisy information.

In this paper, we also adopt the non-local attention mechanism and propose SNLA module to model the spatially long-range dependencies.
We apply strided convolution to obtain the correlation between the feature patches so that the required memory footprint is reduced.
In addition, we apply a filtering module in the attention mechanism so that irrelevant information is suppressed when the correlation between the whole hidden state and the frame feature is low.
It makes our SNLA attention mechanism to be selective in contrast to most methods.


%% file: sections/figs/model_all.tex
\begin{figure}[t]
    \centering
    \renewcommand{\wp}{0.45\linewidth}
    \newcommand{\hp}{\hspace{0.1\linewidth}}
    \newcommand{\hhp}{\hspace{0.05\linewidth}}
    \subfloat{\includegraphics[width=0.9\linewidth]{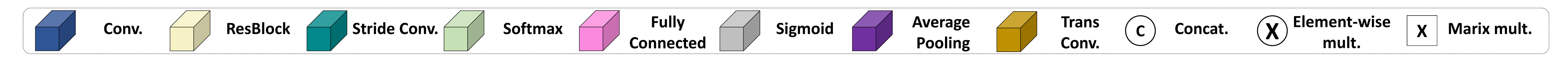}}
    \addtocounter{subfigure}{-1}
    \hspace{-3mm}
    \\
    \subfloat[Overall architecture of PAHS~(Ours) \label{fig:model_all_all}]{
    \includegraphics[width=0.98\linewidth]{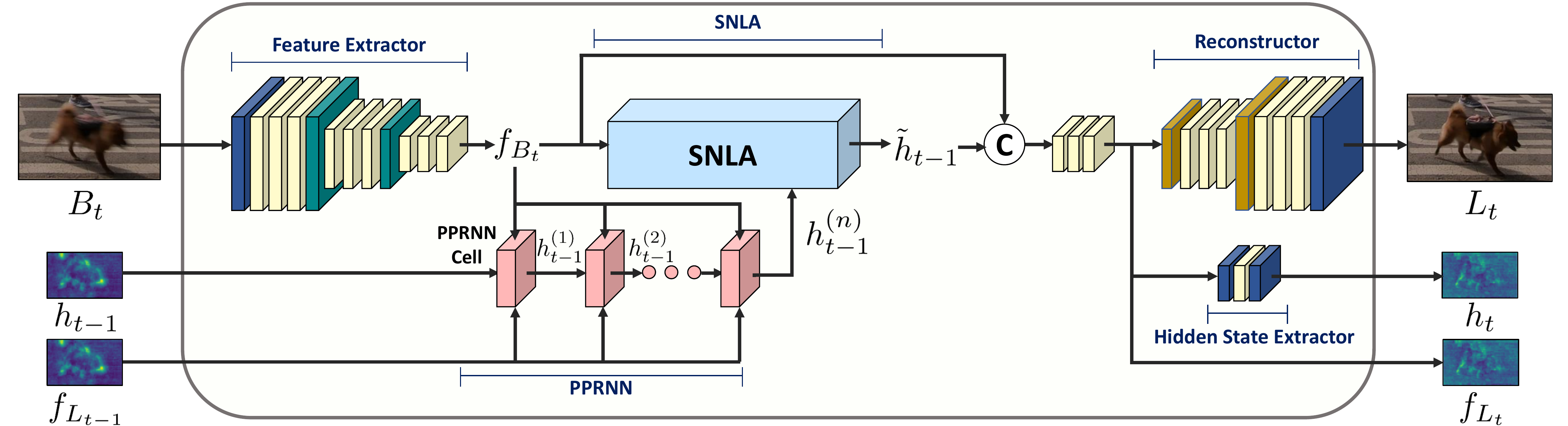}
    }
    \\
    \subfloat[PPRNN cell structure \label{fig:pprnn}]{
    \includegraphics[height=0.15\linewidth]{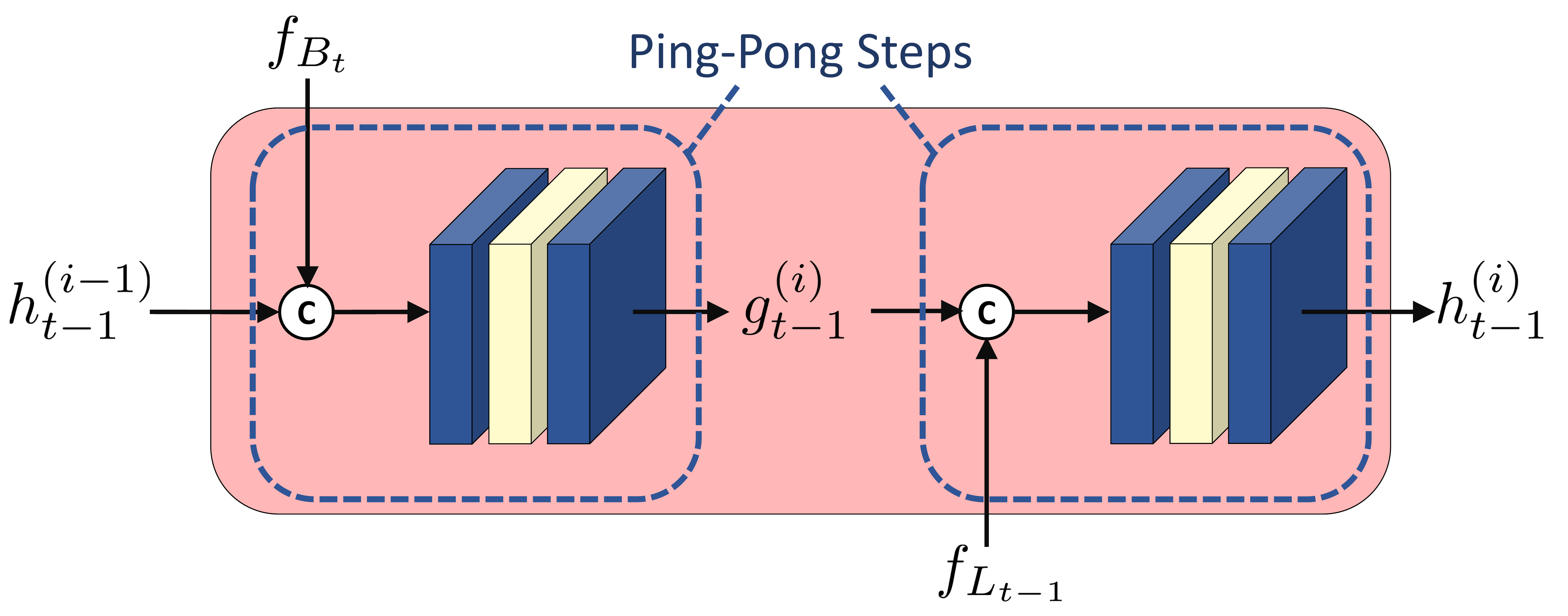}
    }
    \hhp
    \subfloat[SNLA module structure \label{fig:snla}]{
    \includegraphics[height=0.15\linewidth]{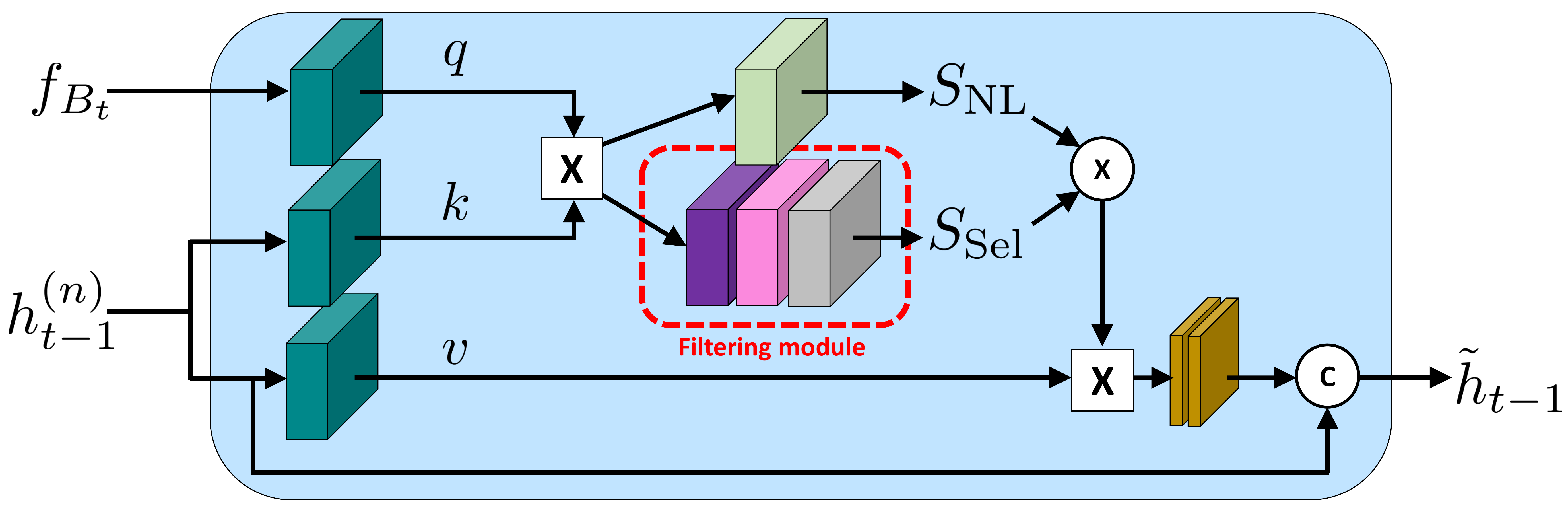}
    }
    \figcspace
    \caption{\textbf{Overall architecture and components of our PAHS.}}
    \label{fig:model}
    \figspace
\end{figure}

%% file: sections/3_proposedmethods.tex
\section{Proposed Method}

Our main goal is to improve the deblurring performance of RNNs by proposing the ways to better utilize hidden states.
We start building our model from a basic RNN architecture with feature extractor, hidden state extractor, and the latent image reconstructor.
With the blurry input frame $B_{t}$, deblurred latent frame $L_{t}$, intermediate feature $f_{L_{t}}$ used to reconstruct $L_{t}$, and the hidden state $h_{t}$, the overall process at time step $t$ could be written as:
\begin{equation*}
   (L_t, h_t, f_{L_t}) = \mathcal{M}_{\text{R}}(B_{t}, h_{t-1}, f_{L_{t-1}}),
\end{equation*}
where $\mathcal{M}_{\text{R}}$ denotes the recurrent neural network architecture.
We propose 2 elements, Ping-Pong RNN and Selective Non-Local Attention module to process hidden states before the reconstruction stage.
As shown in Figure~\ref{fig:model_all_all}, maintaining the recurrent structure, we allocate PPRNN and SNLA modules where the hidden state from the previous time step is received.

\subsection{Ping-Pong Recurrent Neural Network}
Analogous to the RNN-based video deblurring literature modifying the hidden states before its use~\cite{Kim_2017_ICCV,Nah_2019_CVPR,Zhou_2019_ICCV}, we also propose to update the hidden states but in a novel way.
Different from the previous methods only using the current frame information in the update step, such an approach may lose information that could have been important at the final reconstruction stage.
Especially at the moments with abrupt motions, large displacement of features may cause the misaligned information to be lost by the ineffective handling of spatially long-ranged information.
Thus, in order to protect the necessary information from being lost, we propose to update the hidden states in a balanced way by our Ping-Pong update strategy.

\fix{Different from typical RNN structures~\cite{cho2014learning_gru,hochreiter1997long} that update the hidden state only once in a single time step, PPRNN uses both the past and the present frame information alternately in our ping-pong steps.
In addition, unlike the analogous to the RNNs whose purpose is to create output by flowing the hidden states, PPRNN exists only to update the hidden states, solely generating hidden states for the output.}
As shown in Figure~\ref{fig:pprnn}, each Ping-Pong step updates the hidden state by using either the blurry frame feature $f_{B_{t}}$ or $f_{L_{t-1}}$ with alternation.
The PPRNN operation at $i$-th recurrence step is written as:
\begin{align*}
    h_{t-1}^{(0)} &= h_{t-1},\\
    g_{t-1}^{(i)} &= \mathcal{M}_{\text{P}}(f_{B_t}, h_{t-1}^{(i-1)}),\\
    h_{t-1}^{(i)} &= \mathcal{M}_{\text{P}}(f_{L_{t-1}}, g_{t-1}^{(i)}),
\end{align*}
where $\mathcal{M}_{\text{P}}$ denotes a Ping-Pong block in the PPRNN cell.
Each output of the Ping-Pong block, $g_{t-1}^{(i)}$ and $h_{t-1}^{(i)}$ is generated by looking into the feature of the current frame~$f_{B_t}$ and the past deblurred feature~$f_{L_{t-1}}$, respectively.
Thus, the temporal feature relations could be aggregated by the successive recurrent steps.
We design the PPRNN cell with two PPRNN blocks with shared parameters as $\mathcal{M}_{\text{P}}$.
A single block is built with only two convolution layers and a residual block in between to lower the computational burden of PPRNN.
As a result, PPRNN only involves a modest increase in the number of parameters and the computational cost.
Due to the recurrent architecture and the small cost, arbitrary number of recurrence can be chosen without significant speed reduction.
In our experiments, we choose the maximum recurrence number $n=4$.
For the detailed architecture specifics of PPRNN, please refer to the supplementary material.

\subsection{Selective Non-Local Attention}
Although the updated hidden state $h^{(n)}_{t-1}$ obtained from the PPRNN contains potentially rich information from the past and the present, such information could be located without being aligned to each other.
In order for $h^{(n)}_{t-1}$ to be better utilized at deblurring the given frame, the hidden state should be further refined by rearranging the stacked information by relocation.
Thus, we present SNLA module as shown in Figure~\ref{fig:snla}.

In our non-local attention scheme, we take the features from $f_{B_{t}}$ as the query and the features from $h^{(n)}_{t-1}$ as key and value tokens.
The query, key and value tokens are each obtained by strided convolutions to reduce the memory consumption in attention map generation.
\begin{equation}
    (q,k,v) = (Q(f_{B_t}),K(h^{(n)}_{t-1}),V(h^{(n)}_{t-1})),
    \label{eq:nonlocal1}
\end{equation}
where $q$, $k$, and $v$ denote the query, key, and value obtained from three strided convolutions, 
$Q(\cdot), K(\cdot)$, and $V(\cdot)$, respectively.
Thus, the acquired attention map $S_{\text{NL}}$ encodes the spatial correspondence between $f_{B_{t}}$ and $h^{(n)}_{t-1}$.
\begin{equation}
    S_{\text{NL}} = \text{softmax}(qk^{T}).
    \label{eq:nonlocal2}
\end{equation}
Different from typical self-attention scheme~\cite{dosovitskiy2020image,vaswani2017attention}, our attention map indicates the important pixels in $h^{(n)}_{t-1}$ that are located from $f_{B_{t}}$.
Henceforth, our non-local attention could internally relocate the information in the hidden state for a more optimal state to help deblurring with $f_{B_{t}}$.
\input{sections/tables/overallGR}

\input{sections/figs/dvd}
\input{sections/figs/gopro_reds}

While the relative spatial relevance is obtained as correlation score map $S_{\text{NL}}$, there could be cases where $f_{B_{t}}$ and $h^{(n)}_{t-1}$ are less correlated.
In those cases, irrelevant information could be overemphasized, leading to suboptimal feature aggregation.
Thus, we put a filtering module that allows attenuating the attention map by $S_{\text{Sel}}$ as:
\begin{align}
    \centering
    S_{\text{Sel}} &= \text{sigmoid}(\text{FC}(\text{pool}(q{k}^{T}))),
    \label{eq:nonlocal3}
\end{align}
\begin{align*}
    \text{Att}(q,k,v) &= D(S_{\text{Sel}} \otimes S_{\text{NL}} \times v),\\
    {\tilde{h}}_{t-1} &= h^{(n)}_{t-1} + \text{Att}(q,k,v),
\end{align*}
where \text{FC}, \text{pool}, $\times$, $\otimes$ and $D$ stand for fully-connected layer, average pooling, matrix multiplication, element-wise multiplication, and transposed convolution, respectively.
\text{Att} is the obtained attention output and ${\tilde{h}}_{t-1}$ is the updated final hidden state.
${\tilde{h}}_{t-1}$ is concatenated with $f_{B_{t}}$ to be fed into the reconstructor.

Similarly to designing PPRNN, we keep the computational burden of SNLA low by setting the stride in $Q(\cdot), K(\cdot)$, and $V(\cdot)$ as 4 to reduce the size of the query, key, and value tokens.

\subsection{Bidirectional PAHS structure}
Although PPRNN and SNLA effectively transfer the necessary information from the past into the required location, uni-directional RNN cannot use the information from the future frames.
In order to further improve the performance, we apply bidirectional RNN architecture.
As shown in Figure~\ref{fig:bidirection}, PAHS cell in each forward and backward path generate the two intermediate feature~($f^{f}_{L_t}$ and $f^{b}_{L_t}$), respectively.
We concatenate the two intermediate features and feed them into the reconstructor module.
Following the overall structure of reconstructor in unidirectional inference as shown in Figure~\ref{fig:model}, we modify the first transposed convolution layer's number of input channels by the double.

%% file: sections/tables/overallGR.tex
\begin{table}[t]
    \footnotesize
    \resizebox{1.0\linewidth}{!}{
        \begin{tabular}{cc|cccccccc}
            \toprule
            \multicolumn{2}{c|}{Methods} & EDVR~\cite{wang2019edvr} & OVD~\cite{Kim_2017_ICCV} & IFI-RNN~\cite{Nah_2019_CVPR} & STFAN~\cite{Zhou_2019_ICCV}  &  
            ESTRNN~\cite{zhong2020efficient} & TSP~\cite{Pan_2020_CVPR} &
            ARVo~\cite{li2021arvo} &
            \textbf{PAHS}\\
            \midrule
            \multirow{2}{*}{DVD} & PSNR & 28.51 & 29.15 &  30.53 & 31.15 & 31.92 & 32.13  & 32.80 & \textbf{33.28}\\
            & SSIM & 0.8637 & 0.8728 &  0.9069 & 0.9049 & 0.9298 & 0.9268 & 0.9352 & \textbf{0.9565}\\
            \midrule
            \multirow{2}{*}{GOPRO} & PSNR & 30.20 & 28.92 &  28.90 & 28.98 & 31.10 & 31.67 & 32.16 & \textbf{33.82}\\
            & SSIM & 0.9109 & 0.8660 &  0.8713 & 0.8869 &  0.9064 & 0.9279 & 0.9344 & \textbf{0.9612}\\
            \midrule
            \multirow{2}{*}{REDS} & PSNR &  32.02 & 30.46 & 30.16 & 31.44 & 32.59 & 32.08 & 32.96 &  \textbf{34.27}\\
            & SSIM &  0.9053 & 0.8808 & 0.8764 & 0.8921 &  0.9233 & 0.9205 & 0.9322 & \textbf{0.9479}\\
            \midrule
            Model Size & (MB) & 20.6 & \textbf{1.4}  & 3.6 & 5.4 & 22.4 & 16.1 & 20.6 &  21.0 \\
            Speed  & (fps) & 2.70 & 7.69 & \textbf{20.15} & 6.66 & 6.21 & 0.22 & 0.15 &  1.43\\
            \bottomrule
        \end{tabular}
    }
    \tabcspace
    \caption{\textbf{Deblurring accuracy comparison on the DVD~\cite{Su_2017_CVPR}, GORPO~\cite{Nah_2017_CVPR}, and REDS~\cite{Nah_2019_CVPR_Workshops_REDS} datasets.}
    }
    \label{tab:GOPRO_REDS}
    \tabspace
\end{table}

%% file: sections/figs/dvd.tex
\begin{figure}[t]
    \captionsetup[subfloat]{font=scriptsize}
    \renewcommand{\wp}{0.195\linewidth}
    \subfloat[Input \label{fig:dvd_input}]{\includegraphics[width=\wp]{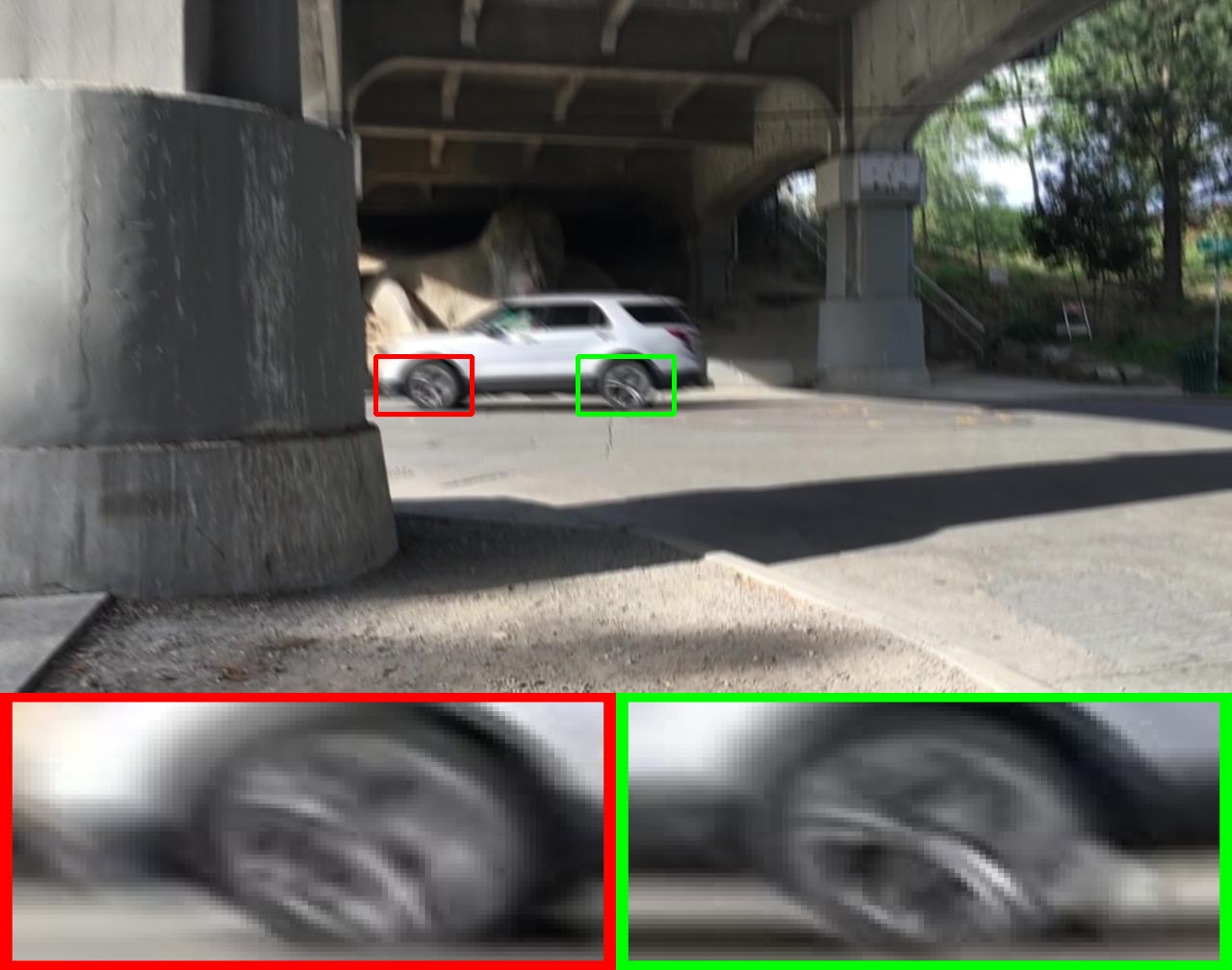}}
    \hfill
    \subfloat[TSP~\cite{Pan_2020_CVPR} \label{fig:dvd_tsp}]{\includegraphics[width=\wp]{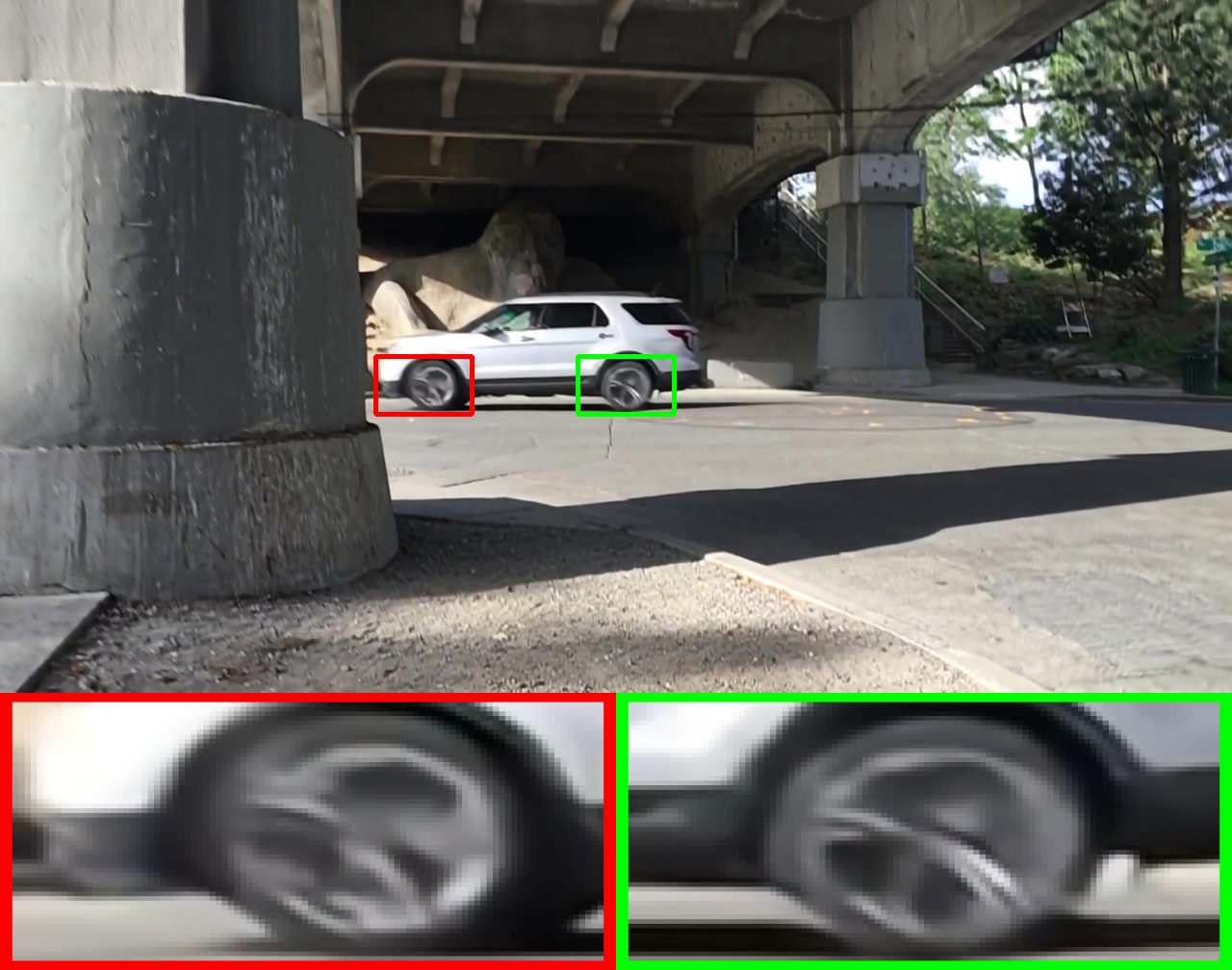}}
    \hfill
    \subfloat[ARVo~\cite{li2021arvo} \label{fig:dvd_arvo}]{\includegraphics[width=\wp]{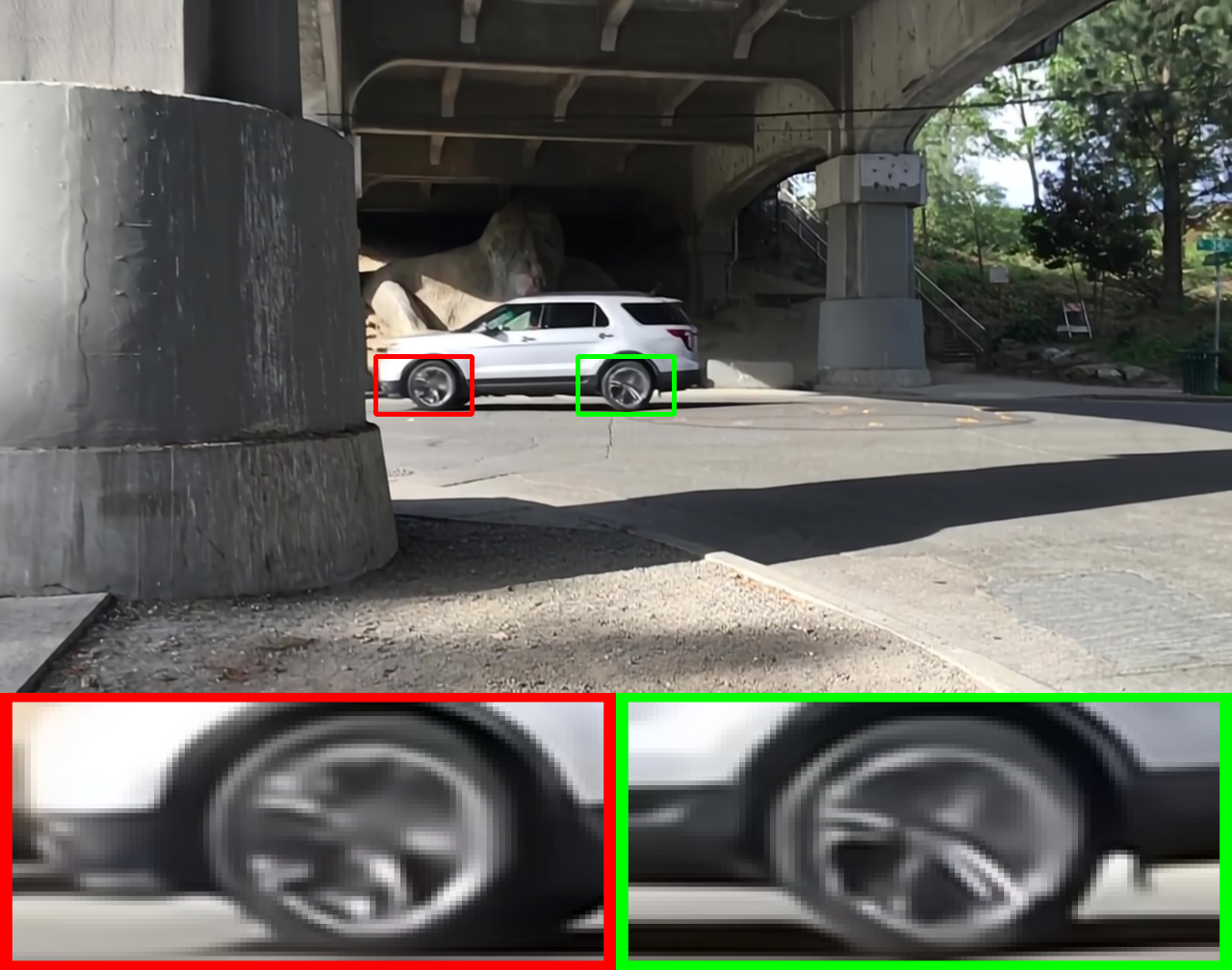}}
    \hfill
    \subfloat[\textbf{PAHS~(Ours) \label{fig:dvd_pahs}}]{\includegraphics[width=\wp]{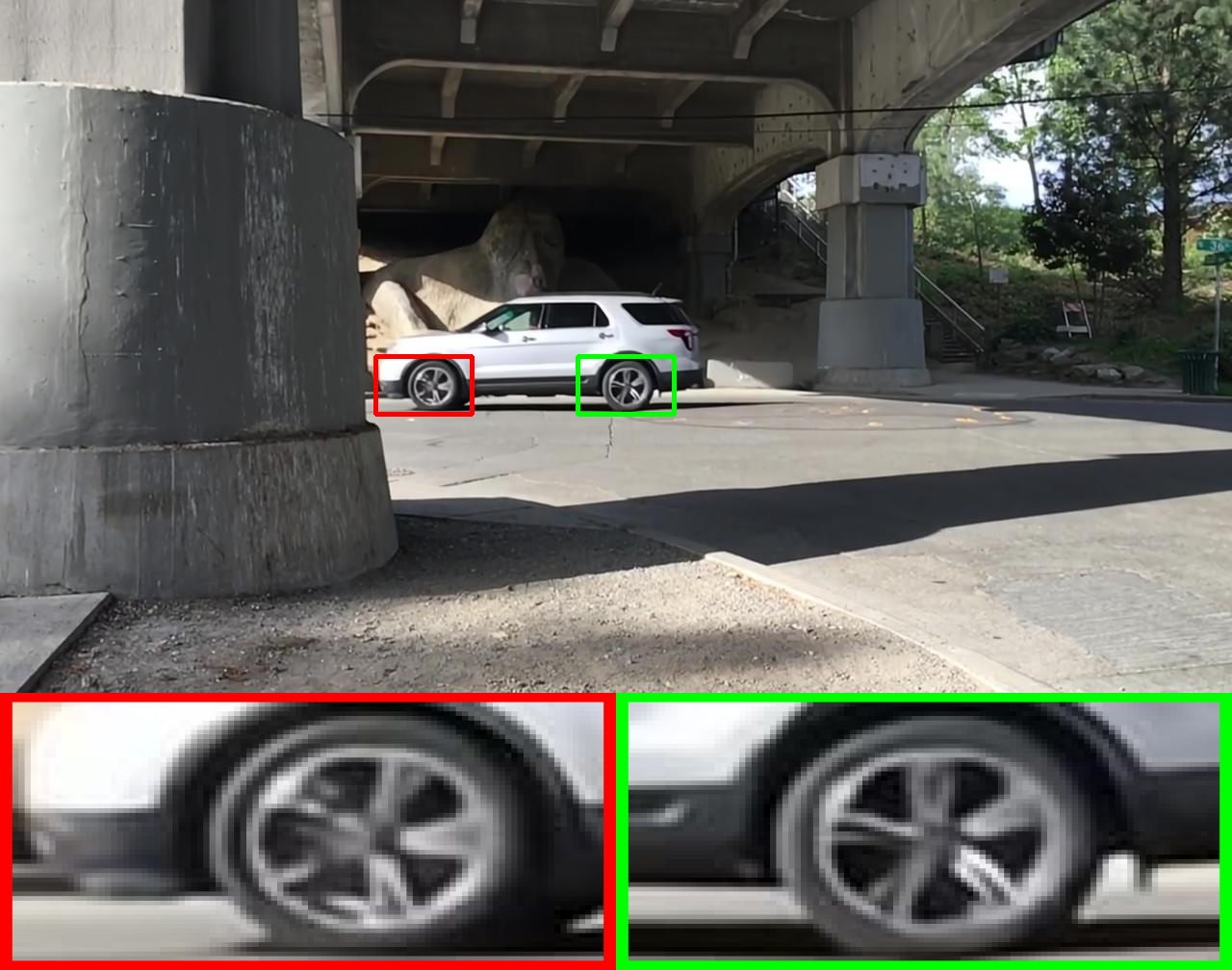}}
    \hfill
    \subfloat[Ground Truth \label{fig:dvd_gt}]{\includegraphics[width=\wp]{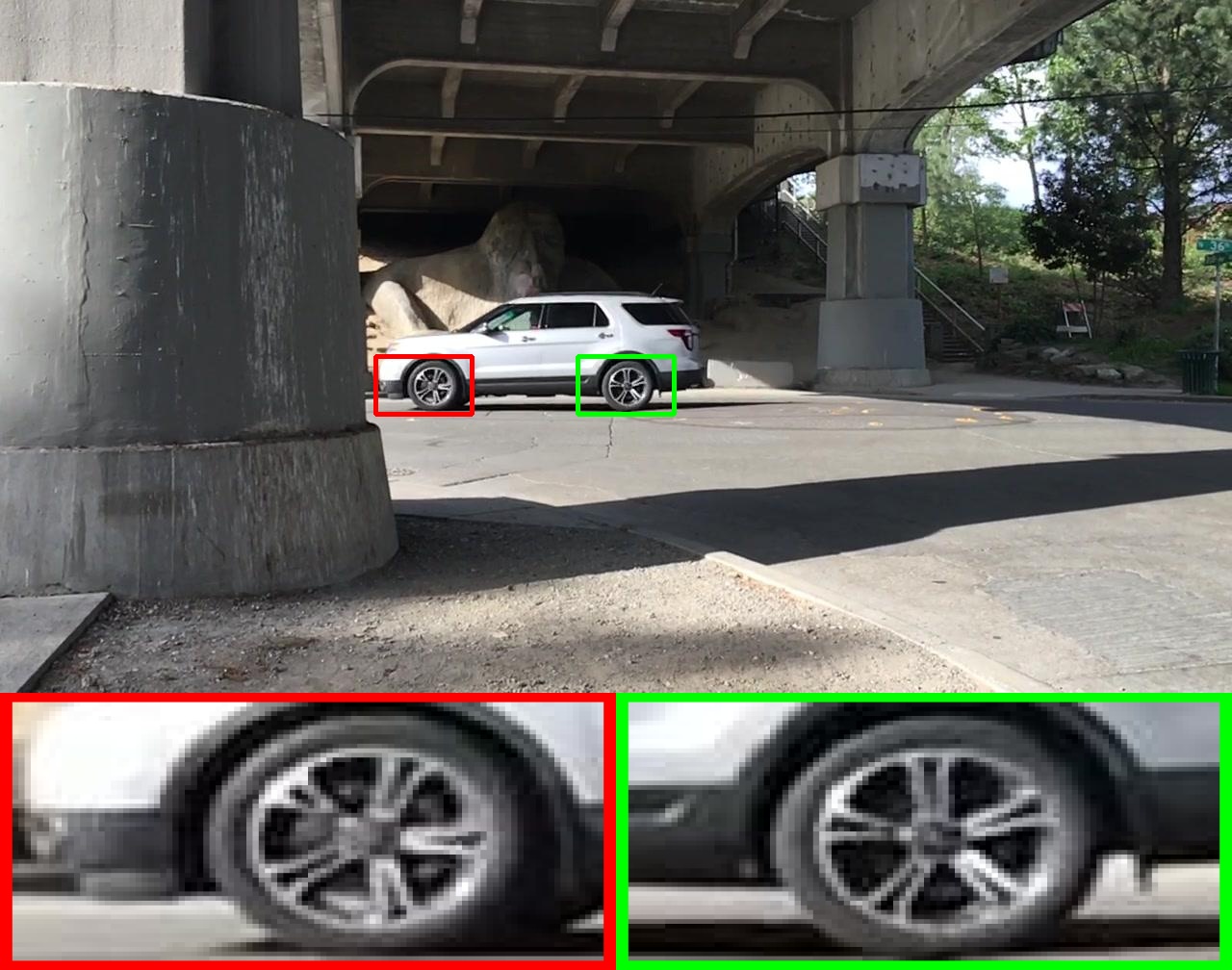}}
    \vspace{-2.8mm}
    \caption{
    \textbf{Qualitative comparison with state-of-the-art methods on DVD~\cite{Su_2017_CVPR} dataset.}
    }
    \label{fig:dvd}
    \figspace
\end{figure}

%% file: sections/figs/gopro_reds.tex
\begin{figure}[t]
    \centering
    \captionsetup[subfloat]{font=scriptsize}
    \renewcommand{\wp}{0.19\linewidth}
    \includegraphics[width=\wp]{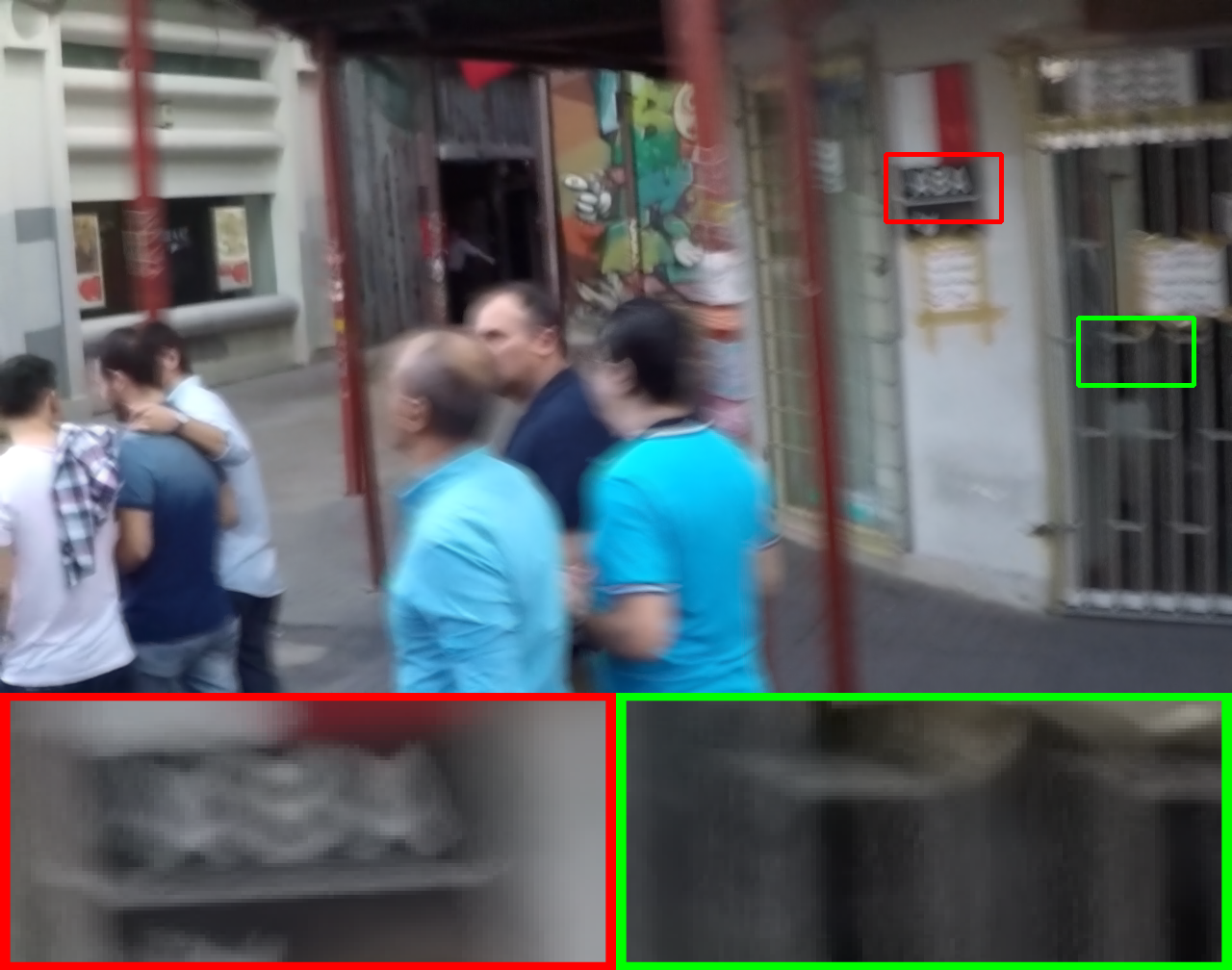}
    \hfill
    \includegraphics[width=\wp]{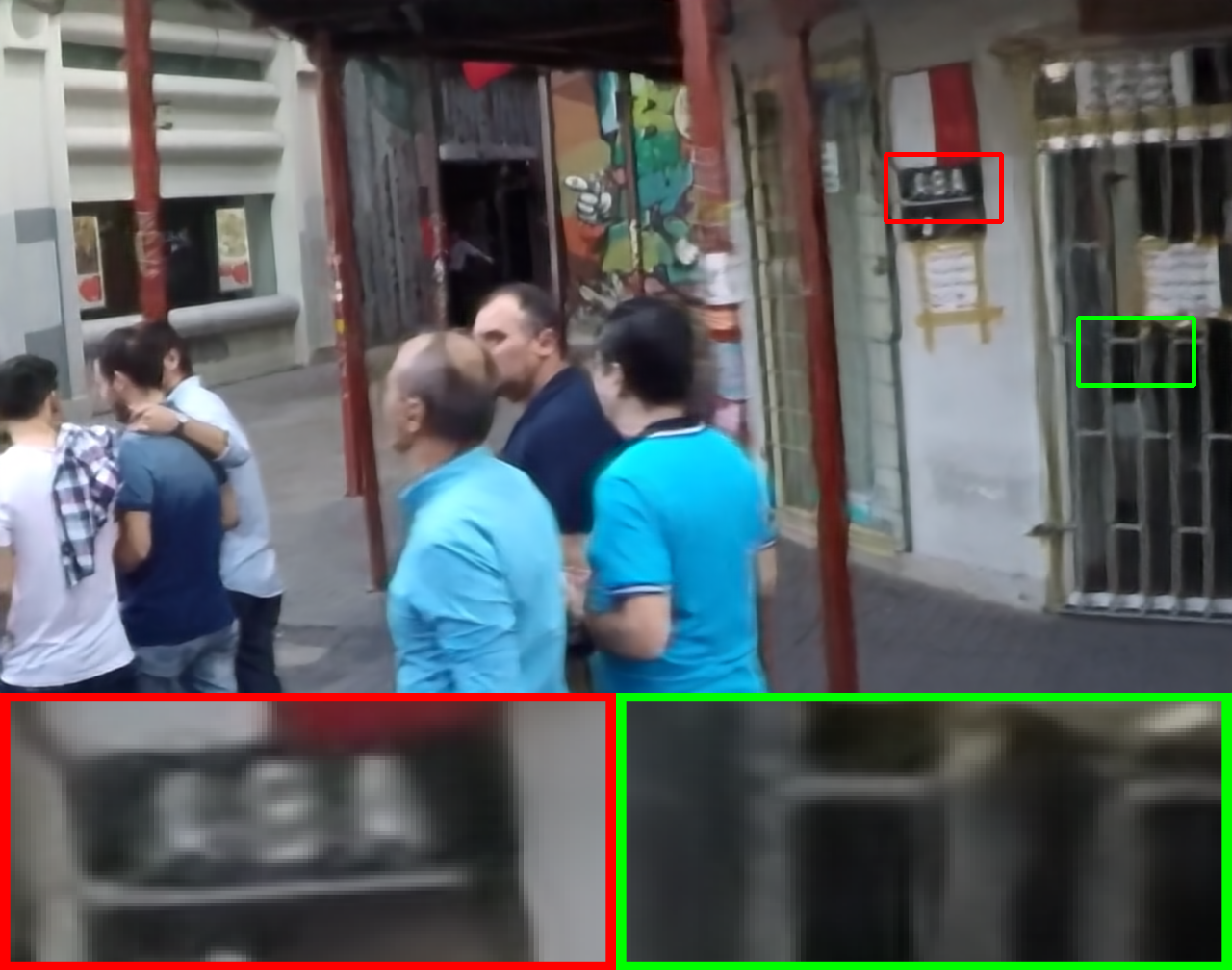}
    \hfill
    \includegraphics[width=\wp]{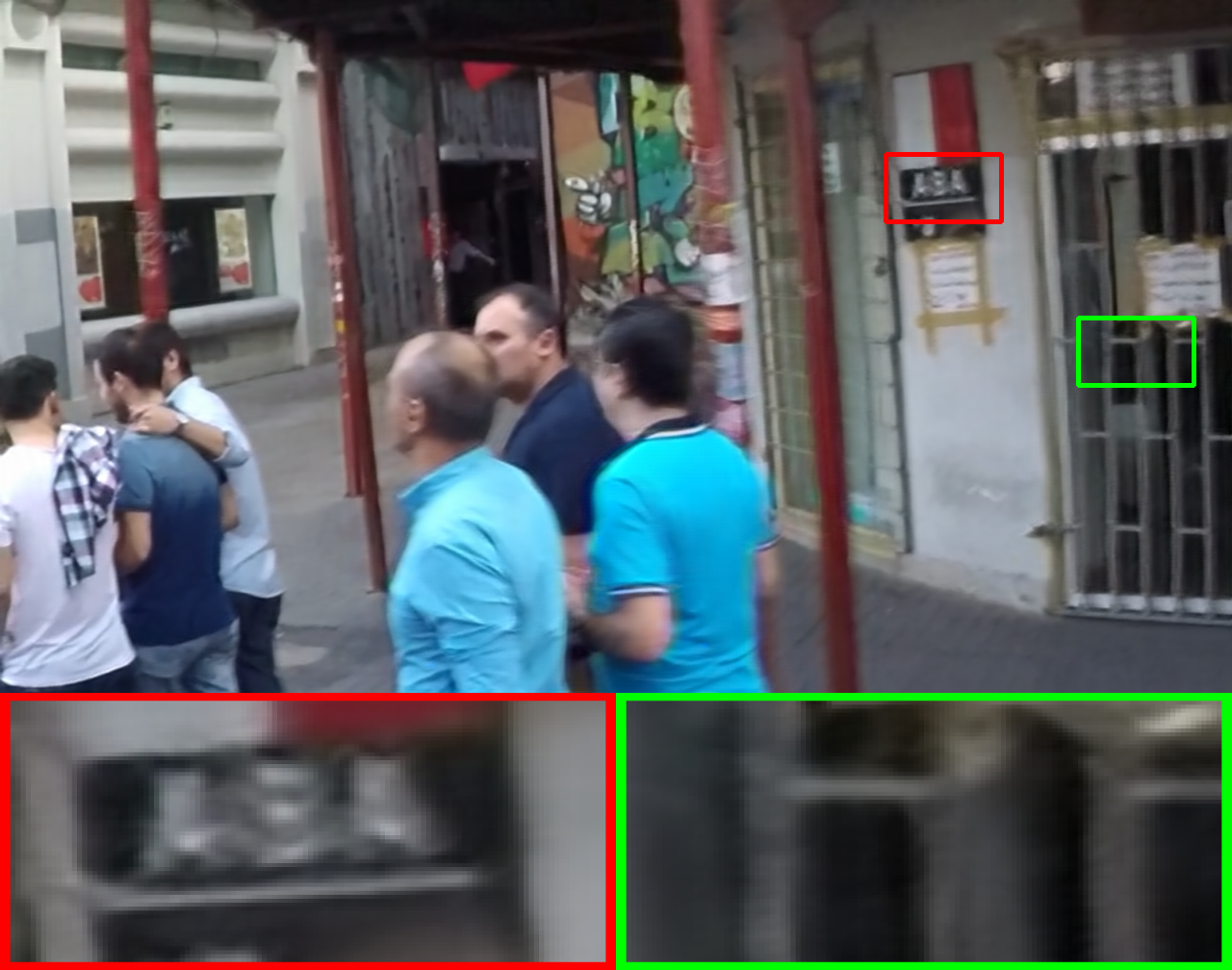}
    \hfill
    \includegraphics[width=\wp]{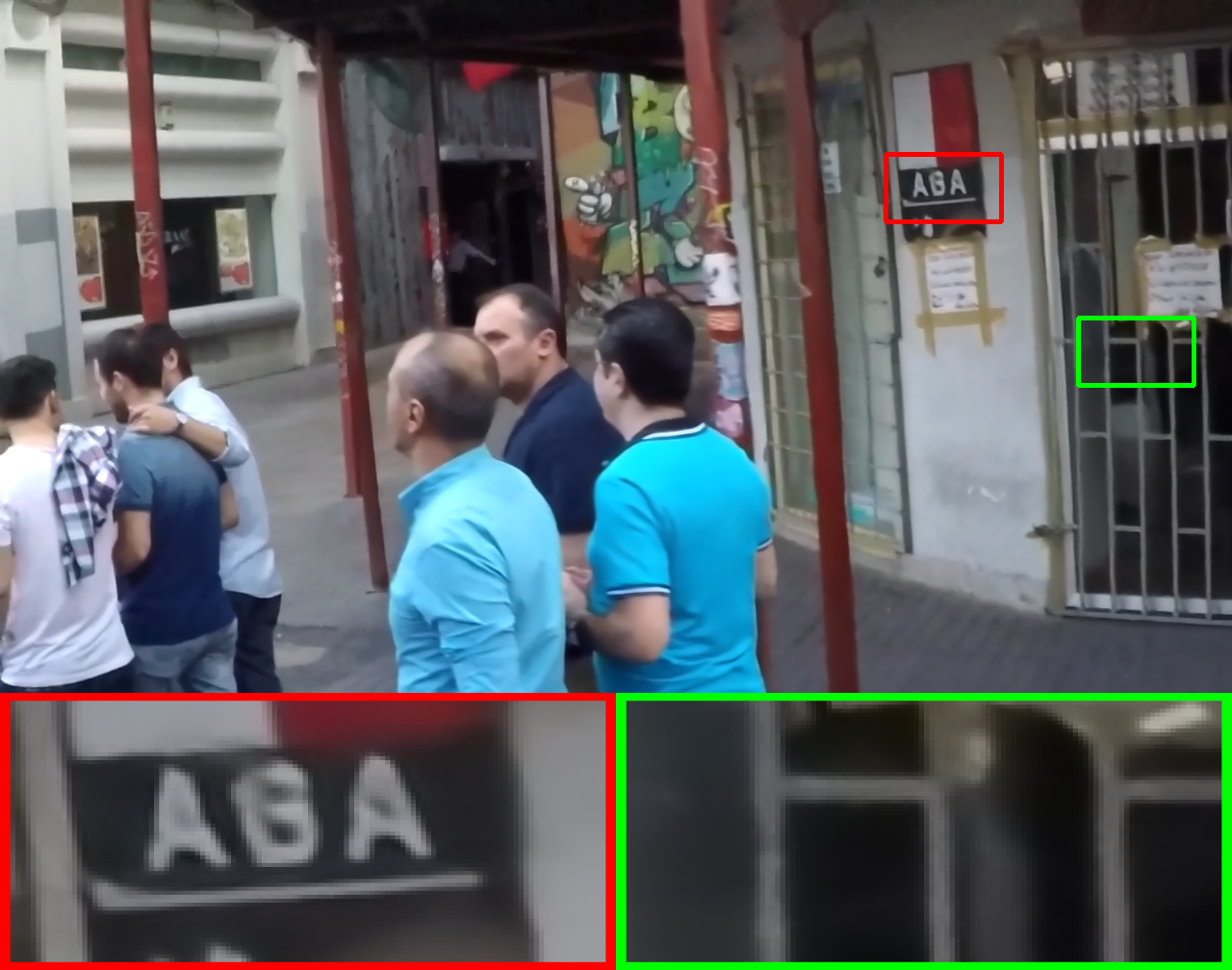} 
    \hfill
    \includegraphics[width=\wp]{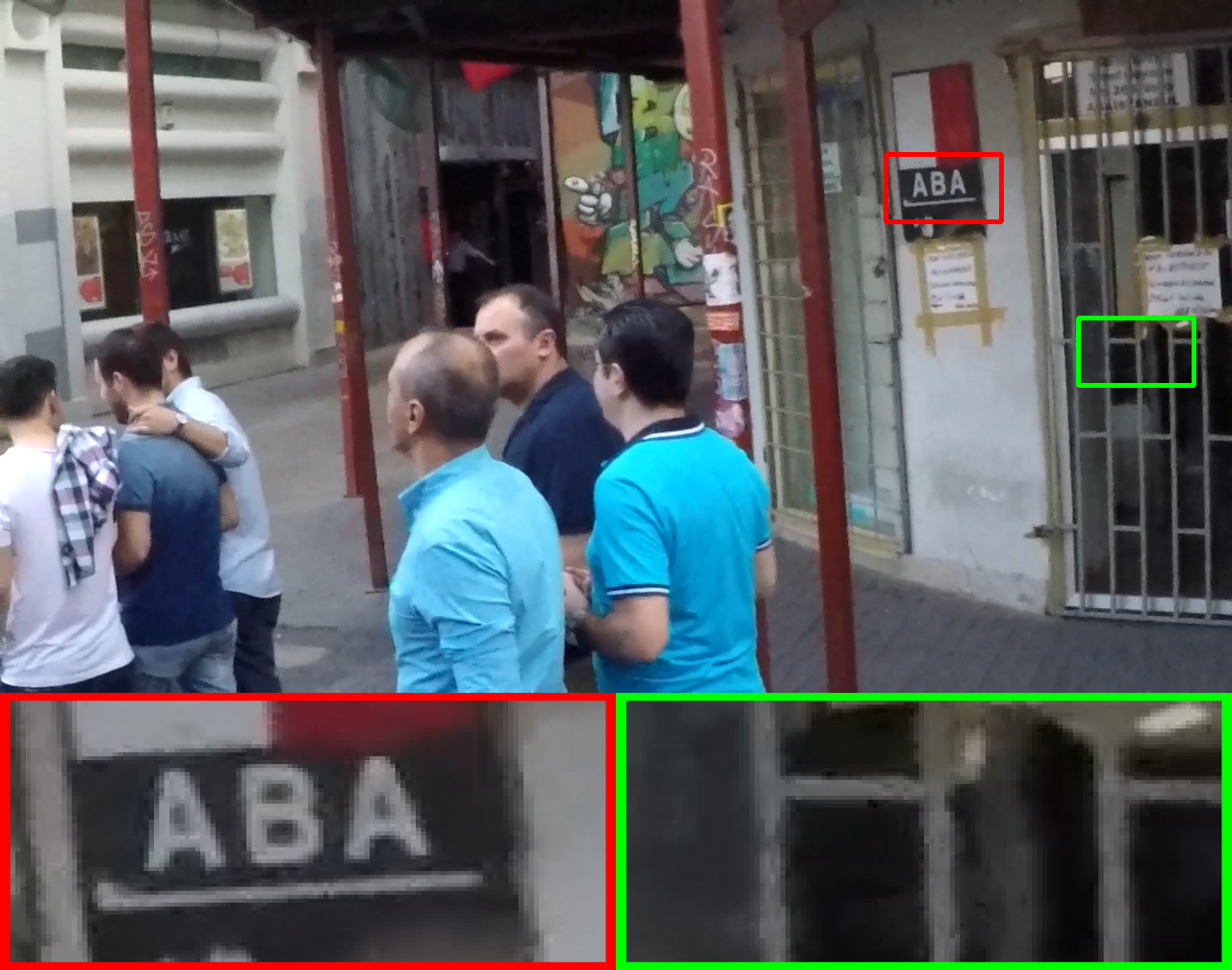}
    \\
    \centering
    \subfloat[Input \label{gr_input}]{\includegraphics[width=\wp]{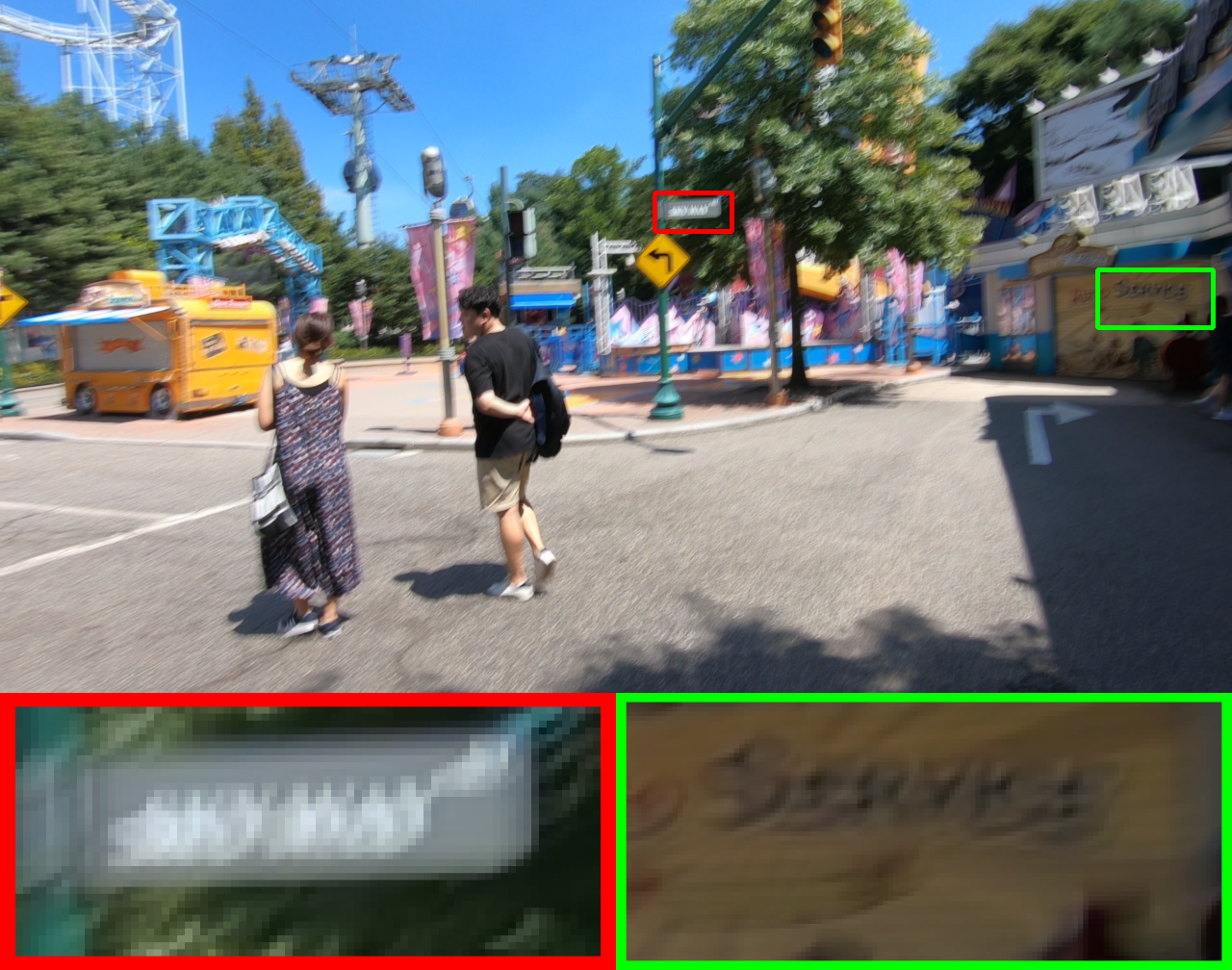}}
    \hfill
    \subfloat[IFI-RNN~\cite{Nah_2019_CVPR} \label{gr_ifirnn}]{\includegraphics[width=\wp]{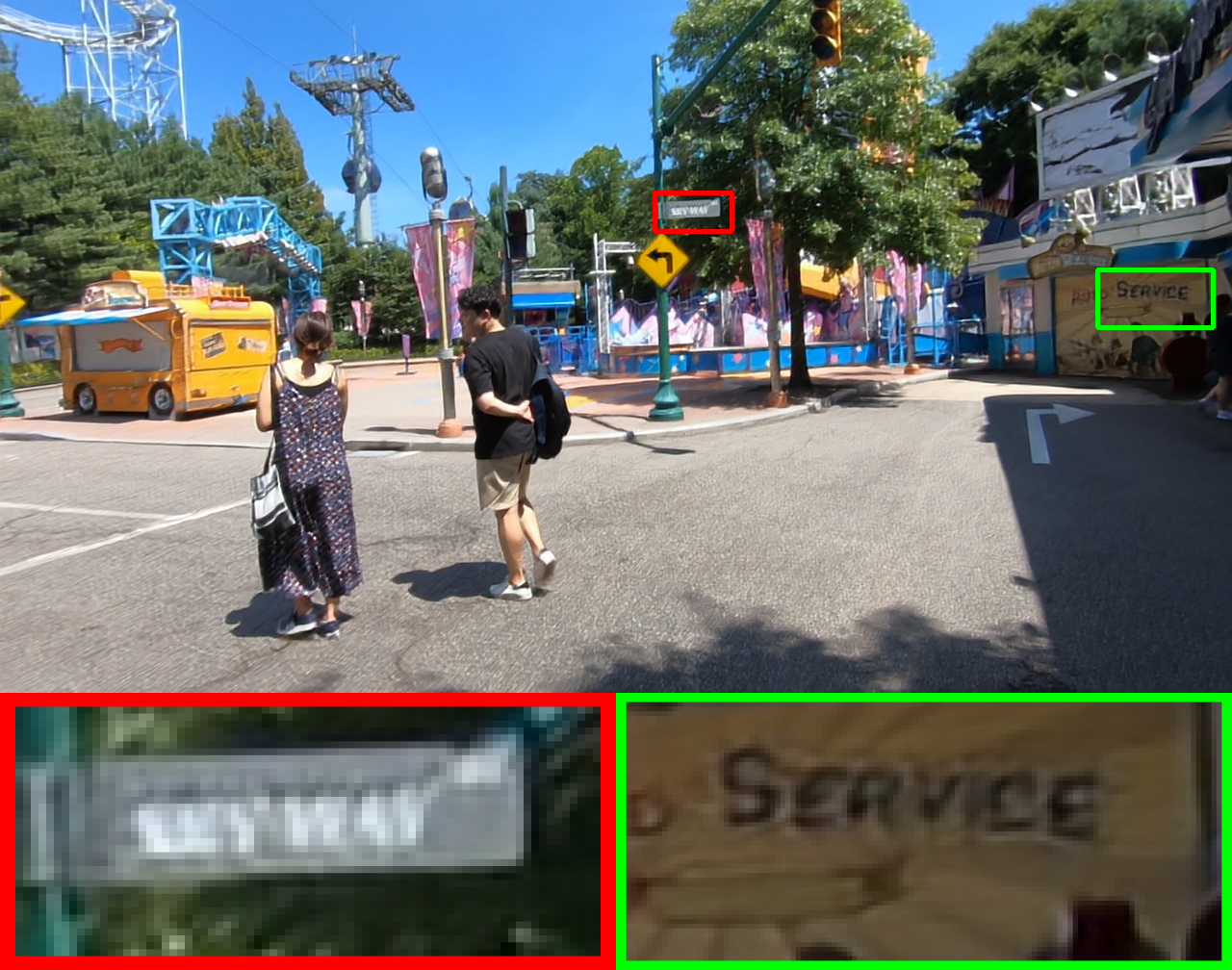}}
    \hfill
    \subfloat[ESTRNN~\cite{zhong2020efficient} \label{gr_estrnn}]{\includegraphics[width=\wp]{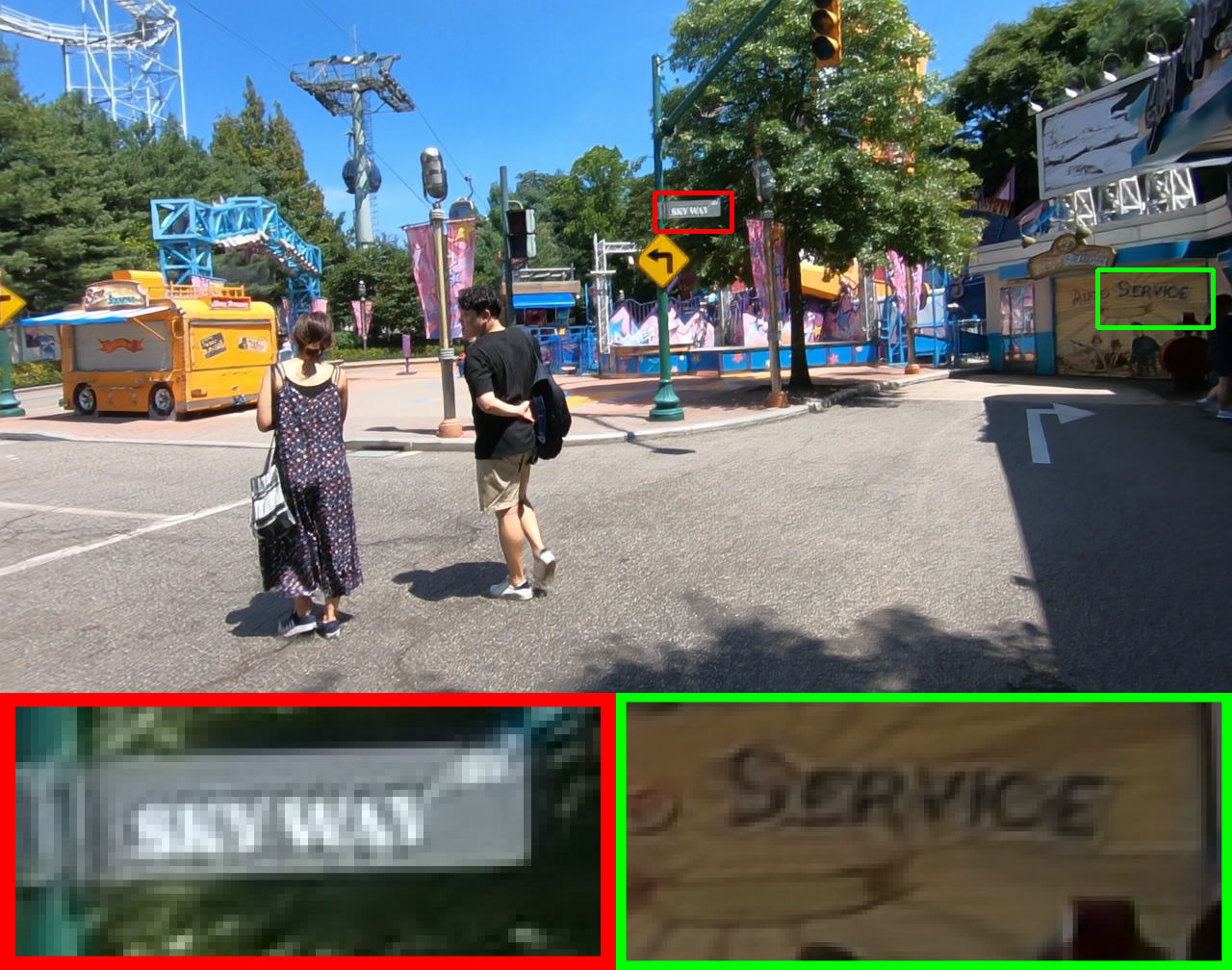}}
    \hfill
    \subfloat[\textbf{PAHS (Ours) \label{gr_pahs}}]{\includegraphics[width=\wp]{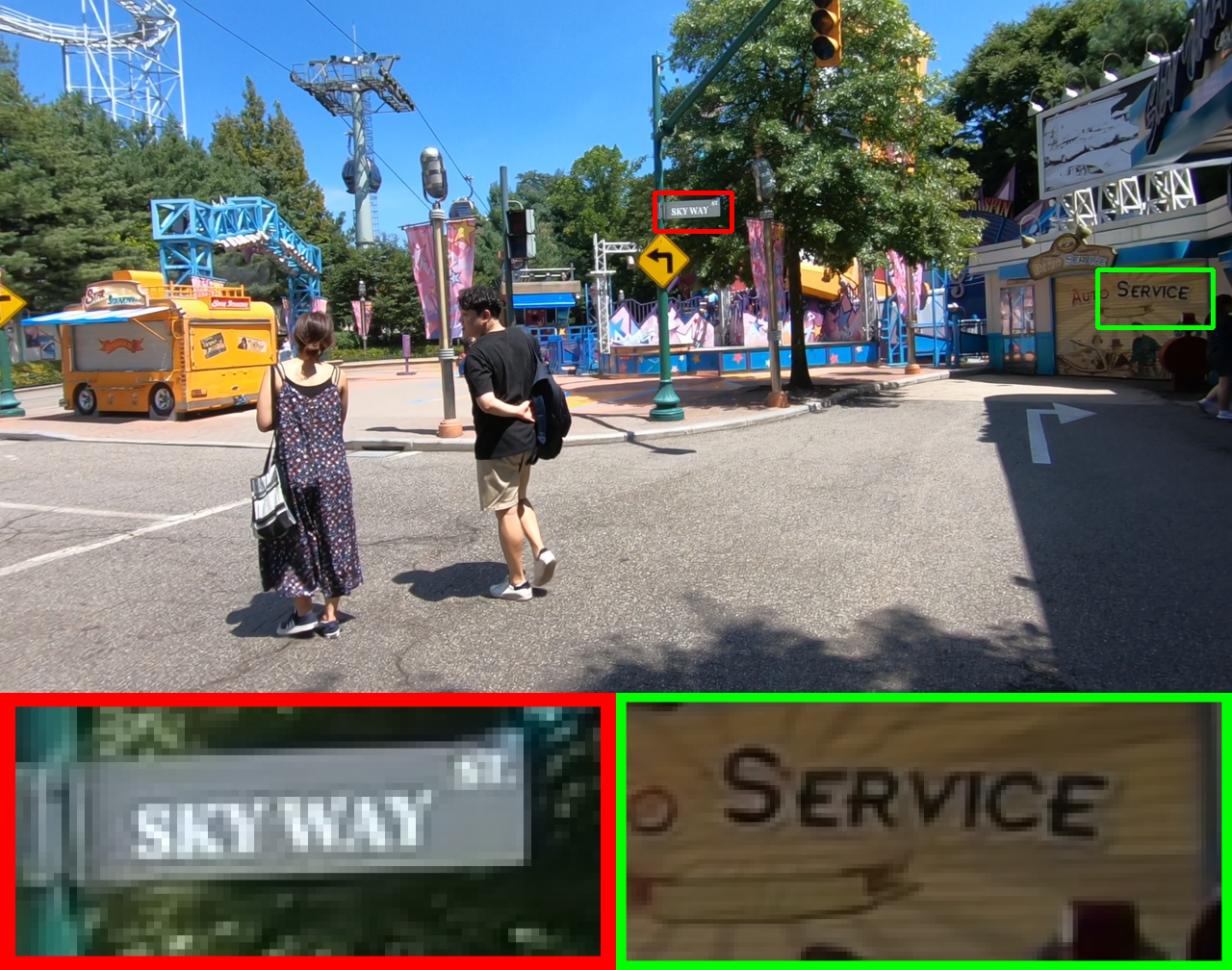}}
    \hfill
    \subfloat[Ground Truth \label{gr_gt}]{\includegraphics[width=\wp]{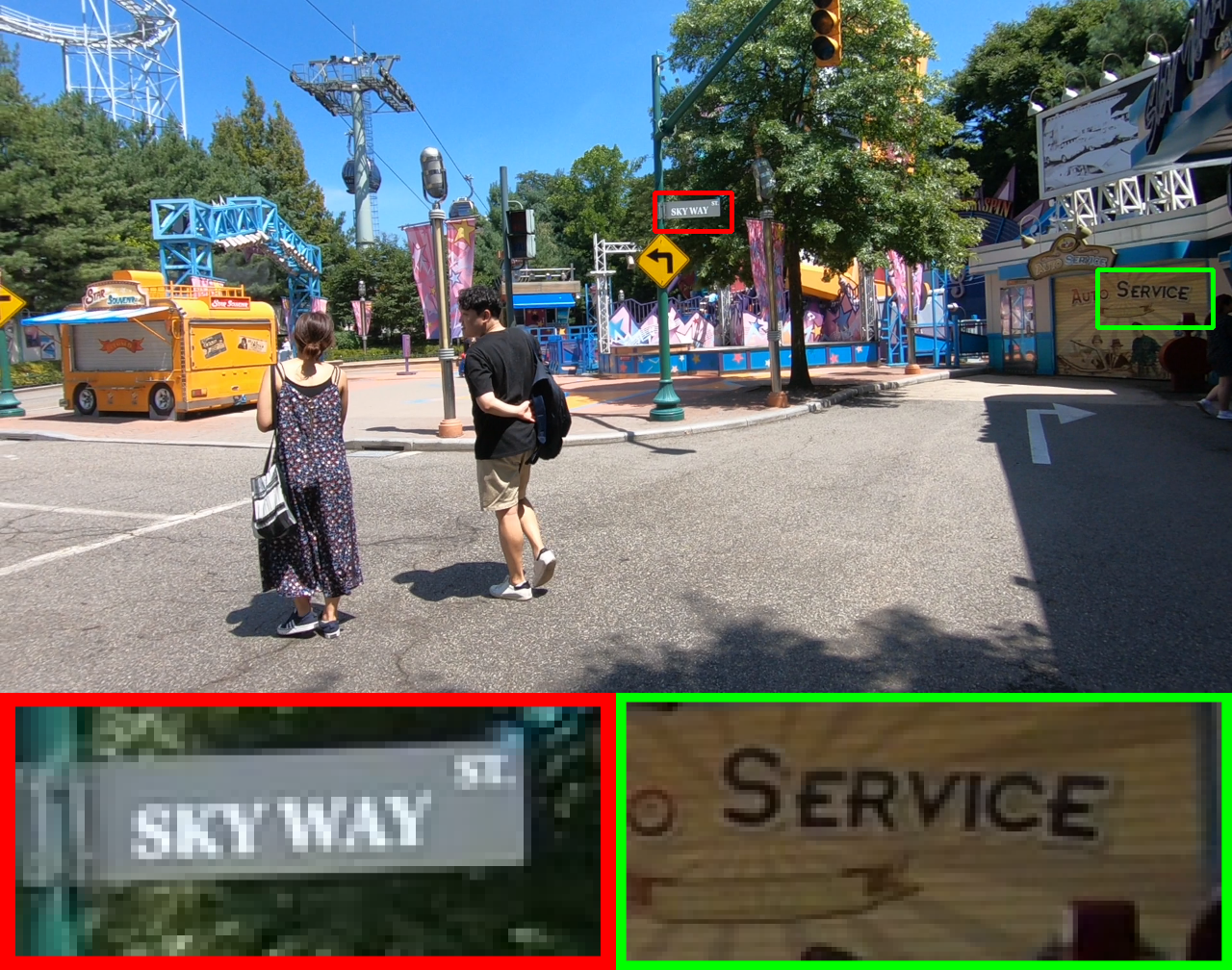}}
    \vspace{-2.8mm}
    \caption{
        \textbf{Qualitative comparison with state-of-the-art methods on GOPRO~\cite{Nah_2017_CVPR} (Top) and REDS~\cite{Nah_2019_CVPR_Workshops_REDS} (Bottom) datasets.}
        }
    \label{fig:gopro_reds}
    \figspace
\end{figure}

%% file: sections/4_experimentalresults.tex
\section{Experiments}

\subsection{Implementation Details}
All implementations were done with PyTorch~\cite{paszke2019pytorch} 1.8 and Adam optimizer~\cite{kingma2014adam} with $\beta_{1}=0.9, \beta_{2}=0.999, \epsilon=10^{-8}$, and batch size 4.
We used sRGB patches of size $256 \times 256$ and the initial learning rate was set to $1\times10^{-4}$ for training.
We train and validate the performance of our model on the widely-used DVD~\cite{Su_2017_CVPR}, GOPRO~\cite{Nah_2017_CVPR}, and REDS~\cite{Nah_2019_CVPR} datasets.
For all datasets, we train our models with simple L1 loss function by comparing the output $L_t$ with the ground-truth sharp image $S_t$ as $\left\|{L_t - S_t}\right\|$.

\subsection{Datasets and Evaluation Metrics}
\noindent
\textbf{DVD} dataset contains 5,708 training samples from 61 sequences and 1,000 evaluation samples from 10 test sequences.
All the videos are taken at 240fps with mobile phones and DSLR. 
Our model was trained for 500 epochs, and the learning rate is used to update weight by half every 200th epochs.

\input{sections/figs/bidirection}

\noindent
\textbf{GOPRO} dataset consists of 2,103 training samples from 22 sequences and 1,111 evaluation samples from 11 sequences.
The blur-sharp pairs are generated from 240 fps videos in $1280 \times 720$ resolution.
For the GOPRO dataset, we trained our model for 1,000 epochs with annealing the learning rate at every 200 epochs from the initial learning rate.

\noindent
\textbf{REDS} dataset provides 240 training sequences and 30 validation sequences.
Each sequence has a 100 frame length, and all the frames are in $1280 \times 720$ resolution.
We trained our model for additional 200 epochs from the model pre-trained on GOPRO dataset, annealing learning rate at 100th, 150th, 180th and evaluated on validation sequences.

\noindent
\textbf{Metrics.} 
We adopt peak signal-to-noise ratio~(PSNR) and structural similarity~(SSIM)~\cite{wang2004image} for evaluating the model's performances on three public datasets, DVD, GOPRO, and REDS.


\subsection{Ablation Study}
\input{sections/figs/graph-pprnnnum}
\noindent
\textbf{Effects of PPRNN.}
To validate the effect of our proposed PPRNN, we compare the deblurring quality from the models with varying number of PPRNN recurrences in Figure~\ref{fig:pprnn-graph}.
Recurrence number $0$ means that PPRNN was not used to update the hidden states.
\fix{Compared to the recurrence number $0$, recurrence number $1$ significantly improves the deblurring accuracy.
This justifies that our PPRNN cell, utilizing past and current frame features to update the hidden states, is effective even if used only once.
In addition, the deblurring quality consistently improves with more recurrences, which verifies our iterative adopting of PPRNN cell.}
In the rest of the experiments, we used 4 recurrences as the marginal gains in terms of PSNR and SSIM diminishes.

Furthermore, we experimentally justify the balanced update from PPRNN cell architecture.
We compare the deblurred results from the different input configurations to PPRNN from None (no update), $f_{L_{t-1}}$ only, $f_{B_{t}}$ only, and both $f_{L_{t-1}}$ and $f_{B_{t}}$ in Table~\ref{tab:pprnnupdate}.
\fix{Updating hidden states with $f_{L_{t-1}}$ only improves the performance slightly because hidden states~$h_{t-1}$, generated from $f_{L_{t-1}}$, already have used information of $f_{L_{t-1}}$.}
Using both features from the past and the present best restores the videos, followed by using $f_{B_{t}}$ to update hidden states.
This is analogous to the previous studies trying to combine the hidden states with the present information~\cite{Kim_2017_ICCV,Nah_2019_CVPR,Zhou_2019_ICCV}.
Moreover, we validate order of $f_{B_{t}}$ and $f_{L_{t-1}}$ to update the hidden states in our supplementary material.

\noindent
\textbf{Effects of SNLA.}
We validate our design choice of SNLA by comparing the effect of basic Non-Local Attention~(NLA) and our SNLA with filtering module in Table~\ref{tab:snla}.
\fix{Our SNLA adds only 0.02\% parameters of the whole network compared with NLA, however, significantly improves the deblurring accuracy ($>$ 0.3dB).}
This verifies that sole attention module is not enough to find the optimal relevance of the features and the hidden states and our filter module helps suppressing the perturbation from irrelevant information.

Furthermore, we show the effect of SNLA in recovering texture in Figure~\ref{fig:snla_ablation}.
With the filtering module put together with the attention mechanism, numbers are recovered to a better readable state.
This is because the filtering module effectively alleviates the overemphasis of irrelevant information.

\input{sections/figs/snla_ablation}
\input{sections/figs/featureablation}
\input{sections/tables/pprnnupdate}
\input{sections/figs/distribution}

\noindent
\textbf{Combination of PPRNN and SNLA.}
In addition to the previous experiments validating each component of our PAHS framework, we also show that PPRNN supplements SNLA well.
By jointly training PPRNN and SNLA, we find that the attention scores found from filtering module of SNLA gets higher by the updates from PPRNN in Figure~\ref{fig:distribution}.
In particular, red bars moving the right side of blue bars show that hidden states updated by PPRNN contain more correlated information to the current frame, demonstrating the effectiveness of both PPRNN and SNLA.
\fix{In Table~\ref{tab:cross_analysis}, we show an ablation study showing the synergy between PPRNN, SNLA, and bidirectional inference in improving deblurring performance.
While each method positively enhances the accuracy, jointly using PPRNN and SNLA, PAHS in Table~\ref{tab:cross_analysis}, further boosts the deblurred video quality.
The gain is due to the PPRNN that helps SNLA by providing both the past and the present information gathered from alternating update so that SNLA would selectively extract informative feature from a wider pool.
In addition, as feature with richer information is produced, bidirectrional inference also benefits from both PPRNN and SNLA.
}

\input{sections/figs/longrange}
\noindent
\textbf{Feature visualization in PAHS.}
We further visualize the feature maps to demonstrate the role of PPRNN and SNLA in Figure~\ref{fig:featurevisualization}.
PPRNN manages feature to successfully contain the information of feature from temporal time steps in Figure~\ref{fig:feature3}.
In addition, SNLA aligns the information of $h^{(n)}_{t-1}$ to be properly used for current time step as shown in Figure~\ref{fig:feature4}.
\fix{Furthermore, we visualize spatially long-range dependency in Figure~\ref{fig:longrange}.
Regardless of spatial distance, our SNLA could find the relevant information between current feature~$f_{B_t}$ and hidden states~$h^{(n)}_{t-1}$.}

\noindent
\textbf{Effects of number of future frames.}
Table~\ref{tab:input_seq} shows that utilizing more future frames improves deblurring accuracy in the bidirectional inference.
As our methods are employed in bidirectional way, number of future frames can affect the performance.
The Table~\ref{tab:input_seq} suggests that exploring more extended time frames can improve the backward hidden states.
As the performance is saturated after using 19 future frames, we use 19 future frames in our final model.

\noindent
\textbf{Generalization of our proposed methods.} 
Since our SNLA and PPRNN can be applied to most of RNN-based video deblurring models without changing the baseline, we show the compatibility of proposed PPRNN and SNLA by adopting them to the previous RNN-based video deblurring models~\cite{hochreiter1997long,Kim_2017_ICCV,Nah_2019_CVPR,Zhou_2019_ICCV} in Table~\ref{tab:compatibility}.
We trained each model for 500 epochs.
As shown, our proposed methods enhance the performances consistently.

\input{sections/tables/compatibility}

\input{sections/figs/real}
\input{sections/figs/spiderman}

\subsection{Comparisons with State-of-the-Art}
We compare our model with the existing state-of-the-art video deblurring methods~\cite{Kim_2017_ICCV,li2021arvo,Nah_2019_CVPR,Pan_2020_CVPR,park2020blur,Tao_2018_CVPR,Wang_2019_CVPR_Workshops,zhong2020efficient,Zhou_2019_ICCV} on the public DVD, GOPRO and REDS datasets.
All models in comparison are based on deep neural networks.
The quantitative comparisons on DVD~\cite{Su_2017_CVPR}, GOPRO~\cite{Nah_2017_CVPR}, and REDS~\cite{Nah_2019_CVPR_Workshops_REDS} datasets are shown in Table~\ref{tab:GOPRO_REDS} with the corresponding model sizes and speed.
The results on the benchmark datasets show that our method PAHS, retains a similar model size as ARVo~\cite{li2021arvo}, presents large performance gains in deblurring.
Specifically, PAHS is 9.53 times faster than ARVo, the second highest performing model in Table~\ref{tab:GOPRO_REDS}.

\fix{The qualitative comparisons are in Figures~\ref{fig:dvd} and \ref{fig:gopro_reds}.
CNN-based methods~\cite{li2021arvo,Pan_2020_CVPR} use a predetermined number of consecutive frames.
Thus they cannot handle the long-term dependency and fail to restore the structure of the scene if the consecutive frames are severely blurred like the middle frame as shown in Figure~\ref{fig:dvd_tsp} and Figure~\ref{fig:dvd_arvo}.
For detailed validation, please refer to the supplementary material.
Moreover, although RNN-based methods~\cite{Kim_2017_ICCV,Nah_2019_CVPR,zhong2020efficient,Zhou_2019_ICCV} manage long-term dependency via hidden states, they overlook and do not utilize spatially distant but relevant information of hidden states.
Thus their methods are less effective when spatial information is not preserved during the temporal steps due to the strong motion blur.
As can be seen in Figure~\ref{gr_ifirnn} and Figure~\ref{gr_estrnn}, they fail to deblur the structure of scene objects.
In contrast, our SNLA can exploit temporal information regardless of misalignment because it exploits relevant information by considering the correlation between all spatial information of the hidden states and the current frame.
In addition, our PPRNN extends the range of information that the hidden states can have to the current time step, helping SNLA bring related spatial information from the current frame itself.
Figures~\ref{fig:dvd_pahs} and \ref{gr_pahs} visually show the efficacy of our PAHS in restoring sharp details on fast-moving objects and textured area.
The validation of the large blur situation is included in the supplementary material.}
Furthermore, we show that our method well generalizes to deblurring real videos in Figures~\ref{fig:real} and \ref{fig:spiderman}.

%% file: sections/figs/bidirection.tex
\begin{figure}[t]
    \centering
    \captionsetup[subfloat]{font=tiny}
    \renewcommand{\wp}{0.98\linewidth}
    \begin{minipage}{0.5\textwidth}
    \includegraphics[width=\wp]{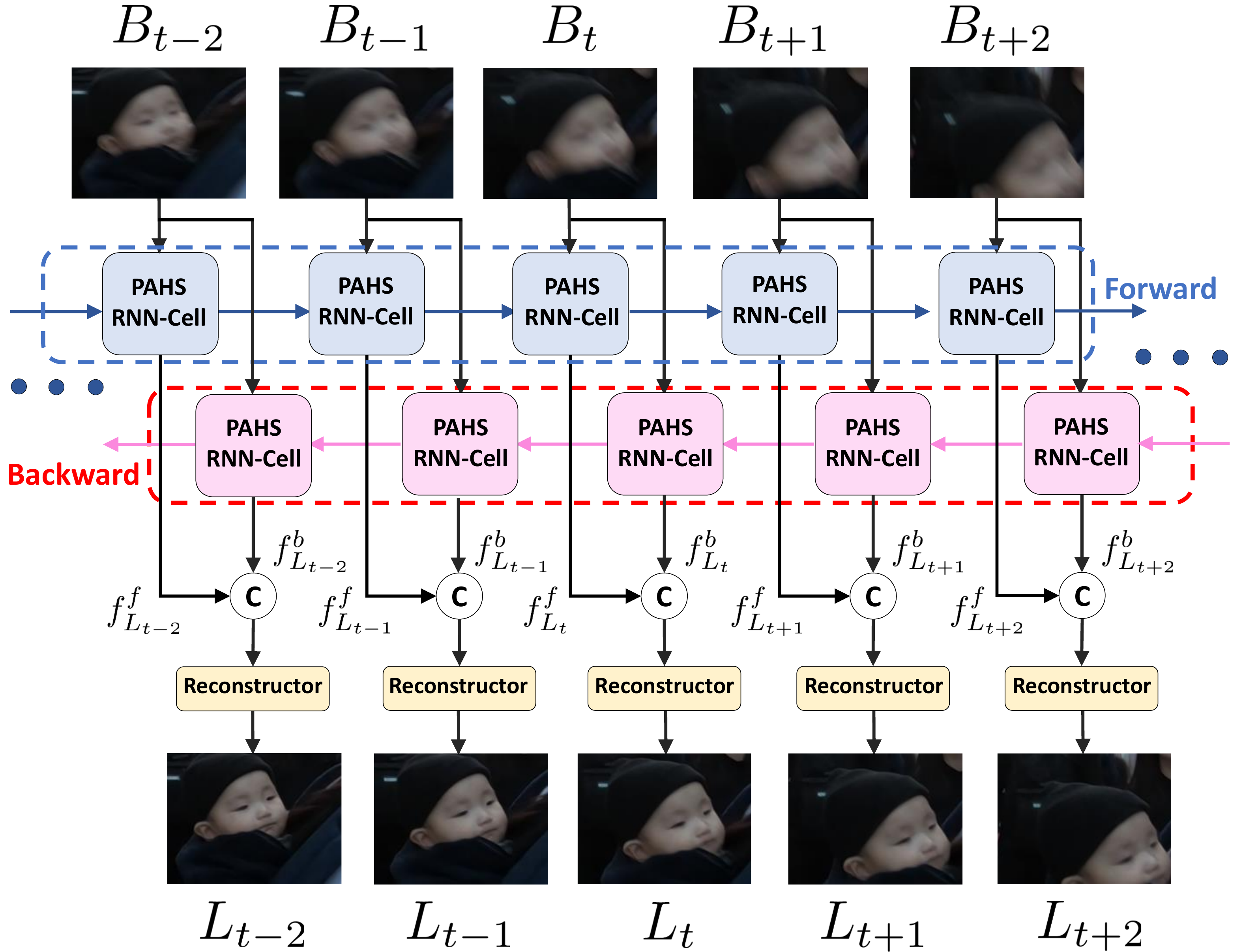}
    \end{minipage}
    \qquad
    \resizebox{0.4\linewidth}{!}{
    \begin{tabular}{c|cc}
    \toprule
    \# future frames & PSNR & SSIM \\
    \midrule
    0 & 32.85 & 0.9380 \\
    3 & 33.52 & 0.9482 \\
    7 & 33.73 & 0.9482 \\
    11 & 33.73 &  0.9502 \\
    15 & 33.78 &  0.9609 \\
    \textbf{19} & \textbf{33.82}  & \textbf{0.9612}\\
    23 & 33.81 & \textbf{0.9612}\\
    \bottomrule
    \end{tabular}
    }
    \captionlistentry[table]{A table beside a figure}
    \label{tab:input_seq}
    \captionsetup{labelformat=andtable}
    \figcspace
    \caption{\textbf{(left)} \textbf{Bidirectional PAHS inference structure.} \textbf{(right)} \textbf{Effects of number of the backward computation in bidirectional inference.} We highlight the setting in our final model in bold.}
    \label{fig:bidirection}
    \figspace
  \end{figure}



%% file: sections/figs/graph-pprnnnum.tex


\begin{figure}[t]
    \newcommand{\cmark}{\textcolor{MyGreen}{\ding{51}}}
    \newcommand{\xmark}{\ding{55}}%
    \centering
    \captionsetup[subfloat]{font=tiny}
    \renewcommand{\wp}{0.98\linewidth}
    \begin{minipage}{0.58\textwidth}
    \includegraphics[width=\wp]{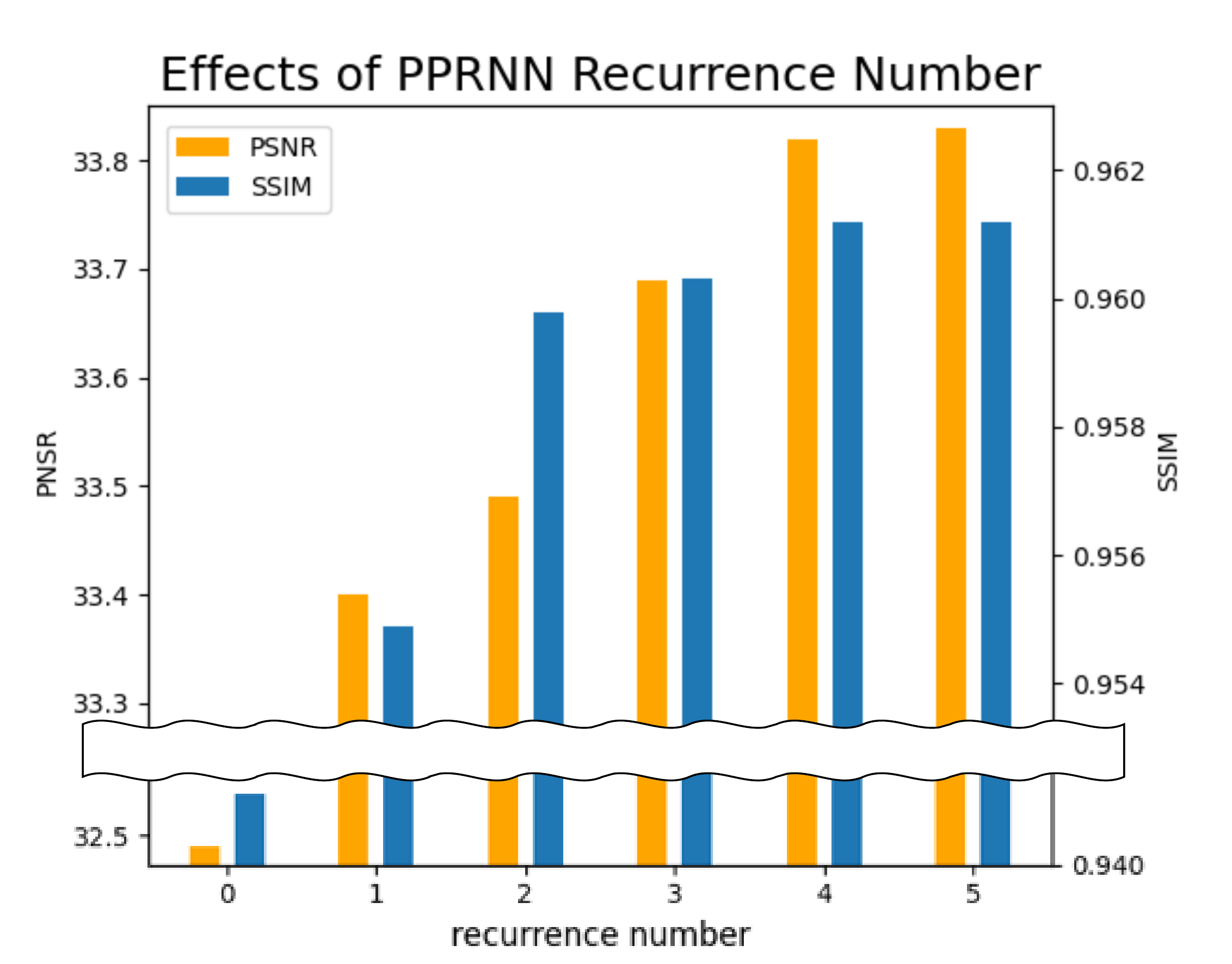}
    \end{minipage}
    \qquad
    \resizebox{0.35\linewidth}{!}{
    \begin{tabular}{cc|cc}
        \toprule
         $f_{L_{t-1}}$ & $f_{B_{t}}$ & PSNR & SSIM\\
        \midrule
        \xmark & \xmark & 32.44 & 0.9430 \\
        \cmark & \xmark & 32.45 & 0.9498   \\
        \xmark & \cmark & 33.01 & 0.9554 \\
        \cmark & \cmark & \textbf{33.82} & \textbf{0.9612}  \\
        \bottomrule
    \end{tabular}
    }
    \captionlistentry[table]{A table beside a figure}
    \captionsetup{labelformat=andtable}
    \label{tab:pprnnupdate}
    \tabcspace
    \caption{\textbf{(left)} \textbf{Effect of the PPRNN recurrence number.} 
    We use 4 recurrences for our final model.
    \textbf{(right)} \textbf{Effects of using $f_{L_{t-1}}$ and $f_{B_{t}}$ to update the hidden states in PPRNN.}}
    \label{fig:pprnn-graph}
    \tabspace
  \end{figure}

%% file: sections/figs/snla_ablation.tex
\begin{figure}[t]
    \centering
    \captionsetup[subfloat]{font=scriptsize}
    \newcommand{\cmark}{\textcolor{MyGreen}{\ding{51}}}
    \newcommand{\xmark}{\ding{55}}%
    \renewcommand{\wp}{0.48\linewidth}
    \begin{minipage}{0.4\textwidth}
    \subfloat[Input \label{fig:snla-a}]{\includegraphics[width=\wp]{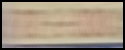}}
    \hspace{1mm}
    \subfloat[Baseline \label{fig:snla-b}]{\includegraphics[width=\wp]{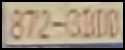}}\\
    \subfloat[NLA \label{fig:snla-c}]{\includegraphics[width=\wp]{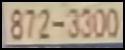}}
    \hspace{1mm}
    \subfloat[SNLA (Ours) \label{fig:snla-d}]{\includegraphics[width=\wp]{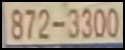}}
    \end{minipage}
    \qquad
    \resizebox{0.45\linewidth}{!}{
    \begin{tabular}{c|cc|cc}
        \toprule
        Method & Attention & Filtering module & PSNR & SSIM \\
        \midrule
        Baseline & \xmark & \xmark & 32.02 & 0.9361 \\
        NLA & \cmark & \xmark & 33.51 & 0.9583 \\
        \textbf{SNLA} & \cmark & \cmark & \textbf{33.82} & \textbf{0.9612}\\
        \bottomrule
    \end{tabular}
    }
    \captionlistentry[table]{A table beside a figure}
    \captionsetup{labelformat=andtable}
    \label{tab:snla}
    \tabcspace
    \caption{\textbf{(left)} \textbf{Qualitative effect of SNLA and its components.}
    Using both attention and filtering modules in SNLA produces the best deblurred results without noisy artifacts. \textbf{(right)} \textbf{Effect of SNLA and its components.} Both the attention and filtering modules make substantial contribution to the performance improvement.}
    \label{fig:snla_ablation}
  \end{figure}

%% file: sections/figs/featureablation.tex
\begin{figure}[t]
    \renewcommand{\wp}{0.242\linewidth}
    \centering
    \subfloat{\includegraphics[width=\wp]{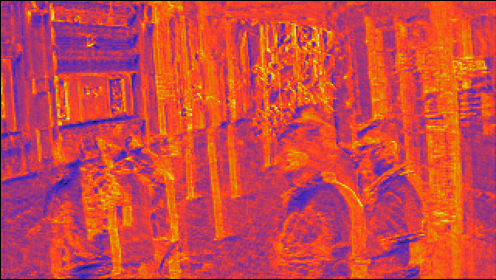}}
    \hfill
    \subfloat{\includegraphics[width=\wp]{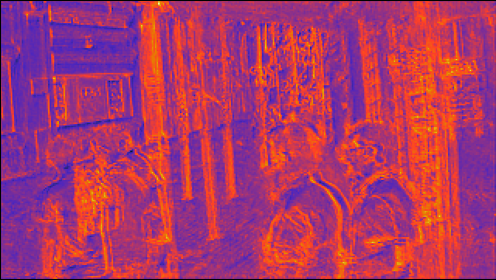}}
    \hfill
    \subfloat{\includegraphics[width=\wp]{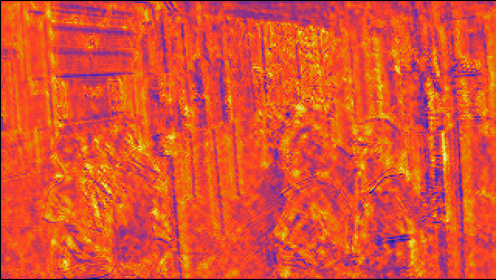}}
    \hfill
    \subfloat{\includegraphics[width=\wp]{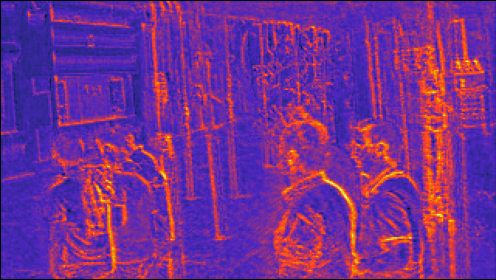}}
    \addtocounter{subfigure}{-4}
    \\
    \centering
    \subfloat[$f_{B_t}$\label{fig:feature1}]{\includegraphics[width=\wp]{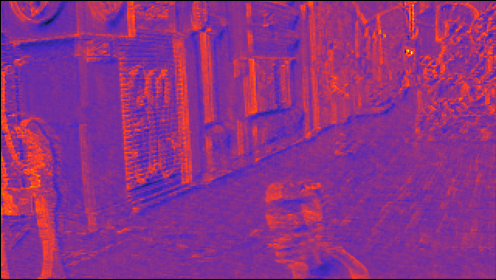}}
    \hfill
    \subfloat[$h_{t-1}$\label{fig:feature2}]{\includegraphics[width=\wp]{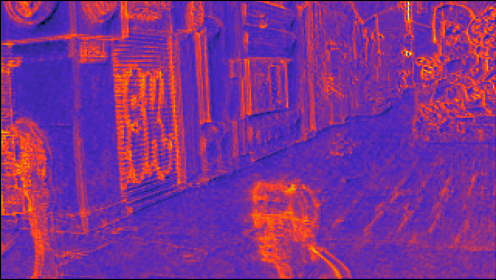}}
    \hfill
    \subfloat[$h^{(n)}_{t-1}$\label{fig:feature3}]{\includegraphics[width=\wp]{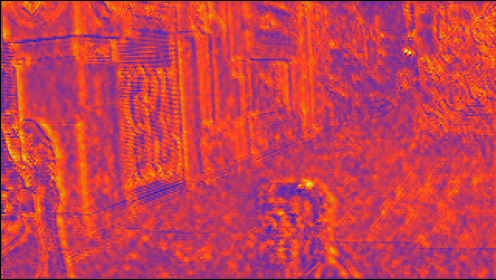}}
    \hfill
    \subfloat[${\tilde{h}}_{t-1}$\label{fig:feature4}]{\includegraphics[width=\wp]{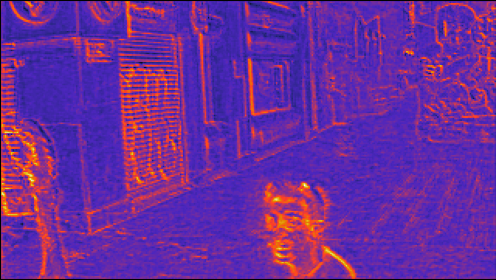}}\\
    \figcspace
    \caption{
        \textbf{Visualization of feature maps from our PAHS.}
        (a) $f_{B_t}$ extracted from current frame.
        (b) hidden state~$h_{t-1}$ without any updating.
        (c) updated hidden state from PPRNN. It successfully aggregates useful information from both (a) and (b).
        (d) output feature of SNLA. SNLA selectively searches and gathers the useful information from every pixels in (c) so that sharp details are constructed in (d).
    }
    \label{fig:featurevisualization}
\end{figure}

%% file: sections/tables/pprnnupdate.tex

%% file: sections/figs/distribution.tex
\begin{figure}[t]
    \centering
    \renewcommand{\wp}{0.51\linewidth}
    \includegraphics[width=\wp]{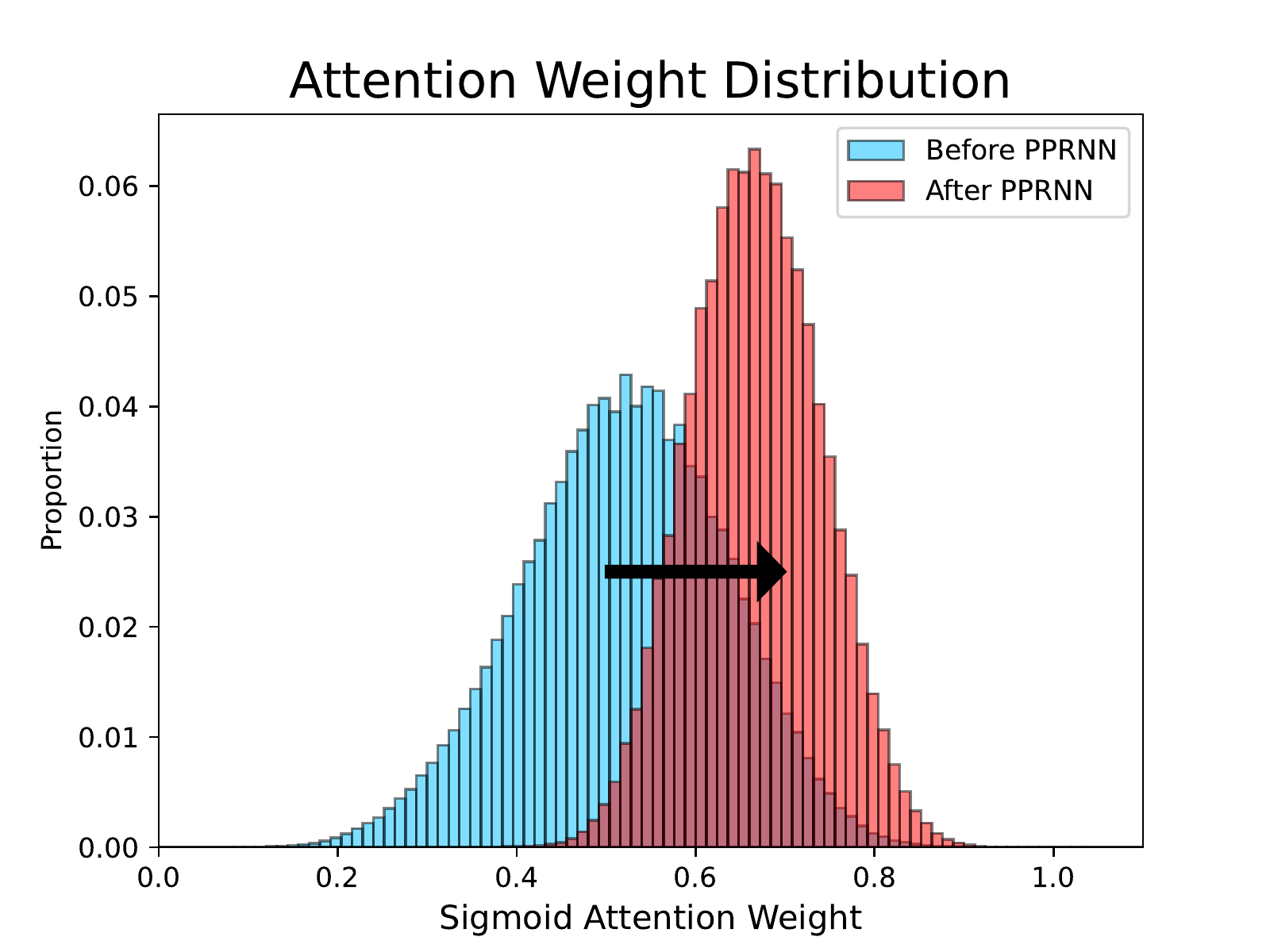}
    \figcspace
    \caption{
        \textbf{Distribution of sum of attention weights in the filtering module of SNLA on GOPRO~\cite{Nah_2017_CVPR} dataset.} 
        $S_{Sel}$ obtained from filtering module can handle the undesired information to be discarded.
    }
    \label{fig:distribution}
\end{figure}

%% file: sections/figs/longrange.tex
\begin{figure}[t]
    \centering
    \renewcommand{\wp}{0.48\linewidth}
    \includegraphics[width=\wp]{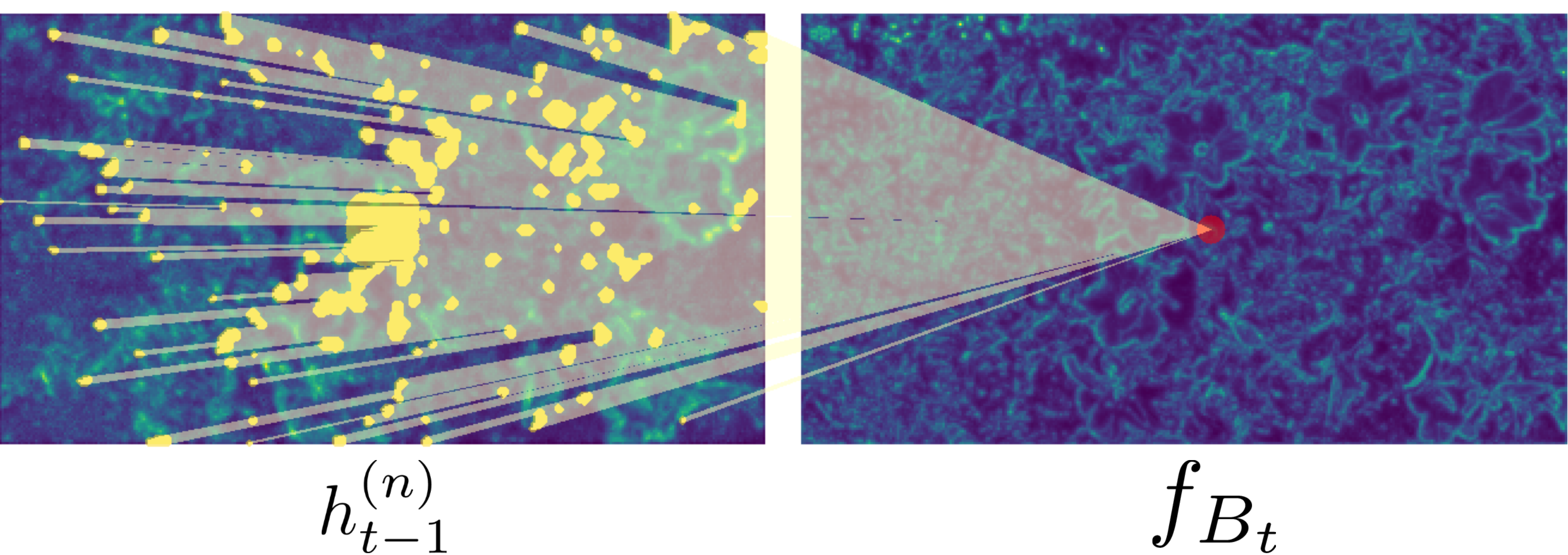}
    \includegraphics[width=\wp]{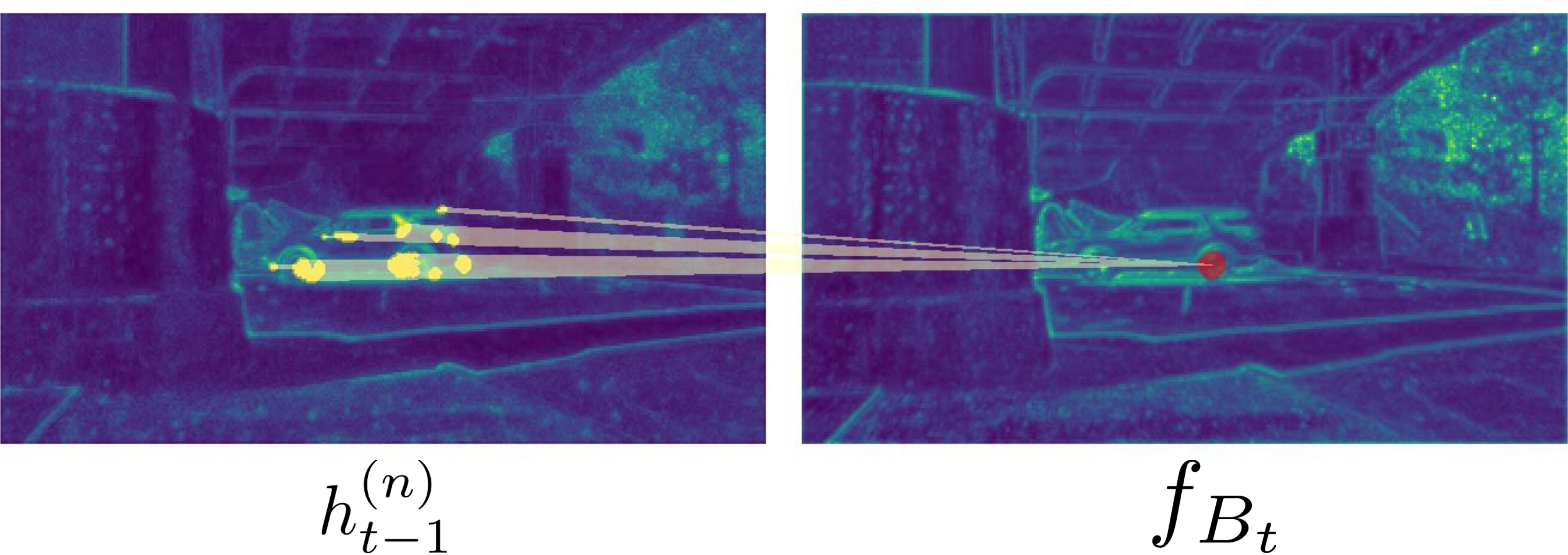}
    \caption{\textbf{Visualization of long-range dependency.} We visualize strong attention, where normalized $S_{\text{NL}}\otimes S_{\text{Sel}}$~(Eq.~\ref{eq:nonlocal2} and \ref{eq:nonlocal3}) is bigger than certain threshold~(0.6), between the all pixels of $h^{(n)}_{t-1}$ and the red point of $f_{B_t}$.
    SNLA exploits information with high relevance to $f_{B_t}$ from spatially remote position of $h^{(n)}_{t-1}$.}
    \label{fig:longrange}
    \figspace
\end{figure}

%% file: sections/tables/compatibility.tex

\begin{table}[t]
\centering
\begin{minipage}[t]{0.48\linewidth}
\begin{tabular}{c|cc|cc}
    \toprule
    \multirow{2}{*}{models} &
    \multicolumn{2}{c|}{Baseline} & 
    \multicolumn{2}{c}{PAHS} \\
    & PSNR & SSIM & PSNR & SSIM\\
    \midrule
    LSTM~\cite{hochreiter1997long} & 25.22 & 0.7948 & \textbf{27.84} & \textbf{0.8589}\\
    OVD~\cite{Kim_2017_ICCV} & 28.72 & 0.8460 & \textbf{28.83} & \textbf{0.8577}\\
    IFI-RNN~\cite{Nah_2019_CVPR} & 28.30 & 0.8668 & \textbf{29.12} & \textbf{0.8906}\\
    STFAN~\cite{Zhou_2019_ICCV} & 28.77 &  0.8776 & \textbf{30.92} & \textbf{0.9107} \\
    \bottomrule
\end{tabular}
\tabcspace
\caption{\textbf{Generalization of PAHS to other RNN-based models.}
}
\label{tab:compatibility}
\end{minipage}%
\hfill
\begin{minipage}[t]{0.48\linewidth}
\centering
    \begin{tabular}{c|cc|cc}
    \toprule
    \multirow{2}{*}{models} &
    \multicolumn{2}{c|}{uni-dir} & 
    \multicolumn{2}{c}{bi-dir} \\
    & PSNR & SSIM & PSNR & SSIM\\
    \midrule
    {\scriptsize Baseline} & 31.29 & 0.9186 & 31.78 & 0.9216\\
    {\scriptsize +PPRNN only} & 31.57 & 0.9190 & 32.02 & 0.9361\\
    {\scriptsize +SNLA only} & 31.98 & 0.9293 & 32.44 & 0.9430\\
    \textbf{PAHS} & \textbf{32.85} & \textbf{0.9380} &  \textbf{33.82} & \textbf{0.9612}\\
    \bottomrule
\end{tabular}
\tabcspace
\caption{\textbf{Synergy among PPRNN, SNLA, and bidireciton.}}
\label{tab:cross_analysis}
\end{minipage}
\end{table}

%% file: sections/figs/real.tex
\begin{figure}[t]
    \captionsetup[subfloat]{font=scriptsize}
    \renewcommand{\wp}{0.195\linewidth}
    \centering
    \subfloat{\includegraphics[width=\wp]{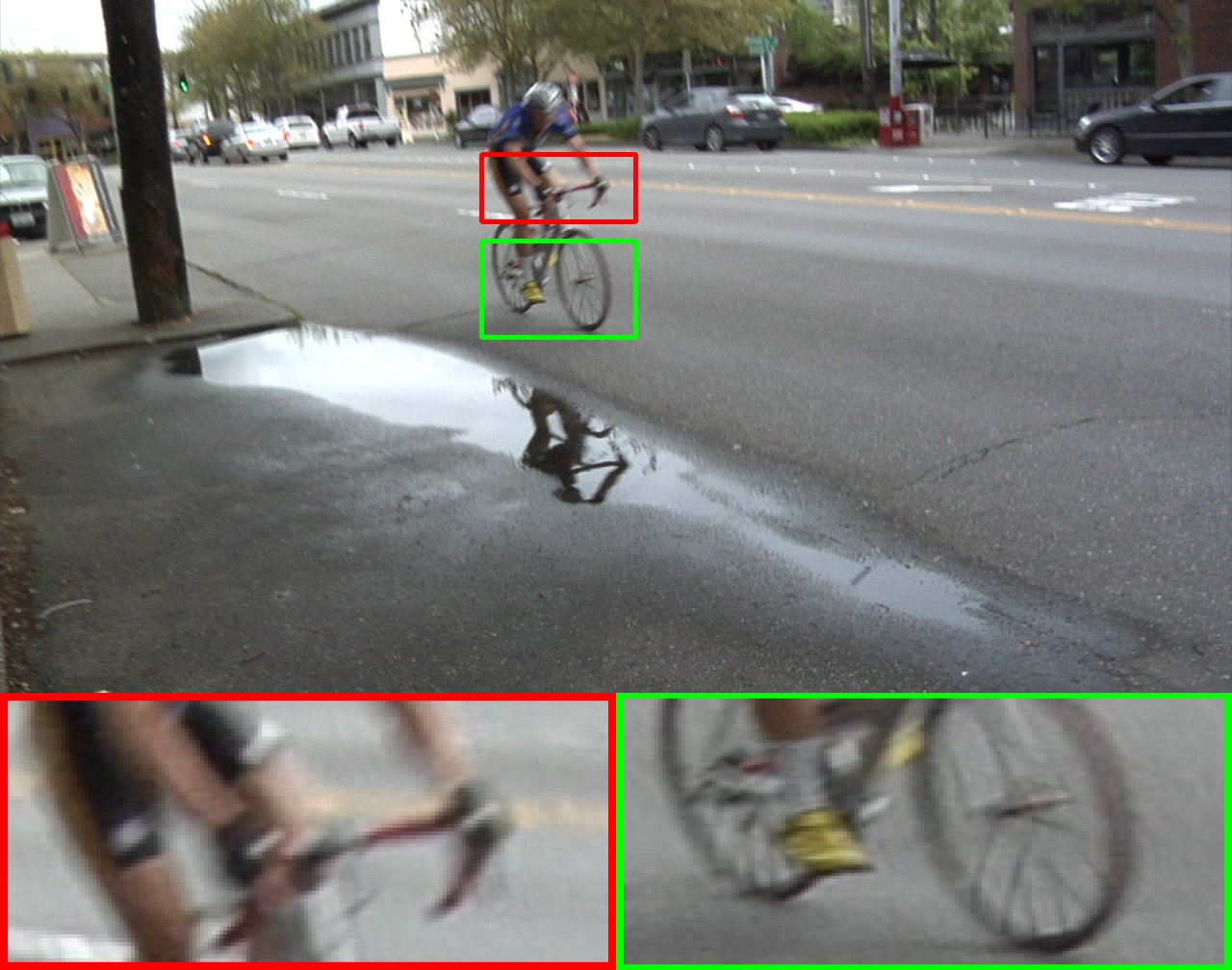}}
    \hfill
    \subfloat{\includegraphics[width=\wp]{figs/real/bike/blur.jpg}}
    \hfill
    \subfloat{\includegraphics[width=\wp]{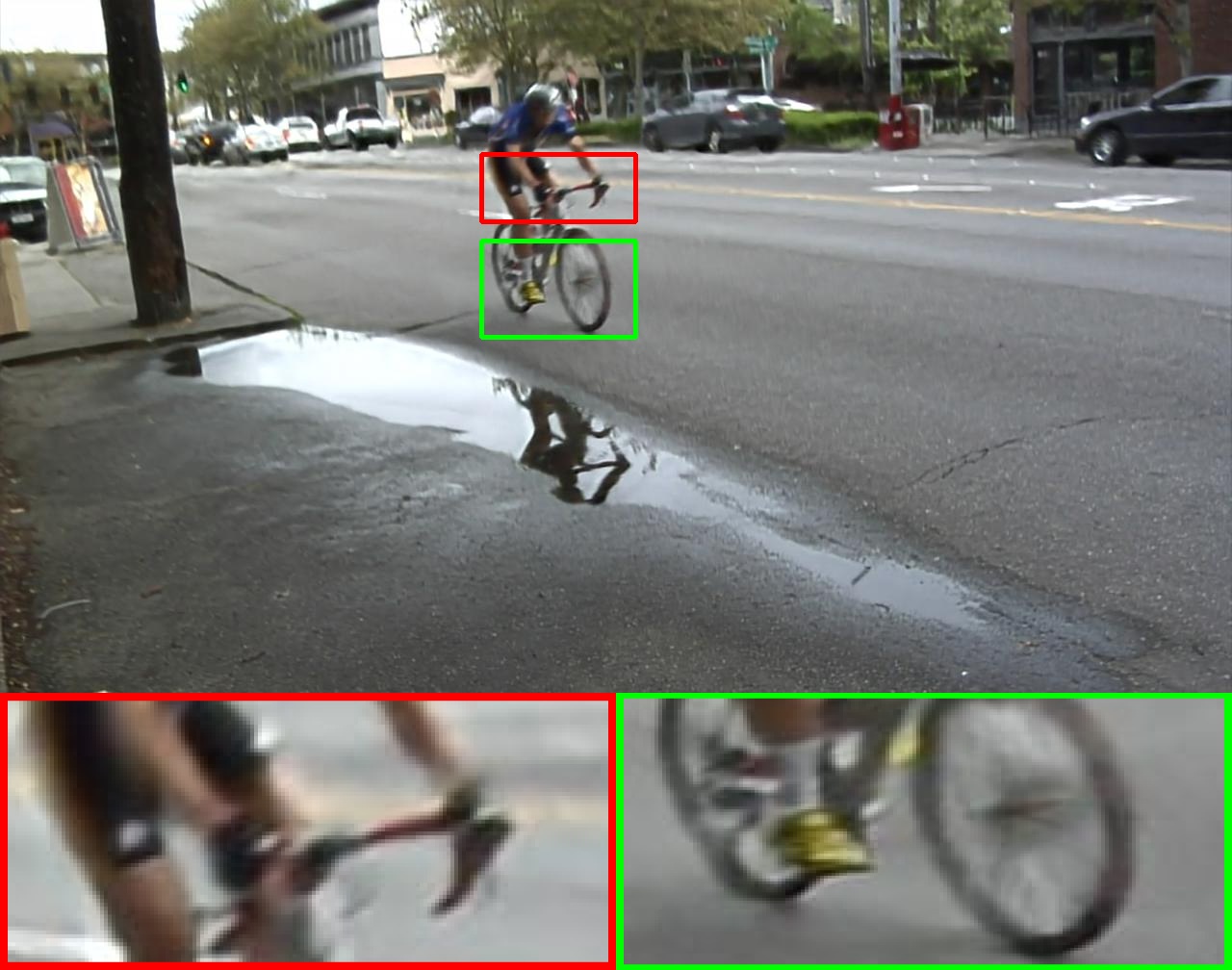}}
    \hfill
    \subfloat{\includegraphics[width=\wp]{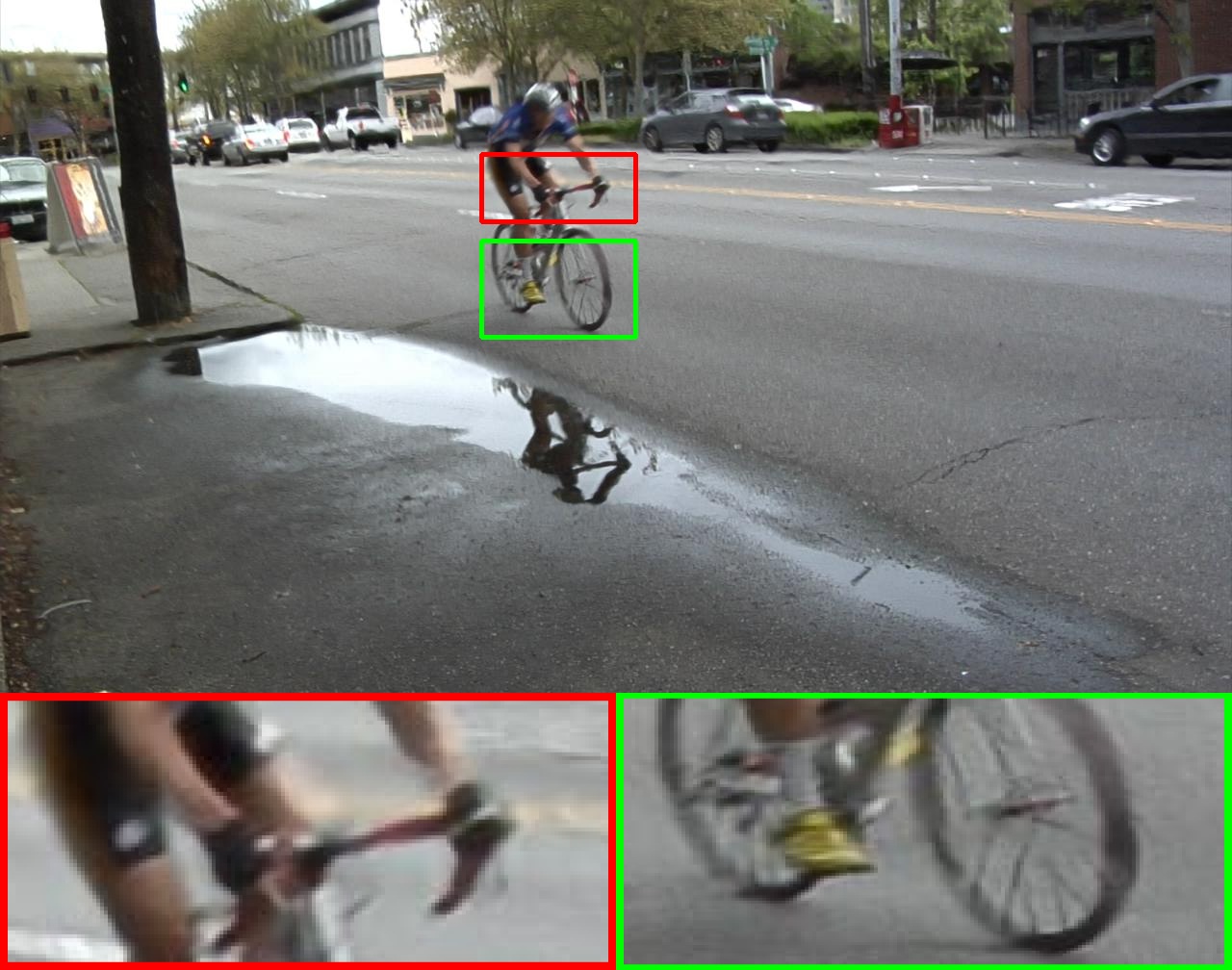}}
    \hfill
    \subfloat{\includegraphics[width=\wp]{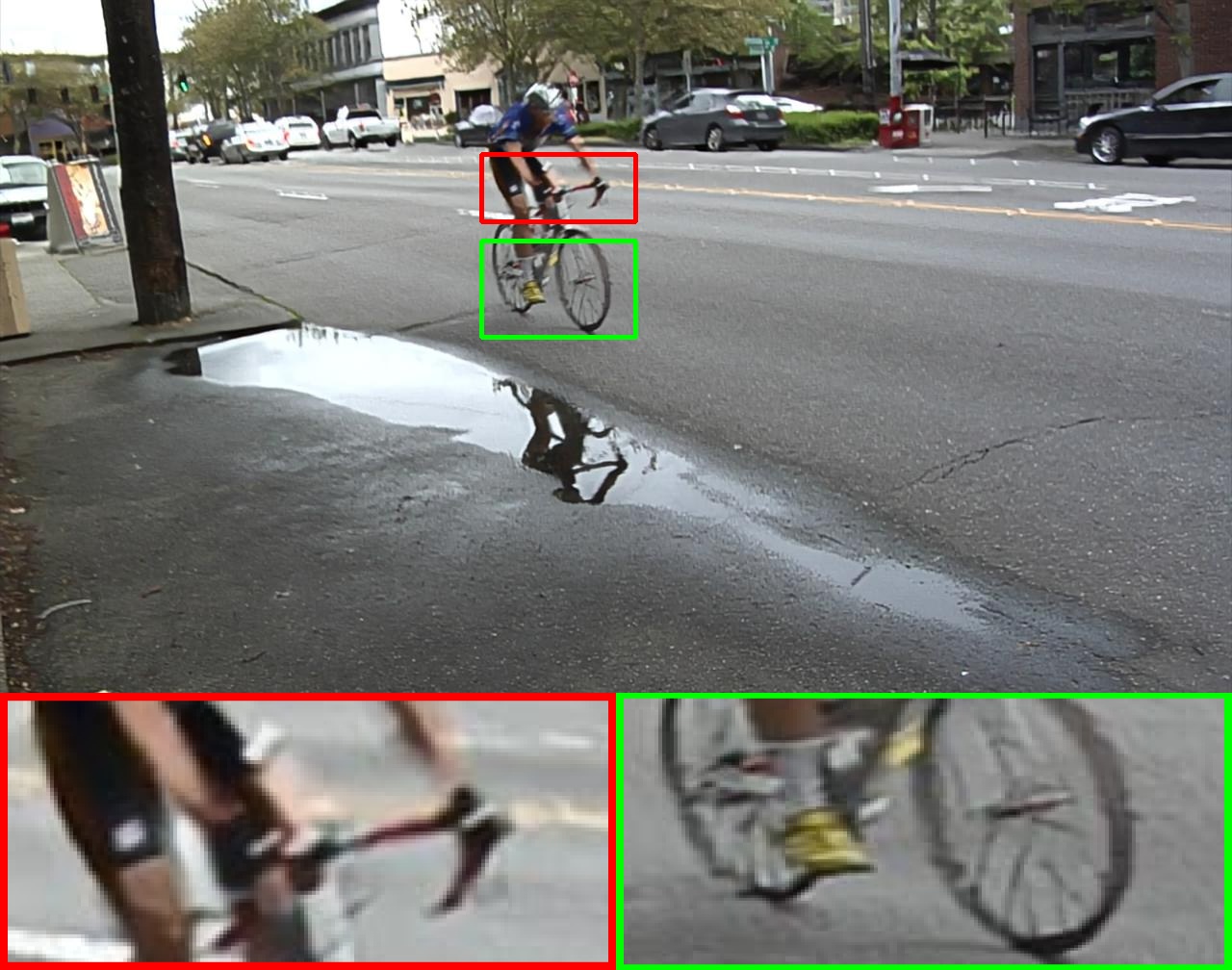}}
    \addtocounter{subfigure}{-5}
    \\
    \centering
    \subfloat[Input]{\includegraphics[width=\wp]{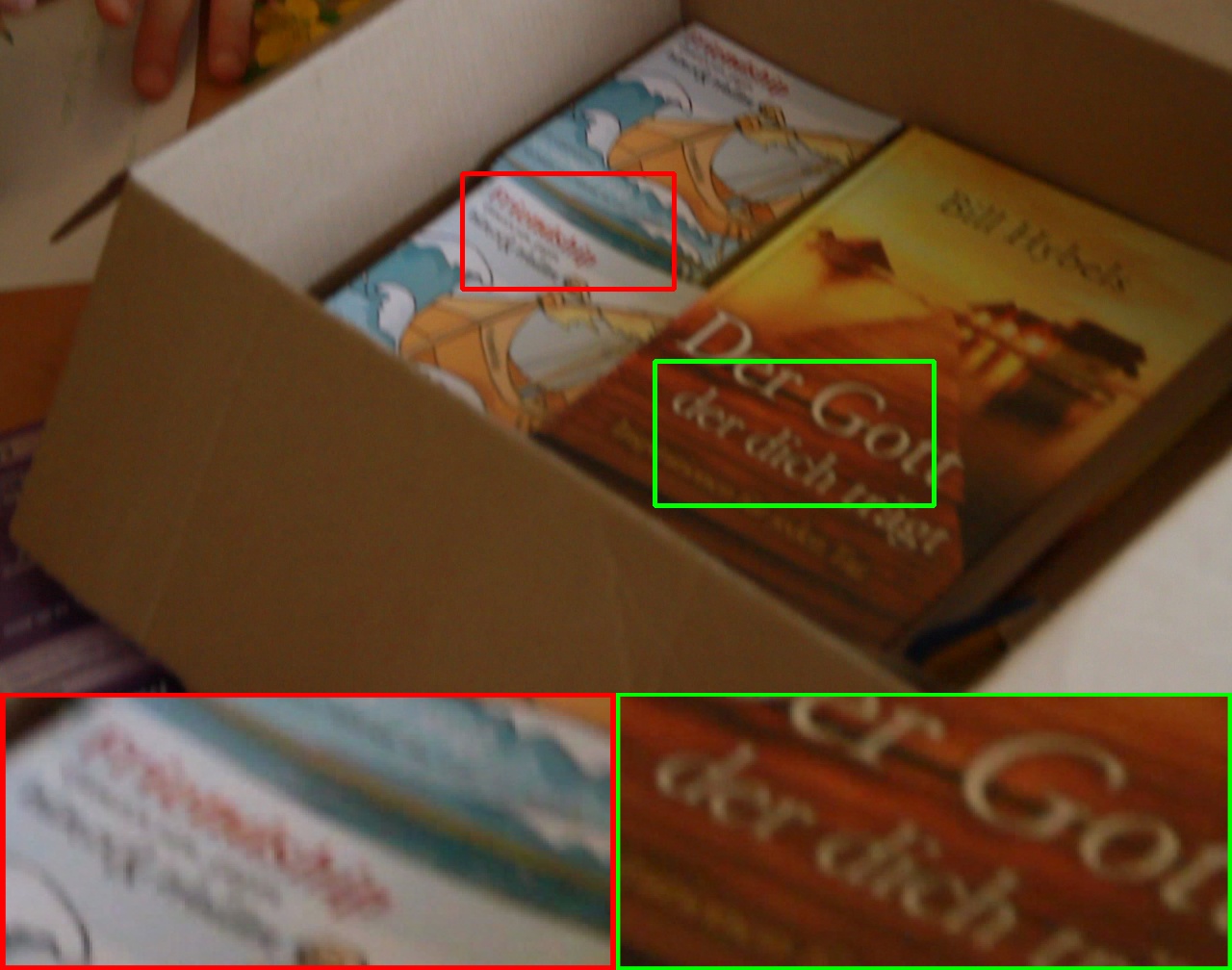}}
    \hfill
    \subfloat[IFI-RNN~\cite{Nah_2019_CVPR}]{\includegraphics[width=\wp]{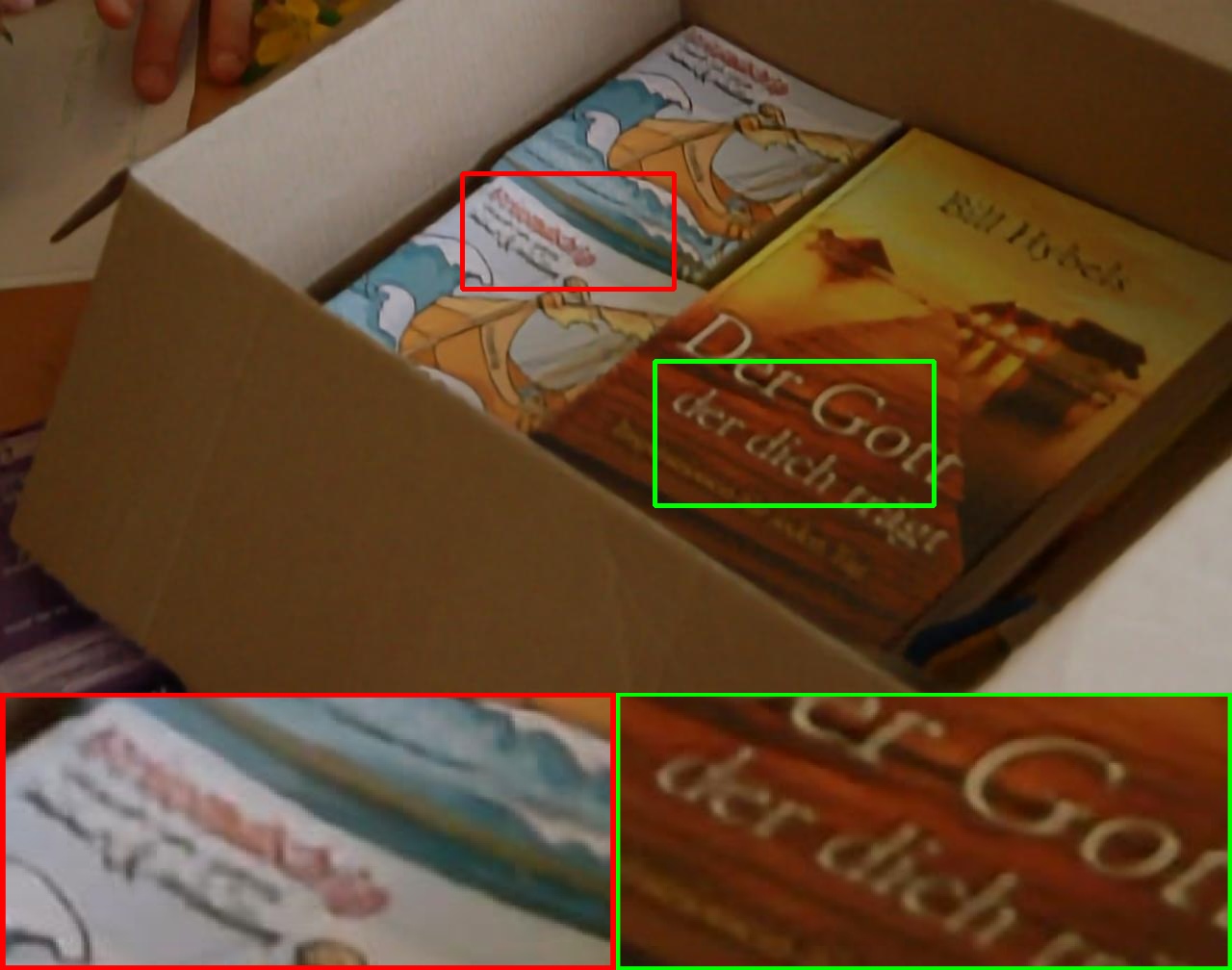}}
    \hfill
    \subfloat[STFAN~\cite{Zhou_2019_ICCV}]{\includegraphics[width=\wp]{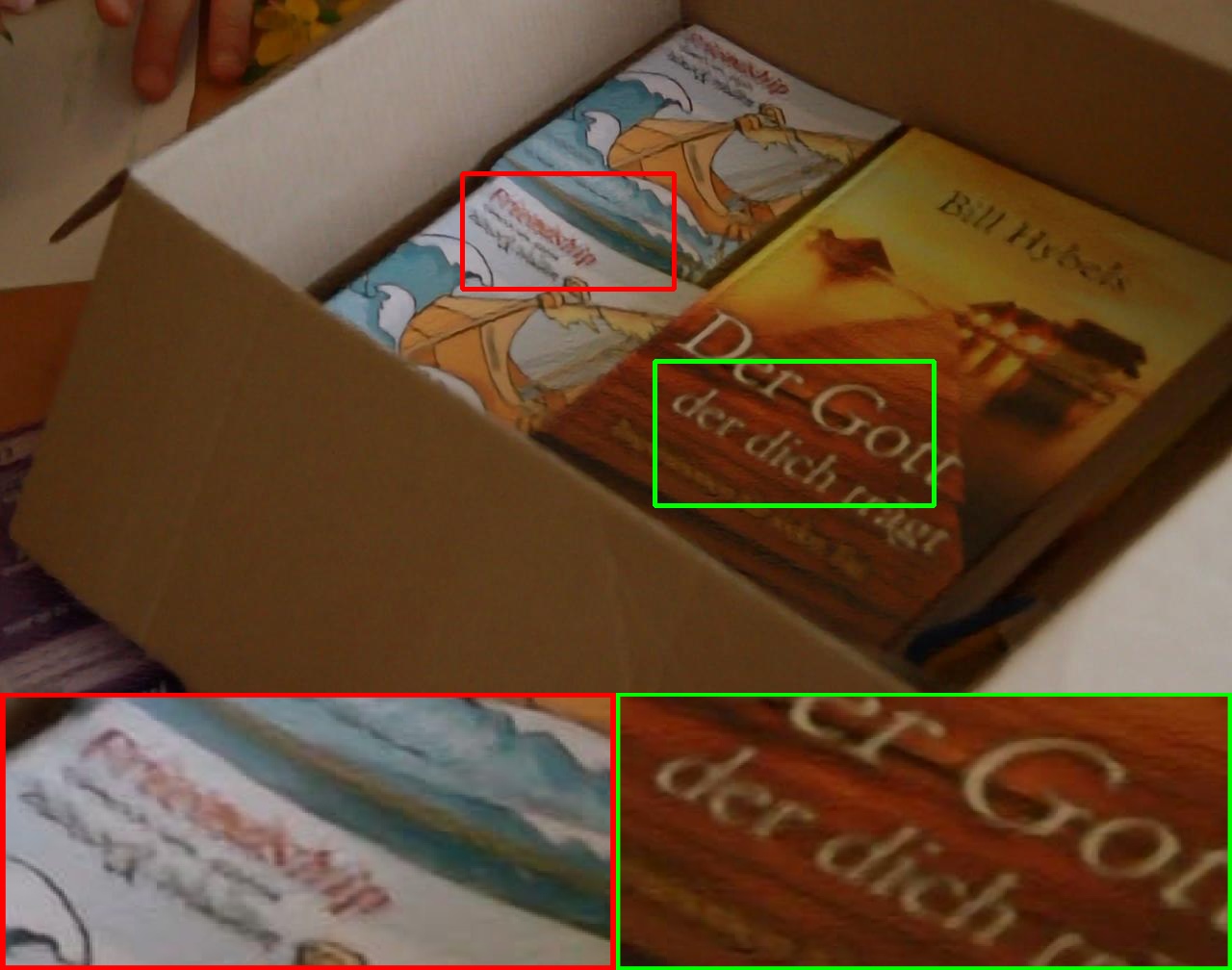}}
    \hfill
    \subfloat[ESTRNN~\cite{zhong2020efficient}]{\includegraphics[width=\wp]{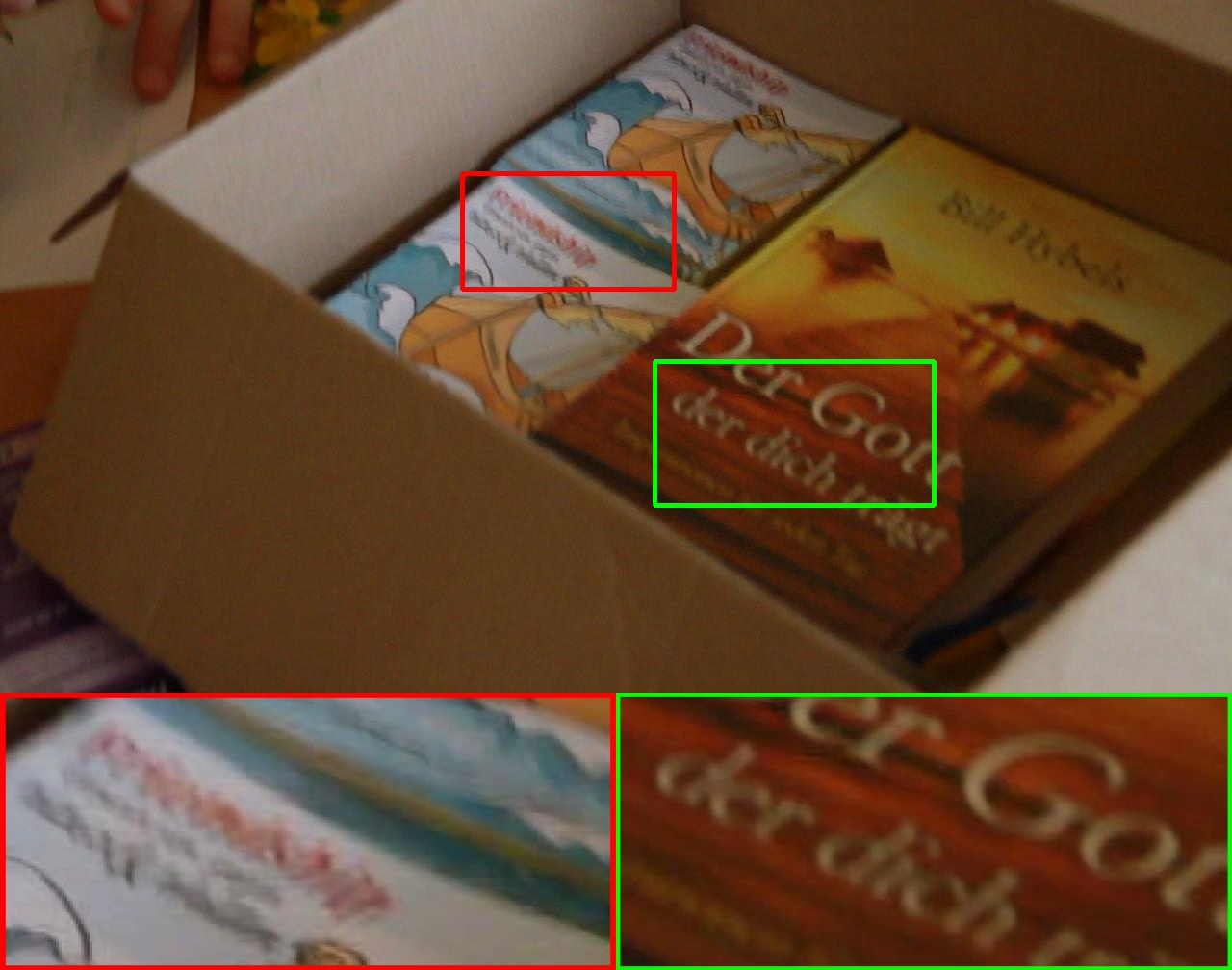}}
    \hfill
    \subfloat[\textbf{PAHS (Ours)}]{\includegraphics[width=\wp]{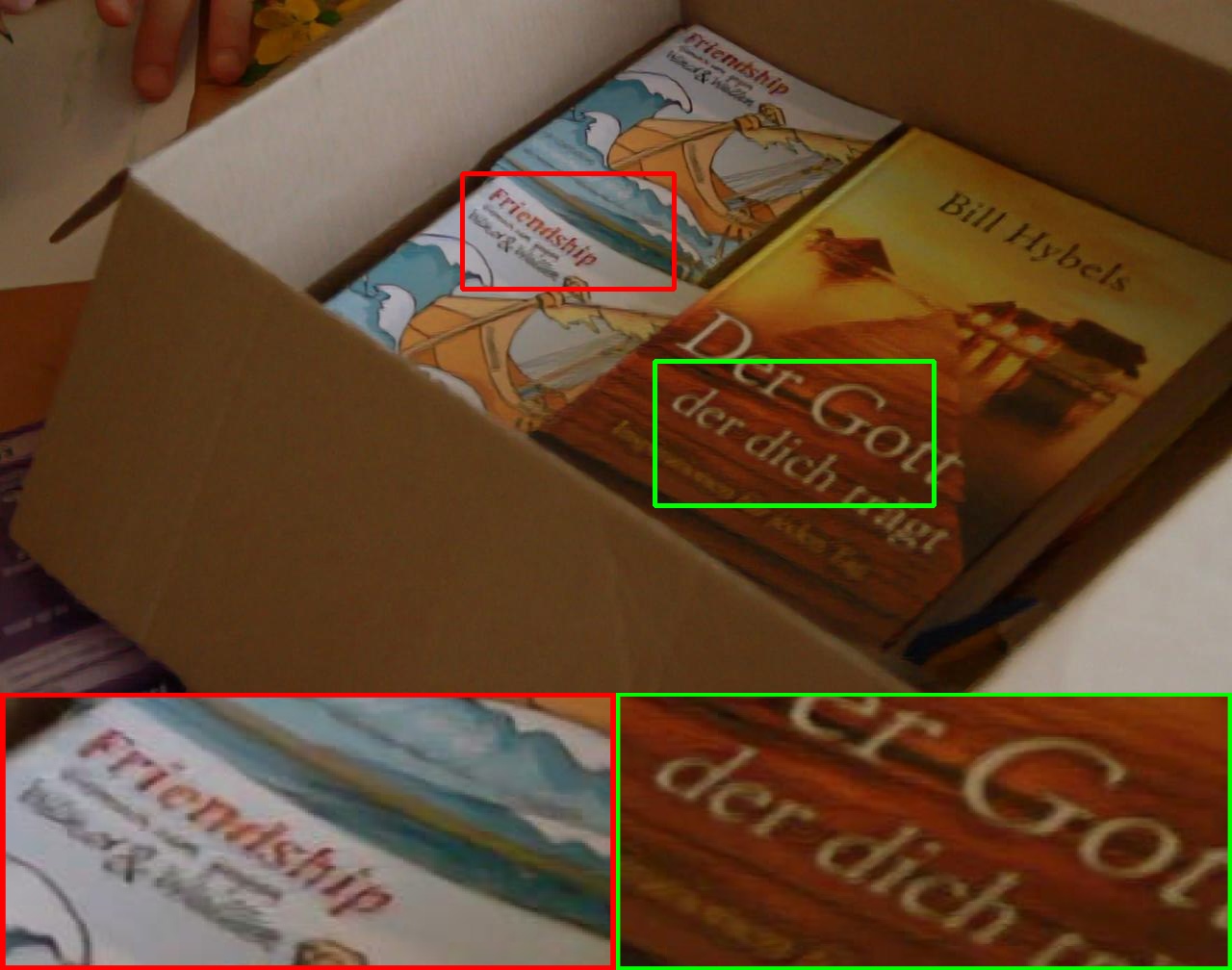}}
    \figcspace
    \caption{\textbf{Qualitative comparison with state-of-the-art methods on real blurry video dataset~\cite{cho2012registration}.}}
    \label{fig:real}
    \figspace
\end{figure}

%% file: sections/figs/spiderman.tex
\begin{figure*}[t]
    \renewcommand{\wp}{0.24\linewidth}
    \subfloat[Input]{\includegraphics[width=\wp]{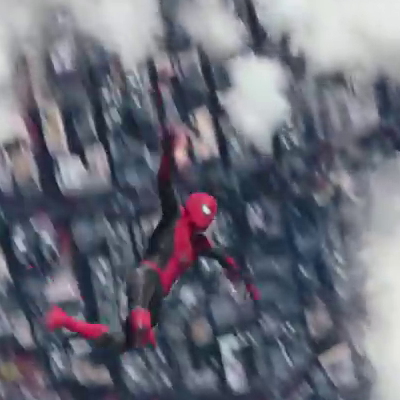}}
    \hfill
    \subfloat[ESTRNN~\cite{zhong2020efficient}]{\includegraphics[width=\wp]{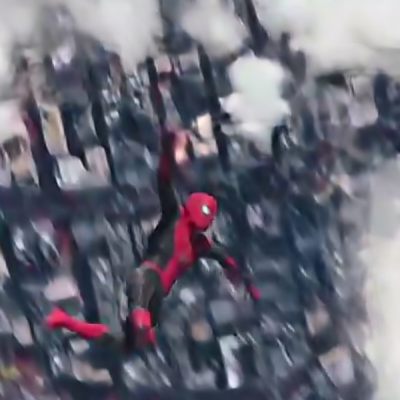}}
    \hfill
    \subfloat[ARVo~\cite{li2021arvo}]{\includegraphics[width=\wp]{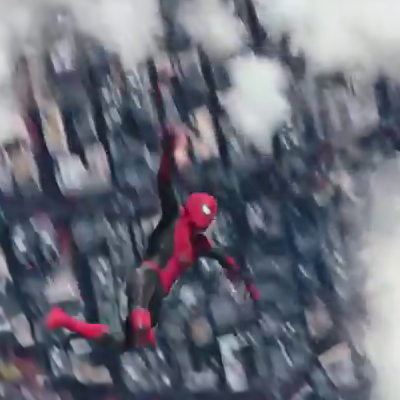}}
    \hfill
    \subfloat[\textbf{PAHS~(Ours)}]{\includegraphics[width=\wp]{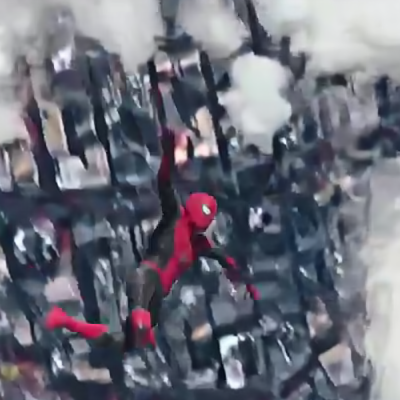}}
    \figcspace
    \caption{\textbf{Qualitative comparison with state-of-the-art methods on a Youtube video.}}
    \label{fig:spiderman}
    \figspace
\end{figure*}

%% file: sections/5_conclusion.tex
\section{Conclusion}
In this work, we proposed PAHS framework with PPRNN and SNLA module to improve the performance of RNN-based
video deblurring.
PPRNN looks into both the previous frame and the current frame to update the hidden state with balance so that useful information from past can be preserved as well.
In addition, SNLA module further adapts the hidden state by the positional information obtained by non-local attention from the input frame feature.
In the filtering process, less relative information is suppressed to let the reconstruction module focus on more useful features.
With both the modules, our PAHS framework can be applied to general RNN-based video deblurring methods.
\fix{Moreover, PPRNN and SNLA create positive synergies because PPRNN enlarges the pool of the hidden states,  utilized in SNLA, by referencing both the past and the present information.}
Our experimental results demonstrate our method consistently improves the deblurring performance by large margin on diverse datasets and real-world examples.

%% file: supplementary/fig/model_all.tex
\begin{figure}[t]
    \centering
    \renewcommand{\wp}{0.45\linewidth}
    \newcommand{\hp}{\hspace{0.1\linewidth}}
    \newcommand{\hhp}{\hspace{0.05\linewidth}}
    \subfloat{\includegraphics[width=0.9\linewidth]{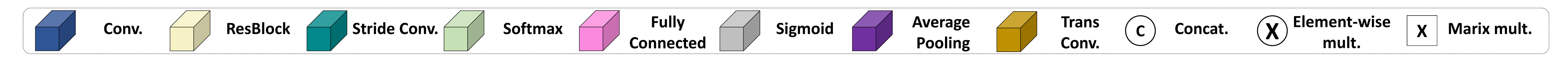}}
    \addtocounter{subfigure}{-1}
    \hspace{-3mm}
    \\
    \subfloat[Overall architecture of PAHS~(Ours) \label{sup_fig:model_all_all}]{
    \includegraphics[width=0.98\linewidth]{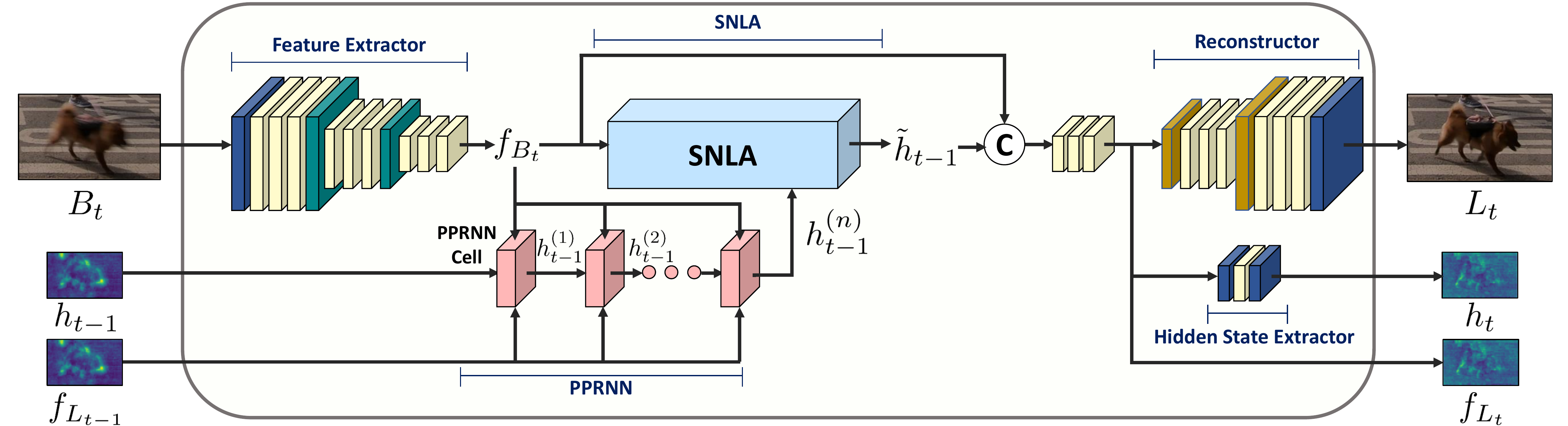}
    }
    \\
    \subfloat[PPRNN cell structure \label{sup_fig:pprnn}]{
    \includegraphics[height=0.15\linewidth]{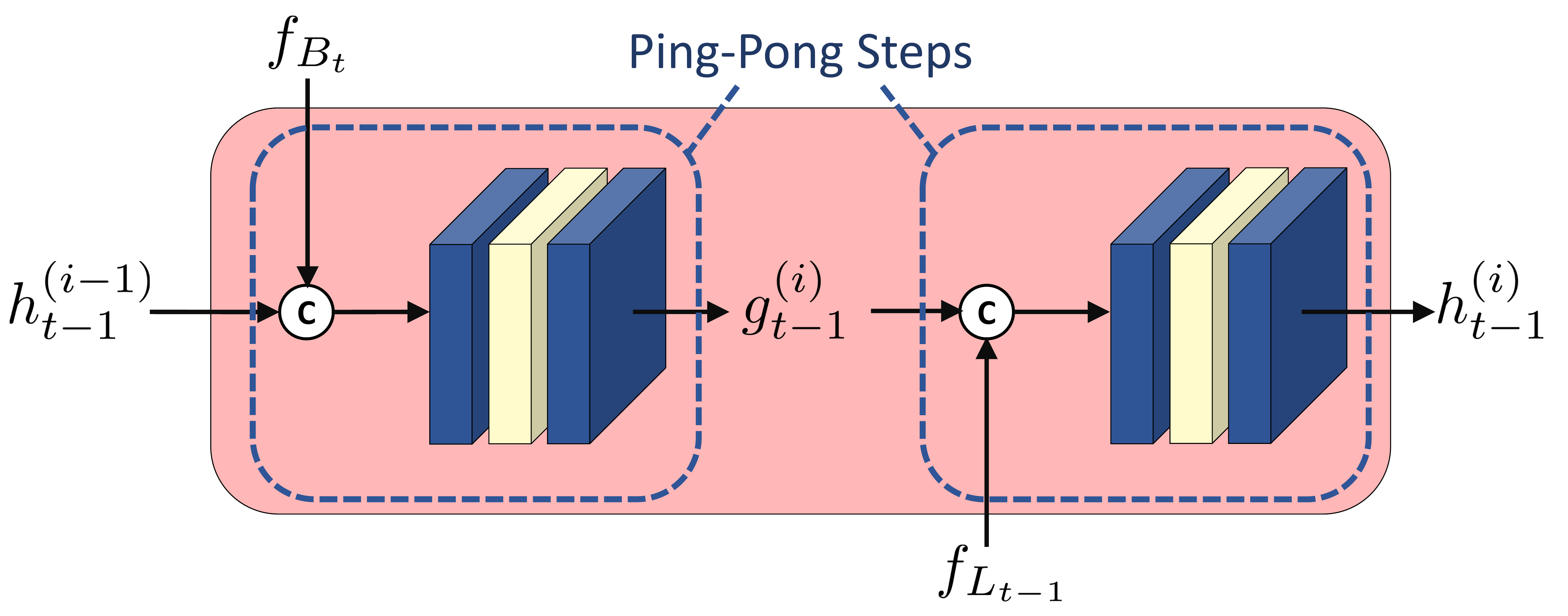}
    }
    \hhp
    \subfloat[SNLA module structure \label{sup_fig:snla}]{
    \includegraphics[height=0.15\linewidth]{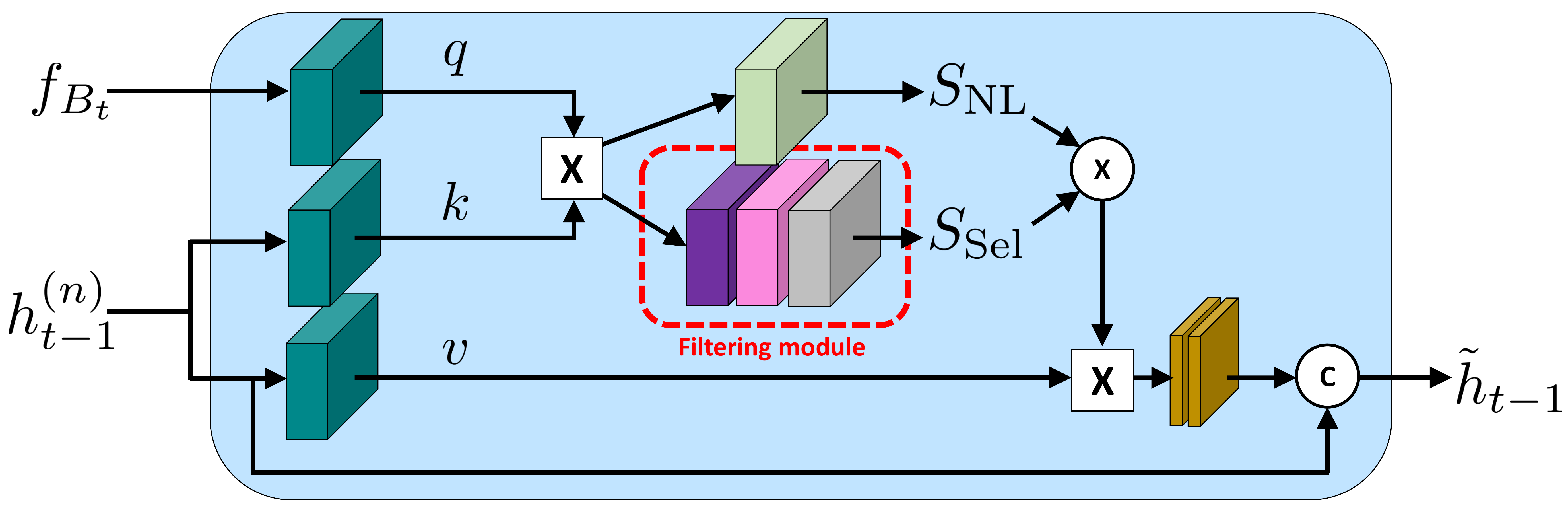}
    }
    \caption{\textbf{Overall architecture and components of our PAHS.}}
    \label{sup_fig:model}
\end{figure}

%% file: supplementary/table/feandre.tex
\begin{table}[t]
\centering
\begin{minipage}[b]{0.48\linewidth}
\resizebox{0.98\linewidth}{!}{
\begin{tabular}{c|c|c|c|c}
        \toprule
        Module & layer & kernel & stride & output shape\\
        \midrule
        $B_{t}$ & input & - & - & $3 \times h \times w$ \\
        \midrule
         & conv & $3 \times 3$ & 1 & $c/3 \times h \times w$\\
         & resblock $\times 5$  & $3 \times 3$ & 1 & $c/3 \times h \times w$\\
         & conv & $5 \times 5$ & 2 & $2c/3 \times h/2 \times w/2$\\
         & resblock $\times 5$  & $3 \times 3$ & 1 & $2c/3 \times h/2 \times w/2$\\
         & conv & $5 \times 5$ & 2 & $c \times h/4 \times w/4$\\
         & resblock $\times 5$  & $3 \times 3$ & 1 & $c \times h/4 \times w/4$\\
        \midrule
        ${f}_{B_t}$ & output &- & - & $c \times h/4 \times w/4$\\
        \bottomrule
\end{tabular}}
\vspace{1.5mm}
\caption{Feature extractor architecture details. $h$, $w$ are the number of height and width.}
\label{sup_tab:feature_extractor}
\end{minipage}%
\hfill
\begin{minipage}[b]{0.48\linewidth}
\centering
\resizebox{0.98\linewidth}{!}{
\begin{tabular}{c|c|c|c|c}
        \toprule
        Module & layer & kernel & stride & output shape\\
        \midrule
        $f_{L_{t}}$ & input & - & - & $c \times h/4 \times w/4$ \\
        \midrule
         & transposed conv & $3 \times 3$ & 2 & $2c/3 \times h/2 \times w/2$\\
         & resblock $\times 3$  & $3 \times 3$ & 1 & $2c/3 \times h/2 \times w/2$\\
         & transposed conv & $3 \times 3$ & 2 & $c/3 \times h \times w$\\
         & resblock $\times 3$  & $3 \times 3$ & 1 & $c/3 \times h \times w$\\
         & conv & $3 \times 3$ & 1 & $3 \times h \times w$\\
        \midrule
        $L_t$ & output &- & - & $3 \times h \times w$\\
        \bottomrule
\end{tabular}}
\vspace{1.5mm}
\caption{Reconstructor architecture details. $h$, $w$ are the number of height and width.}
\label{sup_tab:reconstructor}
\end{minipage}
\tabspace
\end{table}

%% file: supplementary/table/heandpprnn.tex
\begin{table}[h]
\vspace{-5mm}
\centering
\begin{minipage}[t]{0.48\linewidth}
\resizebox{0.98\linewidth}{!}{
\begin{tabular}{c|c|c|c|c}
        \toprule
        Module & layer & kernel & stride & output shape\\
        \midrule
        $f_{L_t}$ & input & - & - & $c \times h/4 \times w/4$ \\
        \midrule
         & conv & $3 \times 3$ & 1 & $c/3 \times h/4 \times w/4$\\
         & resblock  & $3 \times 3$ & 1 & $c/3 \times h/4 \times w/4$\\
         & conv & $3 \times 3$ & 1 & $c/3 \times h/4 \times w/4$\\
        \midrule
        ${h}_{t}$ & output &- & - & $c/3 \times h/4 \times w/4$\\
        \bottomrule
    \end{tabular}}
    \vspace{3mm}
    \caption{Hidden state extractor architecture details. $h$, $w$ are the number of height and width.}
    \label{sup_tab:hidden_state_extractor}
\end{minipage}%
\hfill
\begin{minipage}[t]{0.48\linewidth}
\centering
\resizebox{0.98\linewidth}{!}{
\begin{tabular}{c|c|c|c|c}
        \toprule
        Module & layer & kernel & stride & output shape\\
        \midrule
        $h^{(i-1)}_{t-1}$ or $g^{(i)}_{t-1}$ & input & - & - & $c/3 \times h/4 \times w/4$ \\
        $f_{B_t}$ or $f_{L_{t-1}}$ & input & - & - & $c \times h/4 \times w/4$ \\
        \midrule
         & concat & - & - & $4c/3 \times h/4 \times w/4$ \\ 
         & conv & $3 \times 3$ & 1 & $c/3 \times h/4 \times w/4$\\
         & resblock  & $3 \times 3$ & 1 & $c/3 \times h/4 \times w/4$\\
         & conv & $3 \times 3$ & 1 & $c/3 \times h/4 \times w/4$\\
        \midrule
        $h^{(i)}_{t-1}$ & output & - & - & $c/3 \times h/4 \times w/4$ \\
        \bottomrule
\end{tabular}}
\vspace{1.5mm}
\caption{PPRNN architecture details. $h$, $w$ are the number of height and width.}
\label{sup_tab:pprnn}
\end{minipage}
\vspace{-7mm}
\end{table}

%% file: supplementary/table/pprnn_algo.tex
\algdef{SE}[SUBALG]{Indent}{EndIndent}{}{\algorithmicend\ }
\algtext*{Indent}
\algtext*{EndIndent}

\vspace{-5mm}
\begin{algorithm}
	\caption{PPRNN} 
	\begin{algorithmic}[1]
        \State \textbf{procedure} PPRNN METHOD ($f_{B_t}$,$h_{t-1}$,$f_{L_t}$)
            \Indent
            \State $h^{(0,1)}_{t-1} = h_{t-1}$ 
	        \For{$i=1 \ldots N$}
	            \Indent
	            \State $g^{(i)}_{t-1}$ = ${\mathcal{M}_{\text{P}}}(f_{B_t},h^{(i-1)}_{t-1})$
	            \State $h^{(i)}_{t-1}$ = ${\mathcal{M}_{\text{P}}}(f_{L_{t-1}},g^{(i)}_{t-1})$
	            \EndIndent
	        \EndFor
	        \State \textbf{return} $h^{(n)}_{t-1}$
	        \EndIndent
	\end{algorithmic} 
	\label{algo:pprnn}
\end{algorithm}
\vspace{-5mm}

%% file: supplementary/table/snla_algo.tex
\algdef{SE}[SUBALG]{Indent}{EndIndent}{}{\algorithmicend\ }
\algtext*{Indent}
\algtext*{EndIndent}

\begin{algorithm}[t]
	\caption{SNLA} 
	\begin{algorithmic}[1]
        \State \textbf{procedure} SNLA METHOD ($f_{B_t}$,$h^{(n)}_{t-1}$)
	        \Indent
	        \State $(q,k,v) = (Q(f_{B_t}),K(h^{(n)}_{t-1}),V(h^{(n)}_{t-1}))$
            \State $S_{\text{nl}} = \text{softmax}({q}k^{T})$
            \State $S_{\text{sel}} = \text{sigmoid}(\text{FC}(\text{pool}({q}{k}^{T})))$
            \State $Att(q,k,v) = D(S_{\text{NL}} \otimes S_{\text{sel}}\times v)$
	    \State ${\tilde{h}}_{t-1} = h^{(n)}_{t-1} + Att(q,k,v)$
	    \State \textbf{return} ${\tilde{h}}_{t-1}$
	    \EndIndent
	\end{algorithmic}
	\label{algo:snla}
\end{algorithm}

%% file: supplementary/table/snla.tex
\begin{table}[h]
    \centering
    \resizebox{0.48\linewidth}{!}{
    \begin{tabular}{c|c|c|c|c|c}
        \toprule
        Input & Output &  layer & kernel & stride & output shape\\
        \midrule
        $f_{B_t}$ & - & - & - & - & $c \times h/4 \times w/4$ \\
        $h^{(n)}_{t-1}$ & - & - & - & - & $c/3 \times h/4 \times w/4$ \\
        \midrule
        $f_{B_t}$ & $q$ & conv & $4 \times 4$ & 4 & $c/3 \times h/16 \times w/16$\\
        $h^{(n)}_{t-1}$ & $k$ & conv & $4 \times 4$ & 4 & $c/3 \times h/16 \times w/16$\\
        $h^{(n)}_{t-1}$ & $v$ & conv & $4 \times 4$ & 4 & $c/3 \times h/16 \times w/16$\\
        \midrule
        $q$ & $q$ & reshape & - & - & $(h/16)(w/16) \times c/3$\\
        $k$ & $k$ & reshape & - & - & $(h/16)(w/16) \times c/3$\\
        $v$ & $v$ & reshape & - & - & $(h/16)(w/16) \times c/3$\\
        \midrule
        $q$ $\And$ $k$ & $qk^{T}$ & Matrix mult. & - & - & $(h/16)(w/16) \times (h/16)(w/16)$\\
        $qk^{T}$ & $S_{\text{NL}}$ & softmax & - & - & $(h/16)(w/16) \times (h/16)(w/16)$\\
        $qk^{T}$ & - &  Pool & - & - & $(h/16)(w/16) \times 1$\\
        - & - &  FC & - & - & $(h/16)(w/16) \times 1$\\
        - & $S_{\text{Sel}}$ & sigmoid & - & - & $(h/16)(w/16) \times 1$\\
        $S_{\text{NL}}$ $\And$ $S_{\text{Sel}}$ & $Att(q,k,v)$ & Element mult. & - & - & $(h/16)(w/16) \times (h/16)(w/16)$\\
        \midrule
        $Att(q,k,v)$ $\And$ $v$ & - & Matrix mult. & - & - & $(h/16)(w/16) \times c/3$\\ 
        - & - & reshape & - & - & $c/3 \times h/16 \times w/16 $\\ 
        - & - & transposed conv $\times 2$ & $3 \times 3$ & 2 & $c/3 \times h/4 \times w/4 $\\ 
        - & - & concat & - & - & $2c/3 \times h/4 \times w/4 $\\ 
        - & $\tilde{h}_{t-1}$ & conv & $1 \times 1$ & 1 & $c/3 \times h/4 \times w/4 $\\ 
        \midrule
        $\tilde{h}_{t-1}$ & output &- & - & - & $c/3 \times h/4 \times w/4$\\
        \bottomrule
    \end{tabular}}
    \vspace{1.5mm}
    \caption{SNLA extractor architecture details. $h$, $w$ are the number of height and width.}
    \label{sup_tab:snla}
    \vspace{-7mm}
\end{table}

%% file: supplementary/table/pprnn_ablation.tex
\begin{table}[h]
    \newcommand{\cmark}{\textcolor{MyGreen}{\ding{51}}}
    \newcommand{\xmark}{\ding{55}}%
    \centering
    \resizebox{0.5\linewidth}{!}{
    \begin{tabular}{c|c|c|cc}
        \toprule
        $f_{B_t}$ $\rightarrow$ $f_{L_{t-1}}$ & $f_{L_{t-1}}$ $\rightarrow$ $f_{B_{t}}$& recurrence number & PSNR & SSIM\\
        \midrule
        \xmark & \cmark & 1 & 33.34 & 0.9512 \\
        \cmark & \xmark & 1 & \textbf{33.40} & \textbf{0.9549} \\
        \xmark & \cmark & 2 & 33.42 & 0.9523 \\
        \cmark & \xmark & 2 & \textbf{33.49} & \textbf{0.9598} \\
        \xmark & \cmark & 3 & 33.50 & 0.9600 \\
        \cmark & \xmark & 3 & \textbf{33.69} & \textbf{0.9603} \\
        \xmark & \cmark & 4 & 33.62 & 0.9602 \\
        \cmark & \xmark & 4 & \textbf{33.82} & \textbf{0.9612} \\
        \bottomrule
    \end{tabular}}
    \vspace{1.5mm}
    \caption{Effects of using $f_{B_t}$ and $f_{L_{t-1}}$ in order to update hidden states in PPRNN.}
    \label{sup_tab:pprnn_ablation}
    \vspace{-7mm}
\end{table}

%% file: supplementary/table/snla_ablation.tex
\begin{table}[h]
    \vspace{-5mm}
    \centering
    \begin{tabular}{ccc|cc}
        \toprule
        query & key & value & PSNR & SSIM\\
        \midrule
        $h^{(n)}_{t-1}$ & $h^{(n)}_{t-1}$ & $h^{(n)}_{t-1}$ & 32.64 & 0.9482\\
        $f_{B_t}$ & $h^{(n)}_{t-1}$ & $h^{(n)}_{t-1}$  & \textbf{33.82} & \textbf{0.9612} \\
        \bottomrule
    \end{tabular}
    \vspace{1.5mm}
    \caption{Effects of using heterogeneous inputs for SNLA.}
    \label{sub_tab:snla_ablation}
    \tabspace
    \vspace{-7mm}
\end{table}

%% file: supplementary/table/pahs-s.tex
\begin{table}[h]
    \vspace{-5mm}
    \centering
    \resizebox{0.48\linewidth}{!}{
    \begin{tabular}{c|ccc}
        \toprule
         metric & DVD~[\textcolor{MyGreen}{24}] & GOPRO~[\textcolor{MyGreen}{19}] & REDS~[\textcolor{MyGreen}{18}]\\
        \midrule
        PSNR & 32.65 & 32.66 & 34.08 \\
        SSIM & 0.9459 & 0.9484 & 0.9477 \\
        \bottomrule
    \end{tabular}}
    \vspace{1.5mm}
    \caption{PSNR and SSIM for smaller version of PAHS, PAHS-S.}
    \label{sup_tab:pahs_s}
    \vspace{-7mm}
\end{table}

%% file: supplementary/fig/cnn_based.tex
\begin{figure}[t]
    \centering
    \captionsetup[subfloat]{font=tiny}
    \subfloat[TSP~\citenumber{21}~(28.11/0.836) \label{sup_fig:tsp1}]{
    \includegraphics[width=0.24\linewidth]{figs/dvd/tsp.png}}
    \hfill
    \subfloat[TSP*~\textbf{(28.17/0.838)} \label{sup_fig:tsp2}]{
    \includegraphics[width=0.24\linewidth]{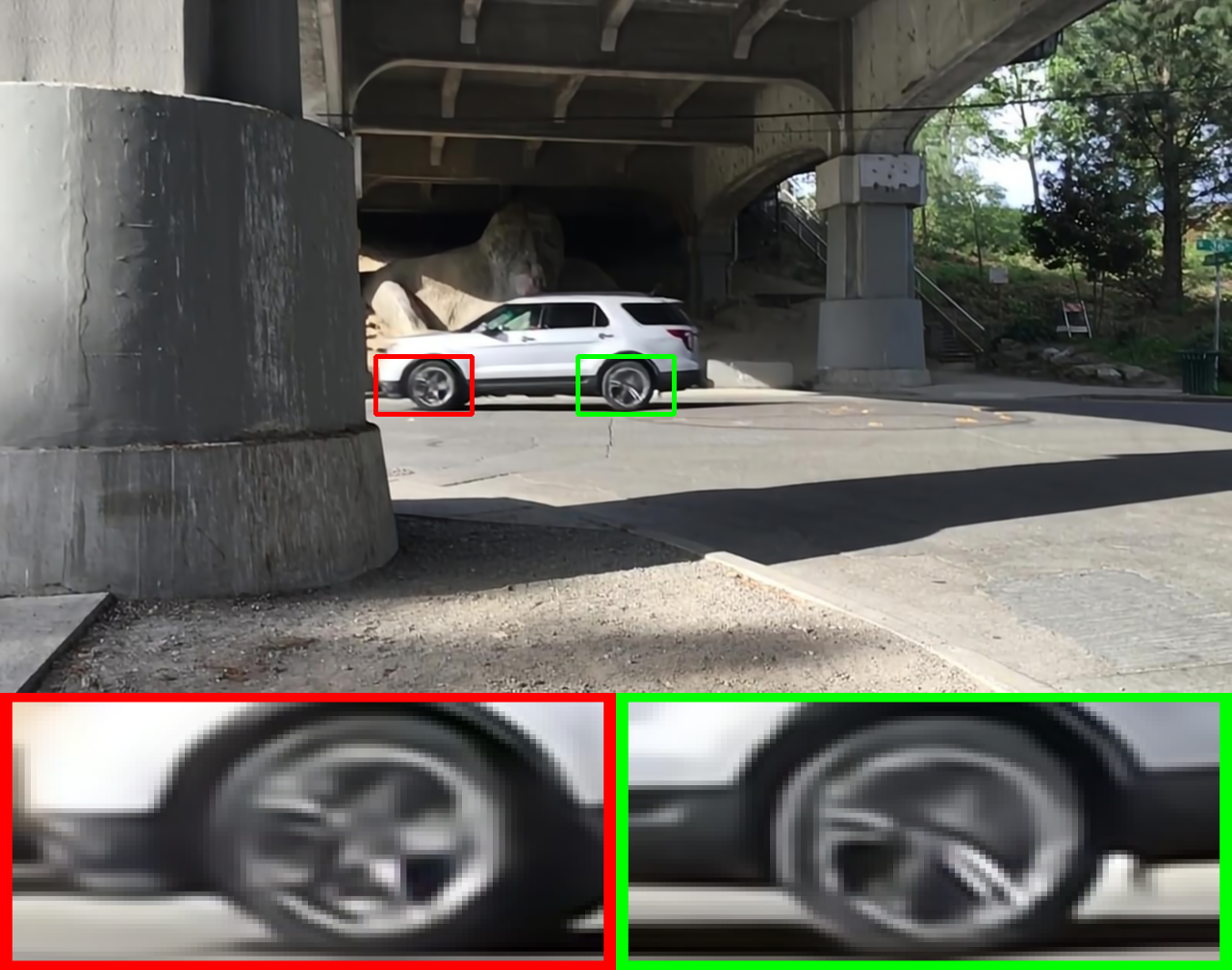}}
    \hfill
    \subfloat[ARVo~\citenumber{14}~(30.90/0.884) \label{sup_fig:arvo1}]{
    \includegraphics[width=0.24\linewidth]{figs/dvd/arvo.png}}
    \hfill
    \subfloat[ARVo*~\textbf{(31.19/0.890)} \label{sup_fig:arvo2}]{
    \includegraphics[width=0.24\linewidth]{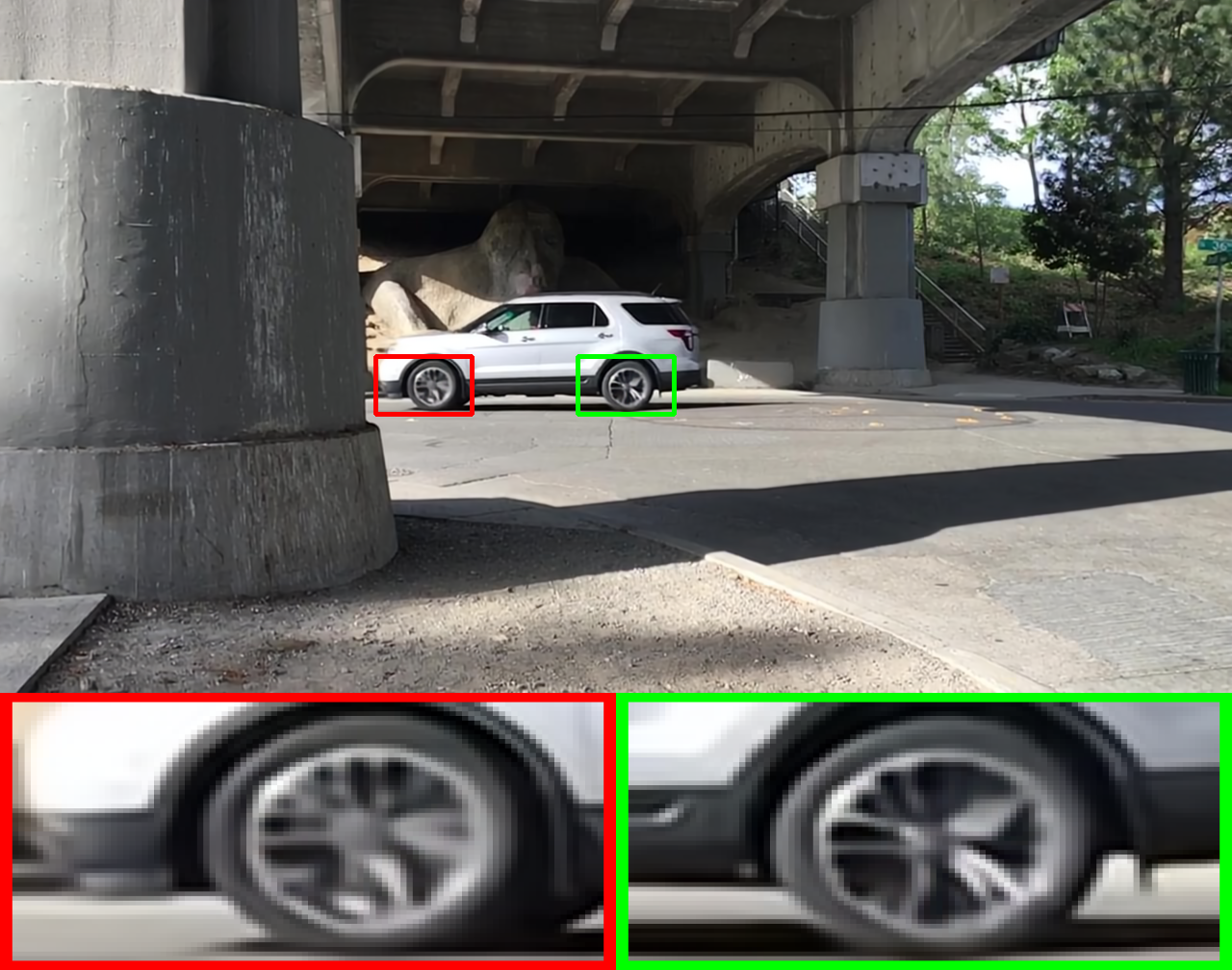}}
    \caption{\textbf{The visual results of replacing one of the blurry frames with sharp frame in CNN-based methods~\citenumber{14,21}.} $*$ denotes replacing one of the blurry frame with the ground truth sharp frame.}
    \label{sup_fig:cnn_based}
\end{figure}

%% file: supplementary/table/large_motion.tex
\begin{table}[h]
    \centering
    \resizebox{0.5\linewidth}{!}{
    \begin{tabular}{c|ccccc}
        \toprule
        models & ESTRNN~[\textcolor{MyGreen}{35}] & TSP~[\textcolor{MyGreen}{21}] & ARVo~[\textcolor{MyGreen}{14}] & PAHS-S & PAHS\\
        \midrule
        PSNR & 26.59 & 26.17 & \uline{27.86} & 27.38 & \textbf{28.93} \\
        SSIM & 0.8412 & 0.8371 & 0.8475 & \uline{0.8679} & \textbf{0.8891}\\
        \bottomrule
    \end{tabular}}
    \vspace{1.5mm}
    \caption{PSNR and SSIM for large-displacement data.}
    \label{sup_tab:large_motion}
    \vspace{-7mm}
\end{table}

%% file: supplementary/fig/real_comparison.tex
\begin{figure}[t]
\captionsetup[subfloat]{font=scriptsize}
    \renewcommand{\wp}{0.49\linewidth}
    \newcommand{\wwp}{0.35\linewidth}
    \newcommand{\hp}{\hspace{1mm}}
    \begin{minipage}{0.6\textwidth}
        \centering
        \subfloat[Blur]{\includegraphics[width=\wp]{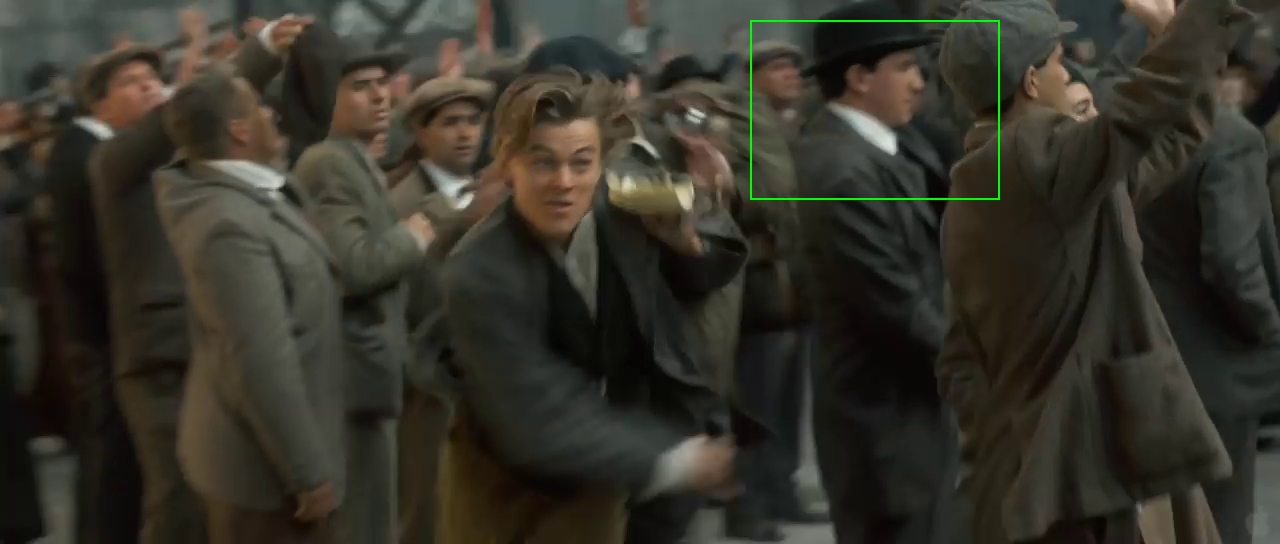}}
        \hp
        \subfloat[Deblurred (PAHS)]{\includegraphics[width=\wp]{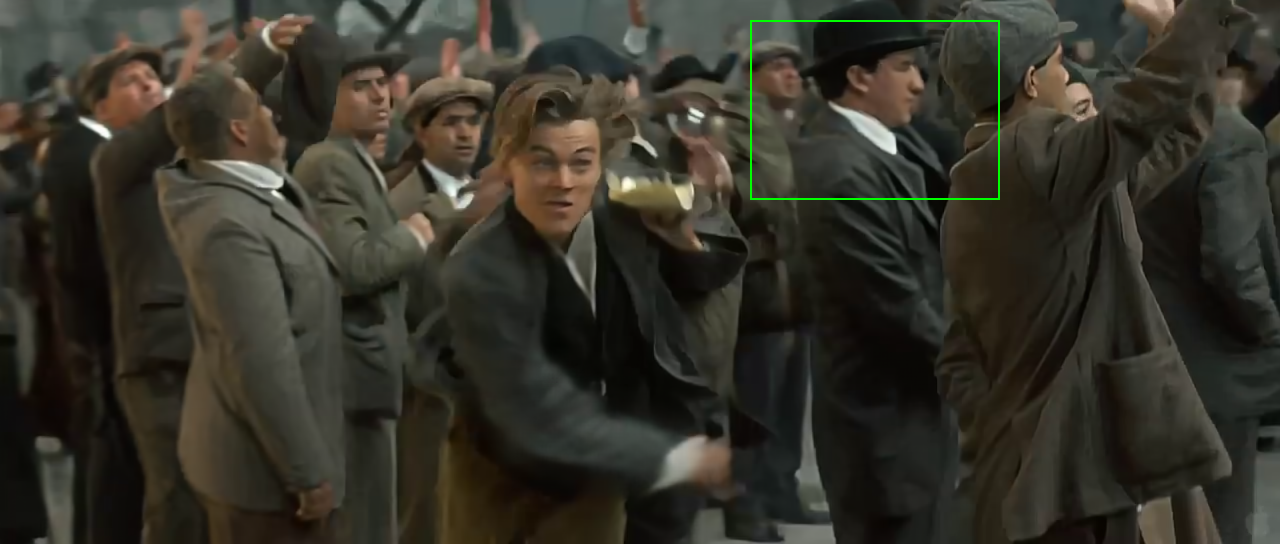}}
        \hp
    \end{minipage}
    \begin{minipage}{0.5\textwidth}
        \subfloat[Blur]{\includegraphics[width=\wwp]{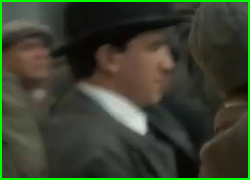}}
        \hp
        \subfloat[ESTRNN~\citenumber{35}]{\includegraphics[width=\wwp]{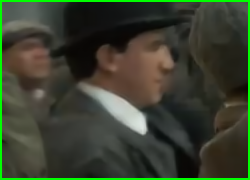}}
        \\
        \subfloat[ARVo~\citenumber{14}]{\includegraphics[width=\wwp]{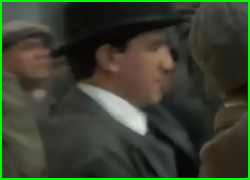}}
        \hp
        \subfloat[PAHS~(Ours)]{\includegraphics[width=\wwp]{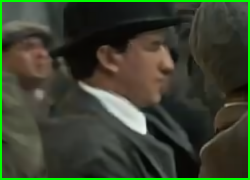}}
    \end{minipage}
    
    \figcspace
    \caption{Visual ablation showing the effect of PAHS on real blur dataset, downloaded from YouTube.}
    \label{fig:real_comparisons}
\end{figure}

%% file: supplementary/fig/dvd_comparison.tex
\begin{figure}[h]
\captionsetup[subfloat]{font=scriptsize}
    \renewcommand{\wp}{0.49\linewidth}
    \newcommand{\wwp}{0.35\linewidth}
    \newcommand{\hp}{\hspace{1mm}}
    \begin{minipage}{0.6\textwidth}
        \centering
        \subfloat[Blur]{\includegraphics[width=\wp]{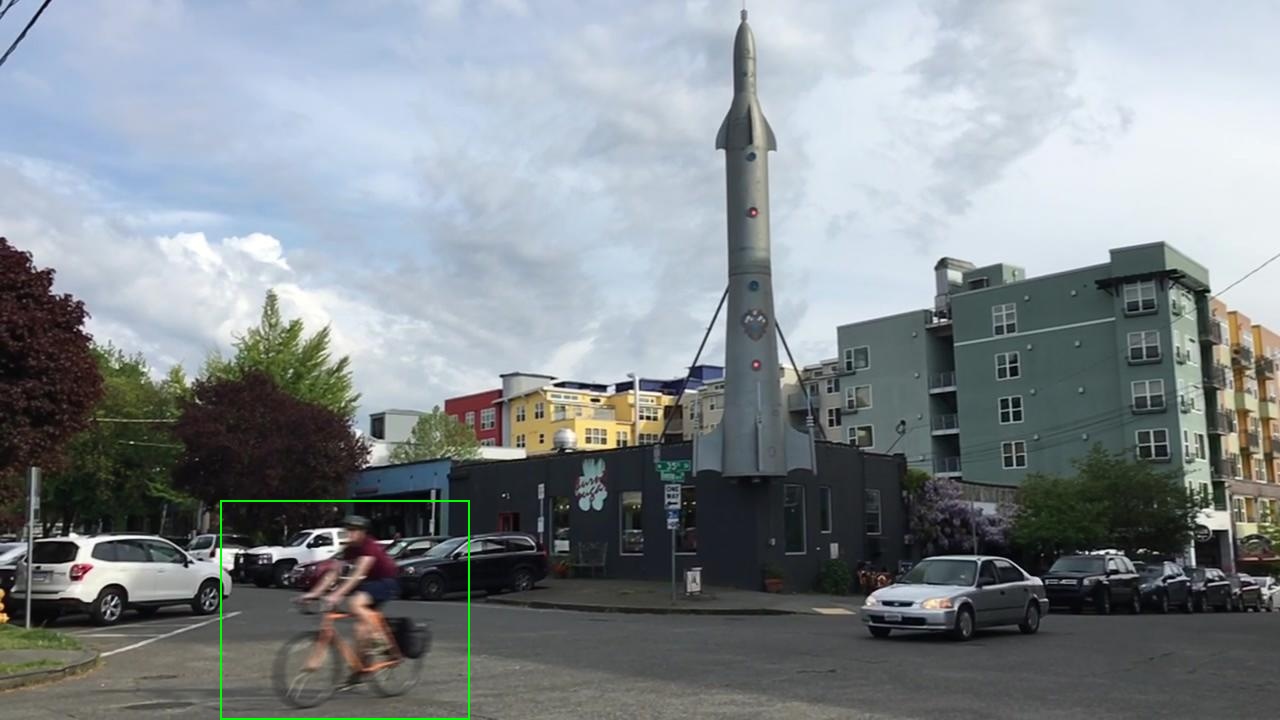}}
        \hp
        \subfloat[Deblurred (PAHS)]{\includegraphics[width=\wp]{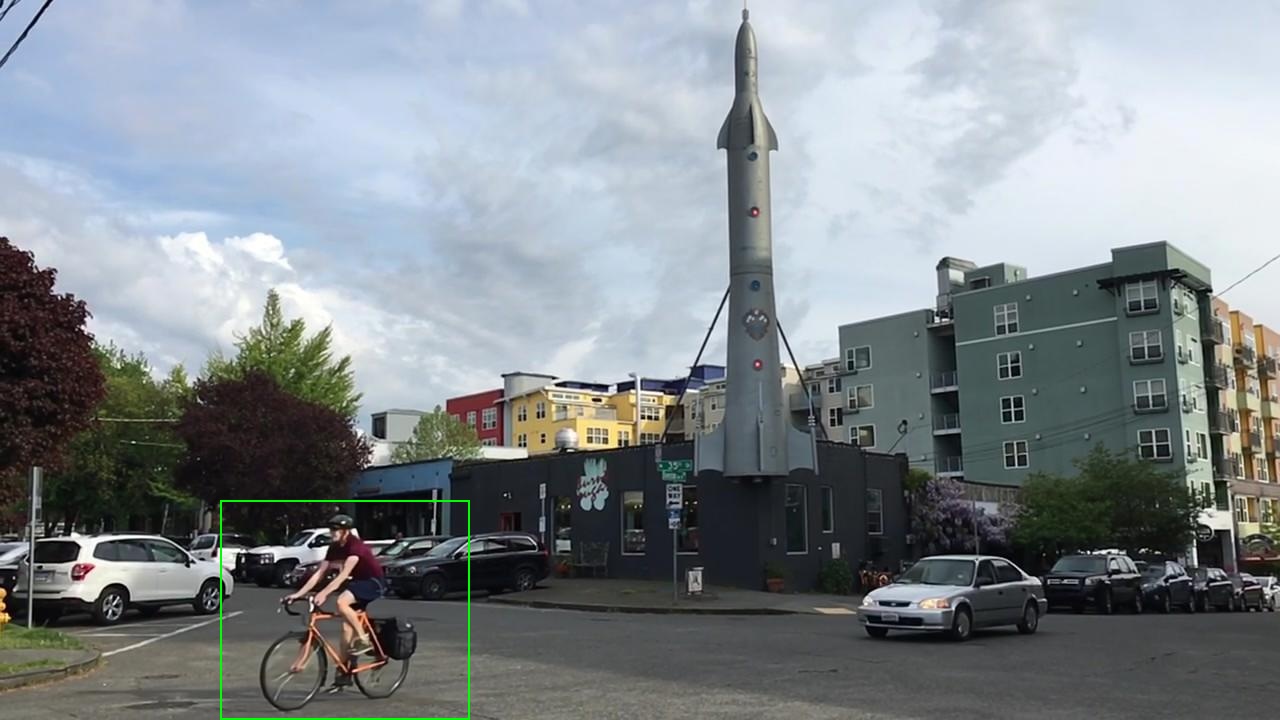}}
        \hp
    \end{minipage}
    \begin{minipage}{0.5\textwidth}
        \subfloat[Blur]{\includegraphics[width=\wwp]{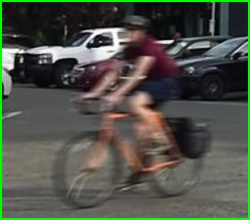}}
        \hp
        \subfloat[TSP~\citenumber{21}]{\includegraphics[width=\wwp]{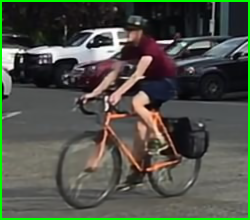}}
        \\
        \subfloat[ARVo~\citenumber{14}]{\includegraphics[width=\wwp]{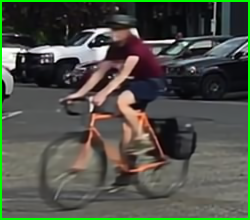}}
        \hp
        \subfloat[PAHS~(Ours)]{\includegraphics[width=\wwp]{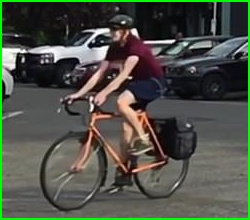}}
    \end{minipage}
    \begin{minipage}{0.6\textwidth}
        \vspace{5mm}
        \centering
        \subfloat[Blur]{\includegraphics[width=\wp]{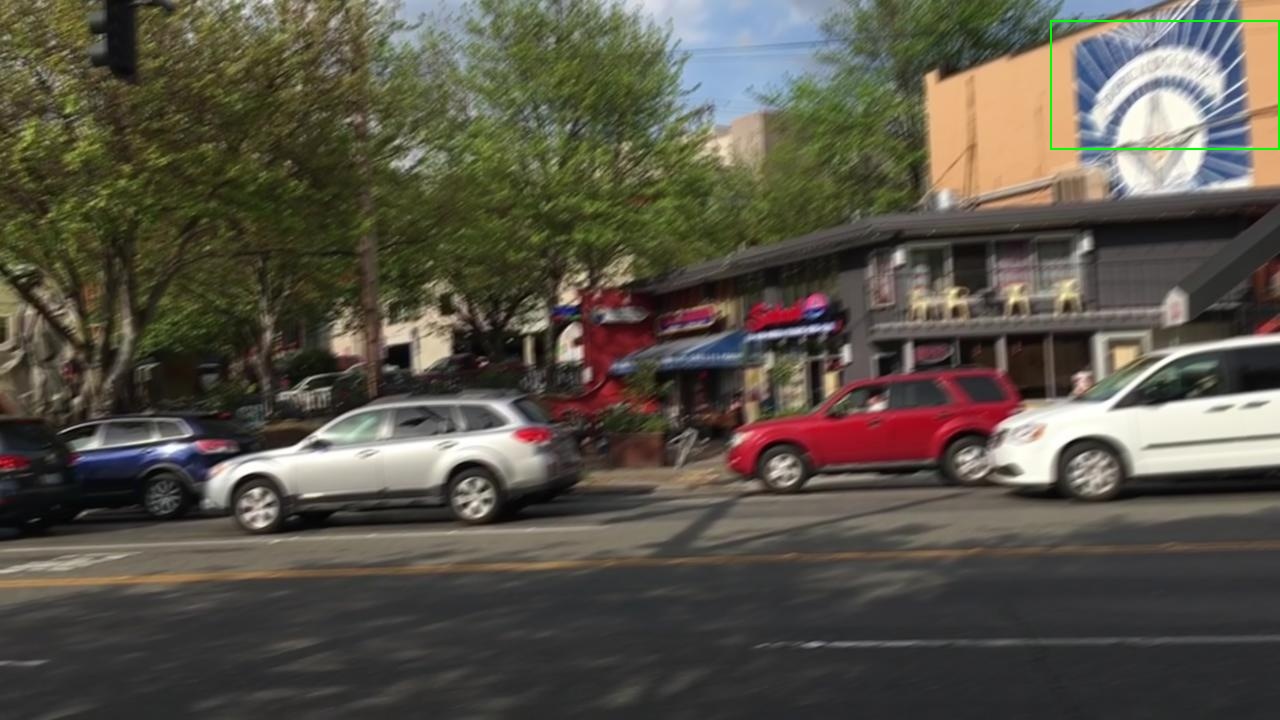}}
        \hp
        \subfloat[Deblurred (PAHS)]{\includegraphics[width=\wp]{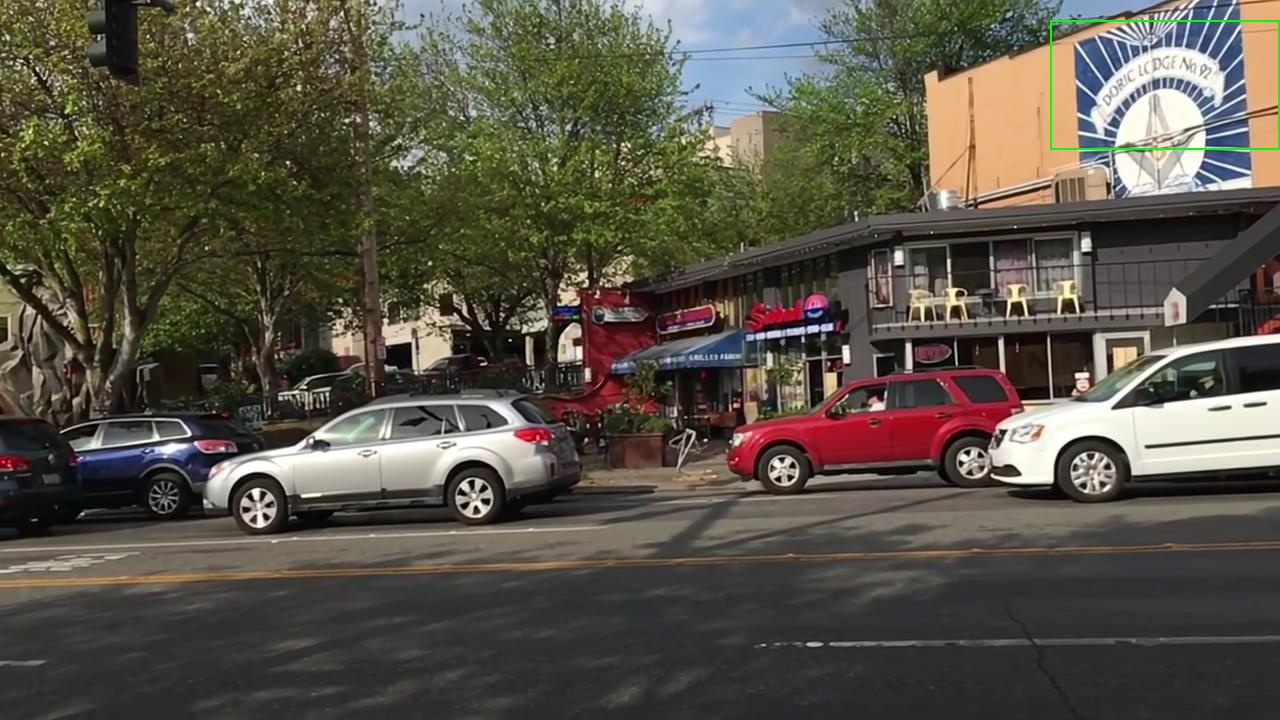}}
        \hp
    \end{minipage}
    \begin{minipage}{0.5\textwidth}
        \vspace{5mm}
        \subfloat[Blur]{\includegraphics[width=\wwp]{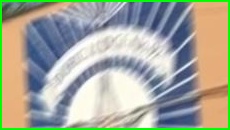}}
        \hp
        \subfloat[TSP~\citenumber{21}]{\includegraphics[width=\wwp]{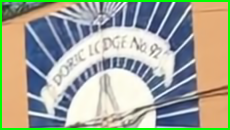}}
        \\
        \subfloat[ARVo~\citenumber{14}]{\includegraphics[width=\wwp]{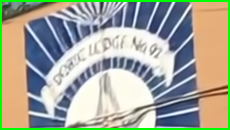}}
        \hp
        \subfloat[PAHS~(Ours)]{\includegraphics[width=\wwp]{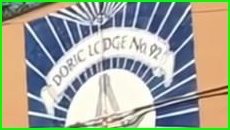}}
    \end{minipage}
    
    \figcspace
    \caption{Visual ablation showing the effect of PAHS on DVD~\citenumber{24} dataset.}
    \label{fig:dvd_comparisons}
\end{figure}

%% file: supplementary/fig/gopro_comparison.tex
\begin{figure}[h]
\captionsetup[subfloat]{font=scriptsize}
    \renewcommand{\wp}{0.49\linewidth}
    \newcommand{\wwp}{0.35\linewidth}
    \newcommand{\hp}{\hspace{1mm}}
    
    \begin{minipage}{0.6\textwidth}
        \centering
        \subfloat[Blur]{\includegraphics[width=\wp]{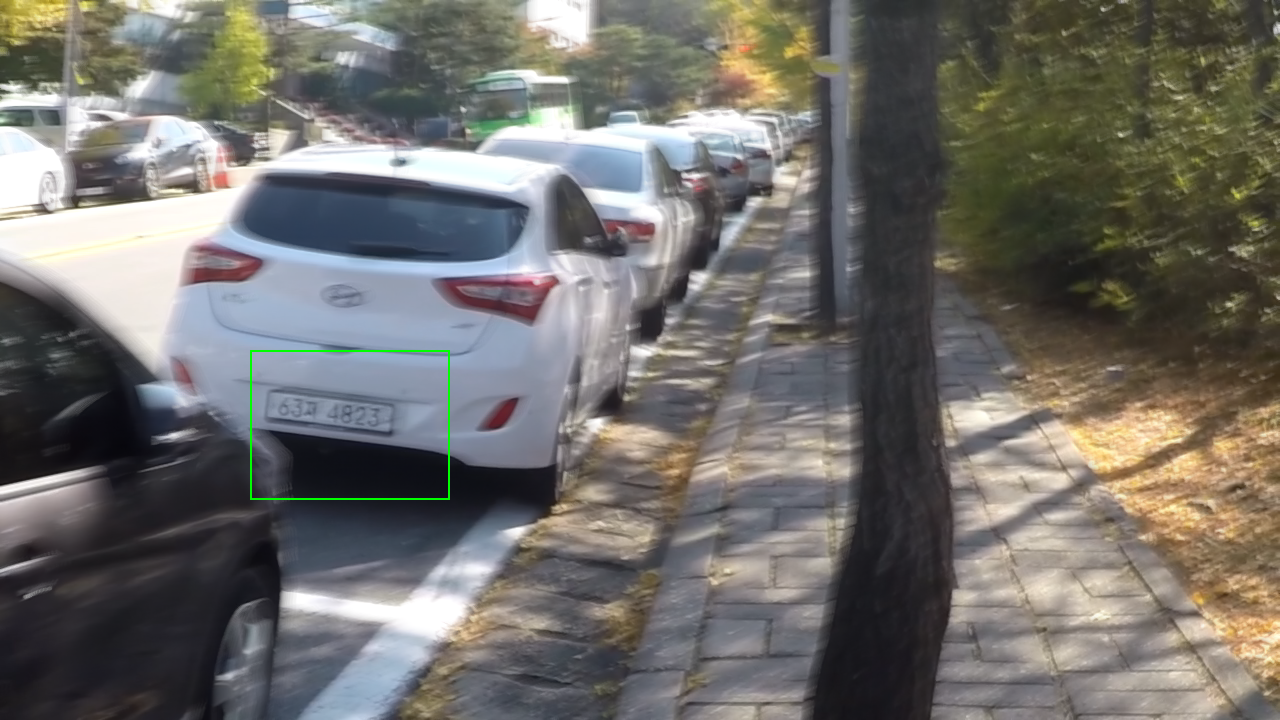}}
        \hp
        \subfloat[Deblurred (PAHS)]{\includegraphics[width=\wp]{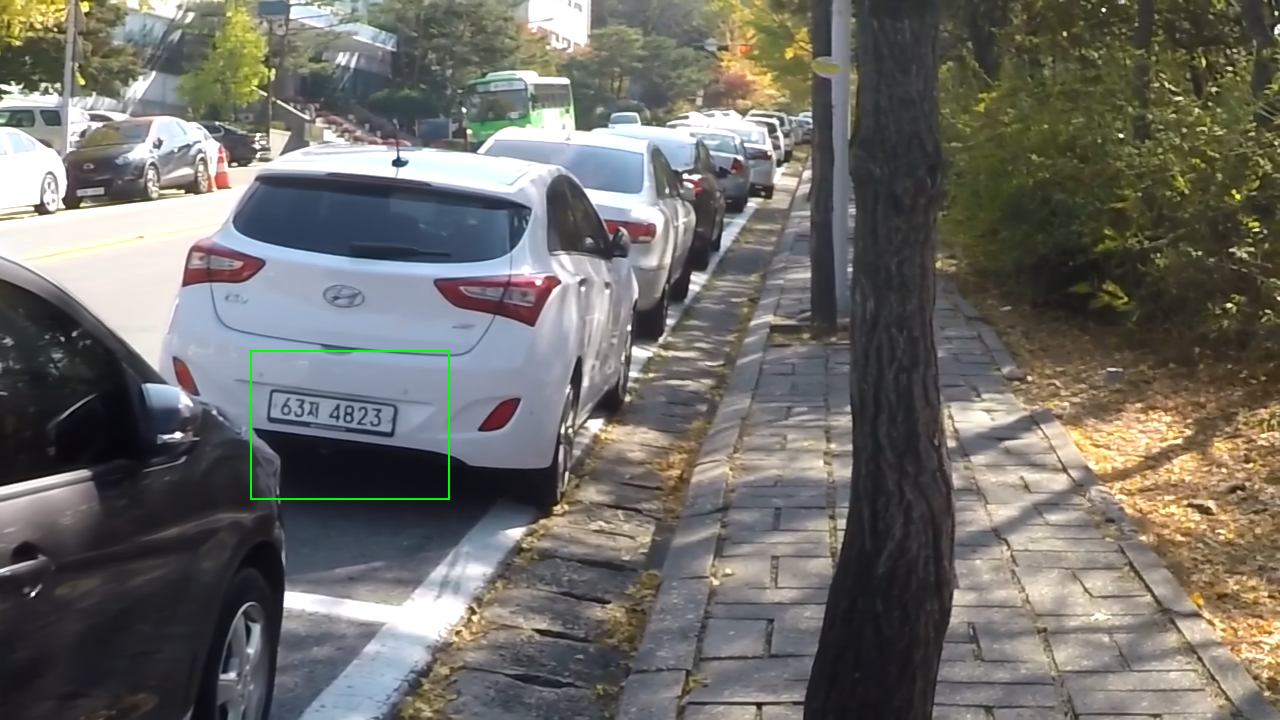}}
        \hp
    \end{minipage}
    \begin{minipage}{0.5\textwidth}
        \subfloat[Blur]{\includegraphics[width=\wwp]{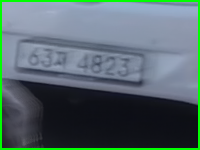}}
        \hp
        \subfloat[ESTRNN~\citenumber{35}]{\includegraphics[width=\wwp]{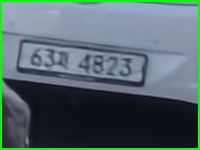}}
        \\
        \subfloat[TSP~\citenumber{21}]{\includegraphics[width=\wwp]{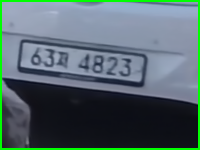}}
        \hp
        \subfloat[PAHS~(Ours)]{\includegraphics[width=\wwp]{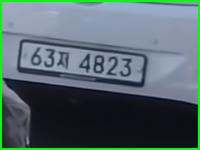}}
    \end{minipage}
    \begin{minipage}{0.6\textwidth}
        \vspace{5mm}
        \centering
        \subfloat[Blur]{\includegraphics[width=\wp]{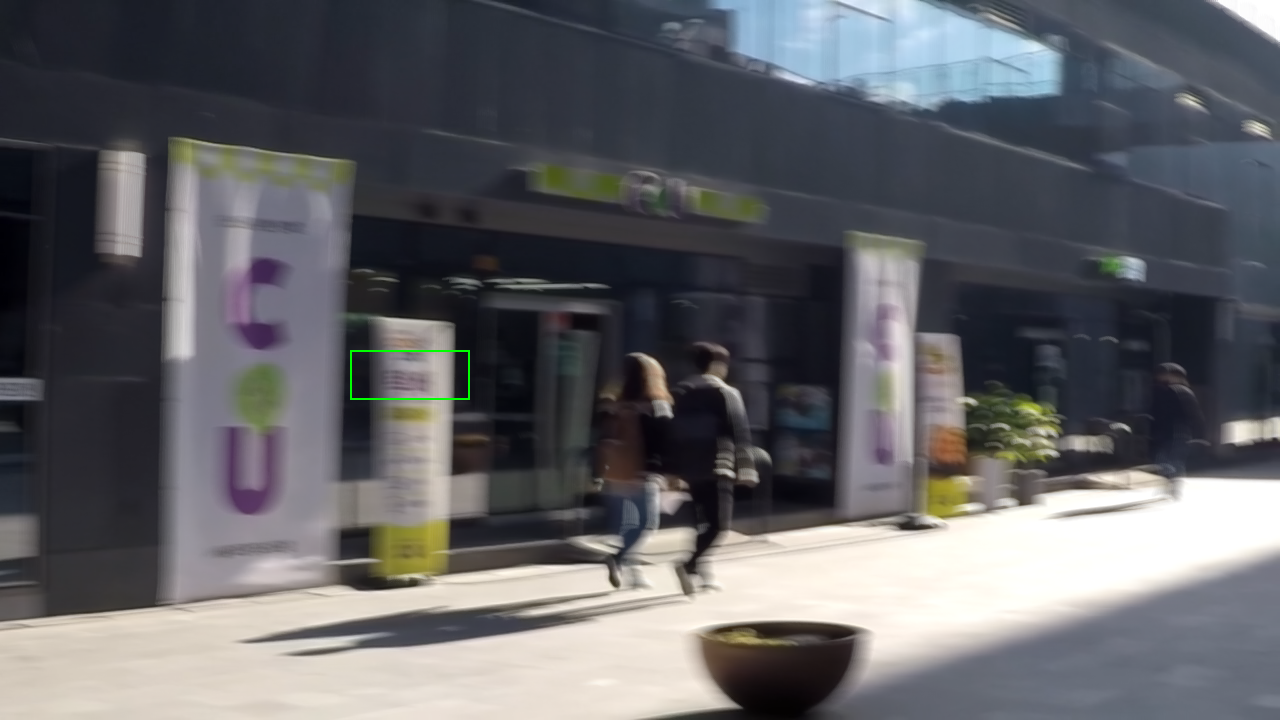}}
        \hp
        \subfloat[Deblurred (PAHS)]{\includegraphics[width=\wp]{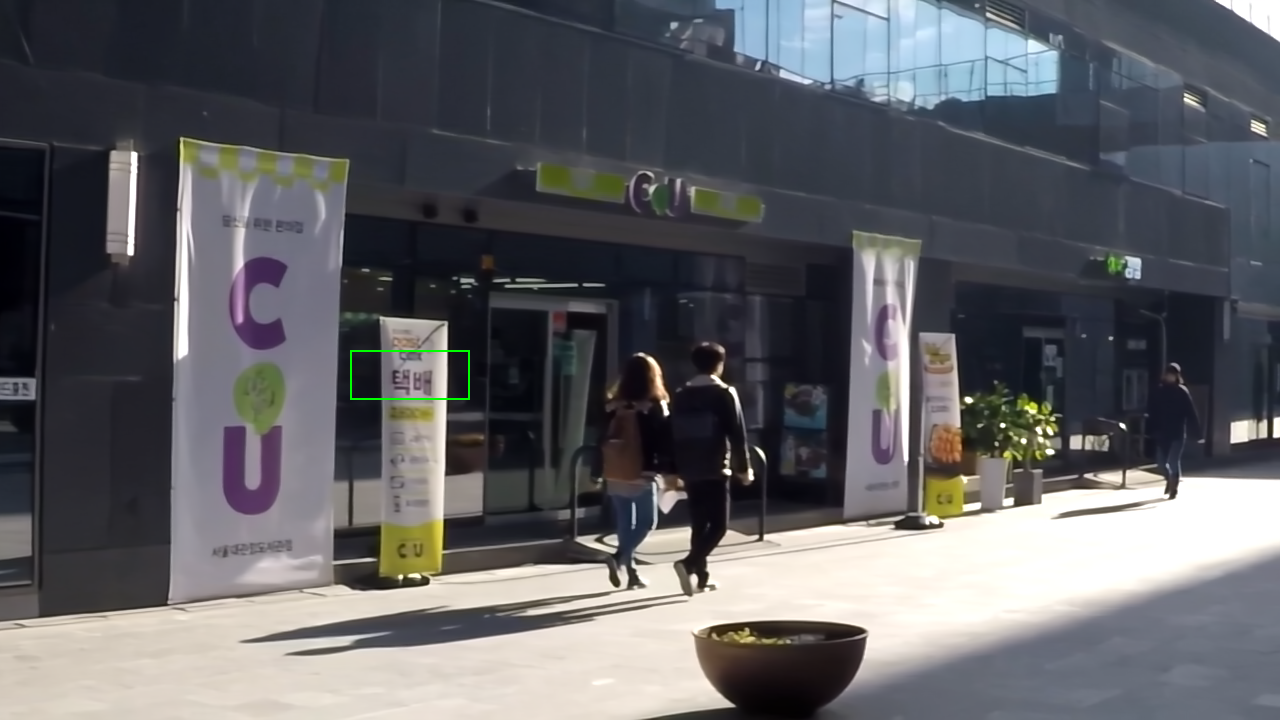}}
        \hp
    \end{minipage}
    \begin{minipage}{0.5\textwidth}
        \vspace{5mm}
       \subfloat[Blur]{\includegraphics[width=\wwp]{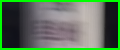}}
        \hp
        \subfloat[ESTRNN~\citenumber{35}]{\includegraphics[width=\wwp]{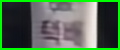}}
        \\
        \subfloat[TSP~\citenumber{21}]{\includegraphics[width=\wwp]{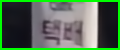}}
        \hp
        \subfloat[PAHS~(Ours)]{\includegraphics[width=\wwp]{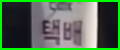}}
    \end{minipage}
    
    \figcspace
    \caption{Visual ablation showing the effect of PAHS on GOPRO~\citenumber{19} dataset.}
    \label{fig:gopro_comparisons}
\end{figure}

%% file: supplementary/fig/reds_comparison.tex
\begin{figure}[h]
\captionsetup[subfloat]{font=scriptsize}
    \renewcommand{\wp}{0.49\linewidth}
    \newcommand{\wwp}{0.35\linewidth}
    \newcommand{\hp}{\hspace{1mm}}
    
    \begin{minipage}{0.6\textwidth}
        \centering
        \subfloat[Blur]{\includegraphics[width=\wp]{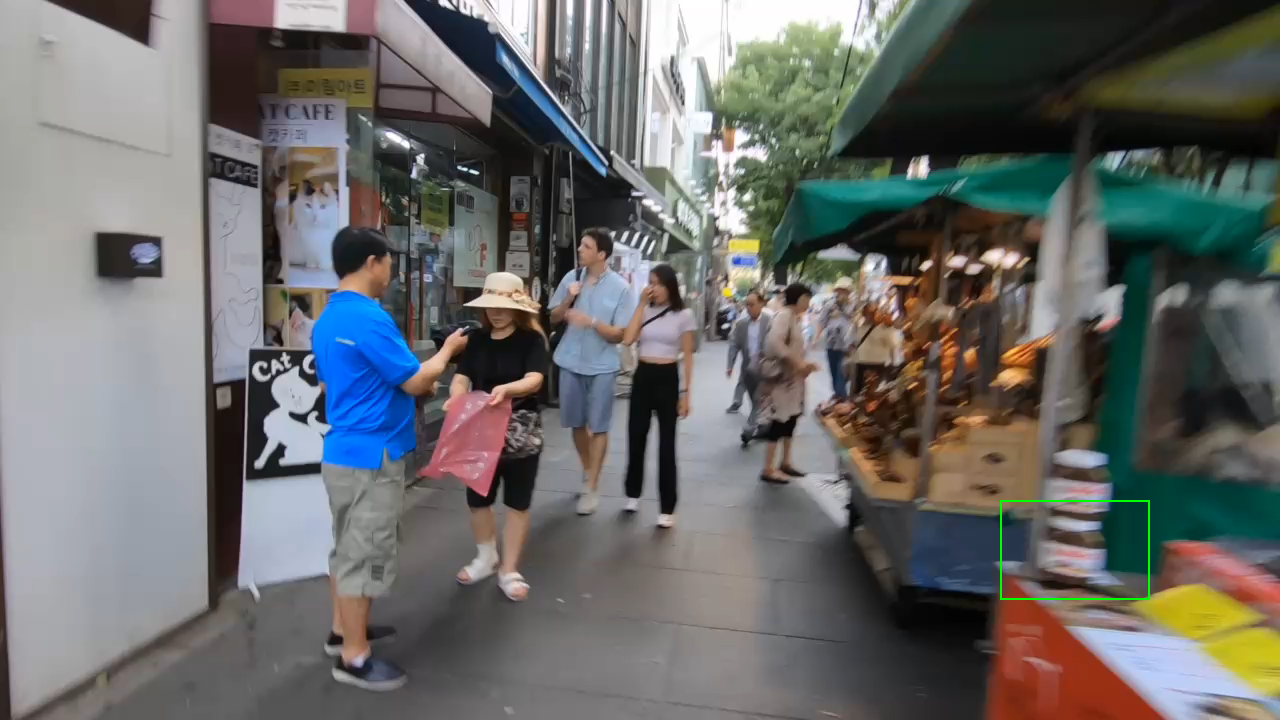}}
        \hp
        \subfloat[Deblurred (PAHS)]{\includegraphics[width=\wp]{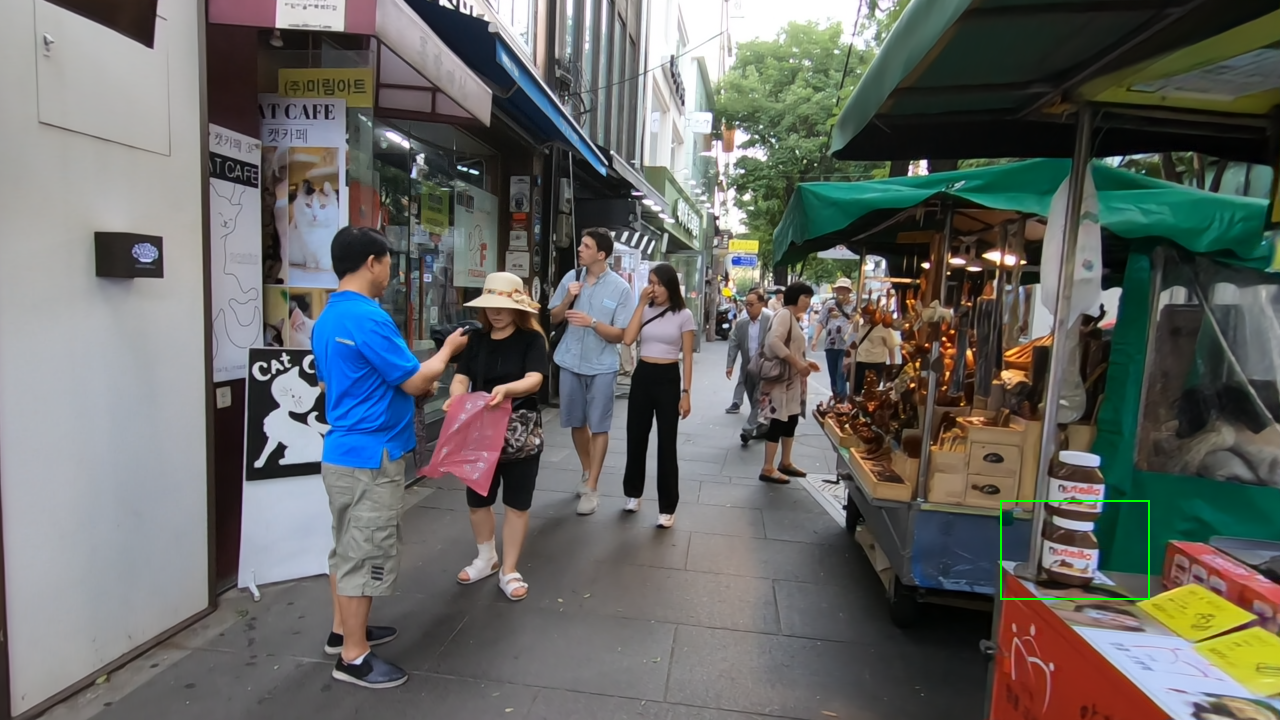}}
        \hp
    \end{minipage}
    \begin{minipage}{0.5\textwidth}
        \subfloat[Blur]{\includegraphics[width=\wwp]{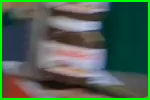}}
        \hp
        \subfloat[IFI-RNN~\citenumber{20}]{\includegraphics[width=\wwp]{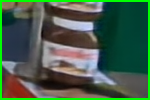}}
        \\
        \subfloat[ESTRNN~\citenumber{35}]{\includegraphics[width=\wwp]{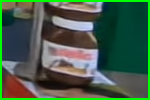}}
        \hp
        \subfloat[PAHS~(Ours)]{\includegraphics[width=\wwp]{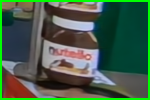}}
    \end{minipage}
    \begin{minipage}{0.6\textwidth}
        \vspace{5mm}
        \centering
        \subfloat[Blur]{\includegraphics[width=\wp]{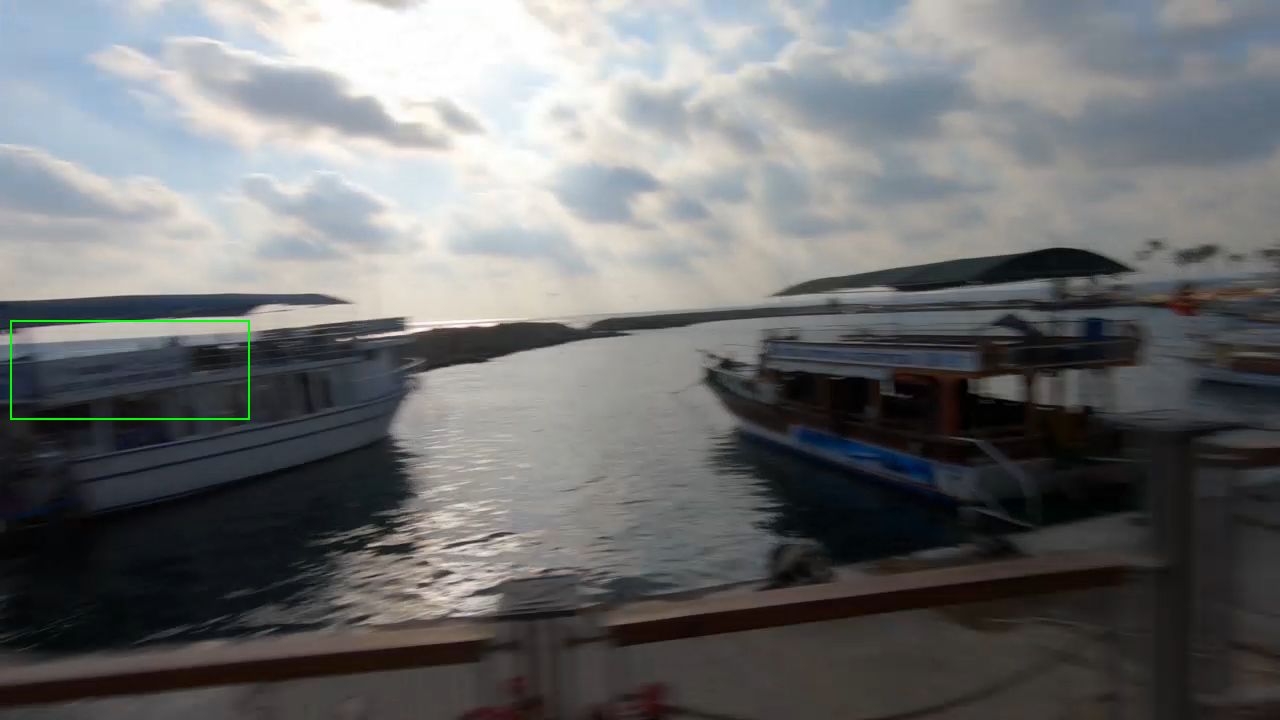}}
        \hp
        \subfloat[Deblurred (PAHS)]{\includegraphics[width=\wp]{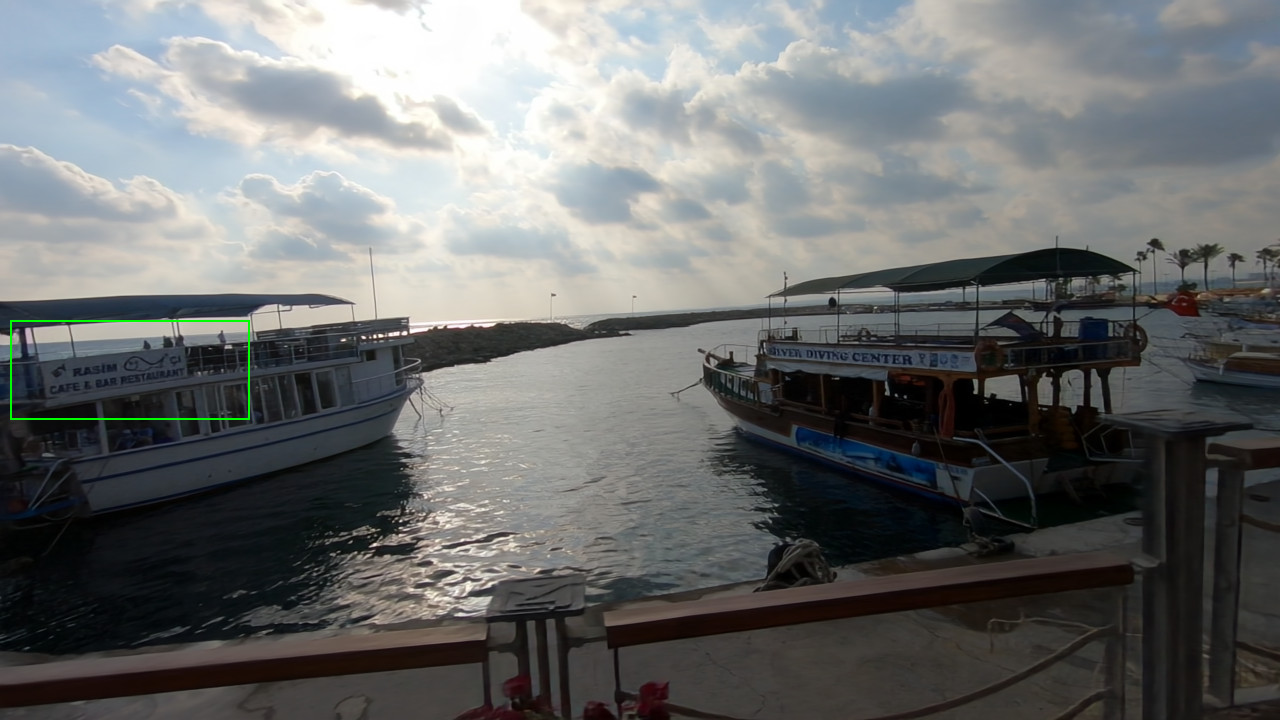}}
        \hp
    \end{minipage}
    \begin{minipage}{0.5\textwidth}
        \vspace{5mm}
        \subfloat[Blur]{\includegraphics[width=\wwp]{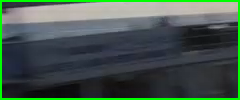}}
        \hp
        \subfloat[IFI-RNN~\citenumber{20}]{\includegraphics[width=\wwp]{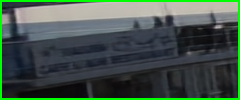}}
        \\
        \subfloat[ESTRNN~\citenumber{35}]{\includegraphics[width=\wwp]{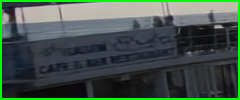}}
        \hp
        \subfloat[PAHS~(Ours)]{\includegraphics[width=\wwp]{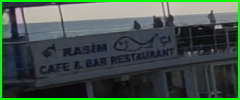}}
    \end{minipage}
    
    \figcspace
    \caption{Visual ablation showing the effect of PAHS on REDS~\citenumber{18} dataset.}
    \label{fig:reds_comparisons}
\end{figure}